\documentclass[conference, 12pt]{IEEEtran}
\IEEEoverridecommandlockouts
\usepackage{cite}
\usepackage{amsmath,amssymb,amsfonts}
\usepackage{algorithmic}
\usepackage{tabularray}
\usepackage{graphicx}
\usepackage{textcomp}
\usepackage{fancyhdr}
\usepackage{xcolor}
\usepackage{float}
\usepackage{booktabs}
\usepackage[T1]{fontenc}
\def\BibTeX{{\rm B\kern-.05em{\sc i\kern-.025em b}\kern-.08em
    T\kern-.1667em\lower.7ex\hbox{E}\kern-.125emX}}
\usepackage{fancyhdr}
\usepackage{url}
\usepackage{graphicx}
\usepackage[normalem]{ulem}
\usepackage{subcaption}
\usepackage{caption}
\usepackage{float}
\usepackage{xspace}
\usepackage{enumitem,kantlipsum}
\usepackage{rotating}

\fancypagestyle{IEEEtitlepagestyle}{
  \fancyhf{} 
   
  \setlength{\headheight}{15.31181pt}
  \fancyhead[R]{\raisebox{-1ex}{Duke Robotics Club | \thepage}} 
}
\begin{document}

\title{\textbf{Technical Design Review of Duke Robotics Club's Oogway: An AUV for RoboSub 2024} 
}

\author{Will Denton, Michael Bryant, Lilly Chiavetta, Vedarsh Shah, Rico Zhu, Philip Xue,\\Vincent Chen, Maxwell Lin, Hung Le, Austin Camacho, Raul Galvez, Nathan Yang,\\ Nathanael Ren, Tyler Rose, Mathew Chu, Amir Ergashev, Saagar Arya, Kaelyn Pieter,\\Ethan Horowitz, Maanav Allampallam, Patrick Zheng, Mia Kaarls, June Wood  }

\maketitle

\begin{abstract}
\normalsize
The Duke Robotics Club is proud to present our robot for the 2024 RoboSub Competition: Oogway. Now in its second year, Oogway has been dramatically upgraded in both its capabilities and reliability. Oogway was built on the principle of independent, well-integrated, and reliable subsystems. Individual components and subsystems were tested and designed separately. Oogway's most advanced capabilities are a result of the tight integration between these subsystems. Such examples include a re-envisioned controls system, an entirely new electrical stack, advanced sonar integration, additional cameras and system monitoring, a new marker dropper, and a watertight capsule mechanism. These additions enabled Oogway to prequalify for Robosub 2024.
\end{abstract}

\section{Competition Goals}
At previous RoboSub competitions, we prioritized versatility over reliability. While our previous designs allowed us to complete more tasks, our runs were inconsistent and unreliable. With only three scoring runs, we need to minimize the complexity of our systems to ensure they are reliable. 

Our strategic vision consists of \textit{core tasks}, which the robot is programmed to perform every run, and one \textit{alternate task} which may be completed at the end of a successful run if more points are necessary. 

RoboSub's design goals focus on four fundamental principles: movement, vision, manipulation, and acoustic tracking. To accomplish our strategic vision we chose to focus on 3 of 4 design goals: movement, vision, and manipulation. This subset of goals aligns with our team's strengths in controls, computer vision (CV), and mechanics. Our focus on these methods enables us to perform the gate, buoy, and bins (our \textit{core tasks}) reliably, while still maintaining the flexibility to attempt the octagon task (our \textit{alternate task}) if needed. 

\vspace{-3pt}

\subsection{Strategic Vision}
Our strategy this year focuses on completing our \textit{core tasks} while having fail-safes in place to ensure a successful run. This year, we upgraded Oogway with new peripherals to ensure that we can always complete our \textit{core tasks} using multiple methods. 

Reflecting on last year's competition, we noticed two large issues which prevented us from achieving our strategic vision: the unreliability of our movement and acoustic subsystems. Since movement is essential to all Robosub tasks we spent the first six months of this season completely remaking our control system and testing it in accordance with our testing plan. When discussing acoustics, we realized that we would not have enough time this year to make the system reliable. Instead of redesigning our complex acoustic system, we instead decided to create a simple and reliable marker dropper system for use in the bins task.

With our new controls system and marker dropper, we decided to change our \textit{core tasks} from last year by removing the octagon and adding the marker dropper. In addition to these new systems, we also added multiple new cameras and a second sonar which allow us to fulfill our strategic vision by adding reliability to the gate, buoy, and bins tasks.

\subsection{Enter the Pacific - Gate}
Since the gate task is required to start a run, we use multiple subsystems to ensure that we can reliably move through the gate. With our CV subsystem, Oogway can identify the gate from any angle and get an estimated position of symbols beneath the gate. If the robot cannot find the gate using CV, we can use our sonar to get the gate's approximate position. Finally, if neither of these methods work, our new controls system is accurate enough to dead reckon through the gate. Given these fallback methods, we will elect to do the coin flip for bonus points. We will always pass through the Hot/CW side of the gate to ensure we can drop our markers in the Hot side of the bins task. Our revamped controls also allows us to gain bonus style points by completing a barrel roll while moving through the gate.

\subsection{Hydrothermal Vent - Buoy}
The buoy task is a core component of our competition strategy. After we complete the barrel roll, we use our newly added downward-facing camera in combination with our CV system to find and track the path. Once in range, CV identifies the position of the buoy. If unsuccessful, we use sonar to search up to 10m away to get a rough estimate of where the buoy is located. Since our new controls is reliable, we will circumnavigate the buoy in the clockwise direction for bonus circumnavigation points.

Since we do not have a torpedo system, we will forego using the torpedoes to hit the buoy.

\subsection{Ocean Temperatures - Bins}
The final component of our core tasks is dropping both markers in the Hot side of the bin. After the buoy, we use the path to navigate towards the bins until the downward camera can identify the red side of the bin. We then line up, hold our position a foot away from the bins, and drop both markers. In case we go off-course from the path, we perform a search around the area until Oogway can find the bins.

When weighing the trade off between last year's final core task, the octagon, and the bins, we decided that since our movement and camera tracking is much more reliable than our acoustic tracking. Thus, completing the bins task will reliably get us more points in this year's competition

\subsection{Collect Samples - Octagon}
With limited reliability in our acoustic tracking systems, we made the octagon task our \textit{alternate task}. If we need more points at the end of a run we can use our current acoustics and sonar systems to attempt to find and surface inside of the octagon. After the bins task, we can enable our acoustics and perform large sonar sweeps until we have a "guess" at the octagon location. Since this is not reliable, we decided to leave this task until the end in case we surface outside of the octagon and end our run.

Since our actuators are still in development, we opted to skip the sample collecting task.

\subsection{Mapping - Torpedoes}
We are not planning on attempting the torpedoes task since Oogway does not have a torpedo system. However, we have this system in development and plan to use it next year.

\subsection{Trade Offs Between Complexity and Reliability}
Given last year's competition experience, we realized that many of our systems were too complex and did not have enough testing. We attempted to perform tasks at competition that were untested, resulting in failed runs. For this year's competition, we decided to remake many of Oogway's systems to decrease overall complexity while increasing reliability. Additionally, we developed an open-source GUI to interface with Oogway while in the water which allowed us to test individual components quickly. By narrowing our focus to 3 out of the 5 tasks during the design process for this year's competition, we allocated more testing time for each task to ensure that our integration can work reliably. 

\section{Design Strategy}

\subsection{Mechanical Design}

The main mechanical structure of Oogway has not changed from last year. However, we fixed major issues with leaks in the capsule and utilized our modular design to add peripherals such as a marker dropper and an adjustable buoyancy system.

\subsubsection{Reworked Capsule Design}

During last year's competition, we had multiple small leaks inside our capsule. The leaks were caused by the four brass inserts in our capsule slowly stripping out from the surrounding plastic due to the stress of holding a seal. To make the capsule watertight, we created a new locking mechanism that secures the capsule using a custom 0.125" Grade V titanium plate, chosen for its corrosion resistance and high yield strength to weight ratio \cite{b4}. 

\vspace{-5pt}

\begin{figure}[h!]
\centering
\includegraphics[width=50mm]{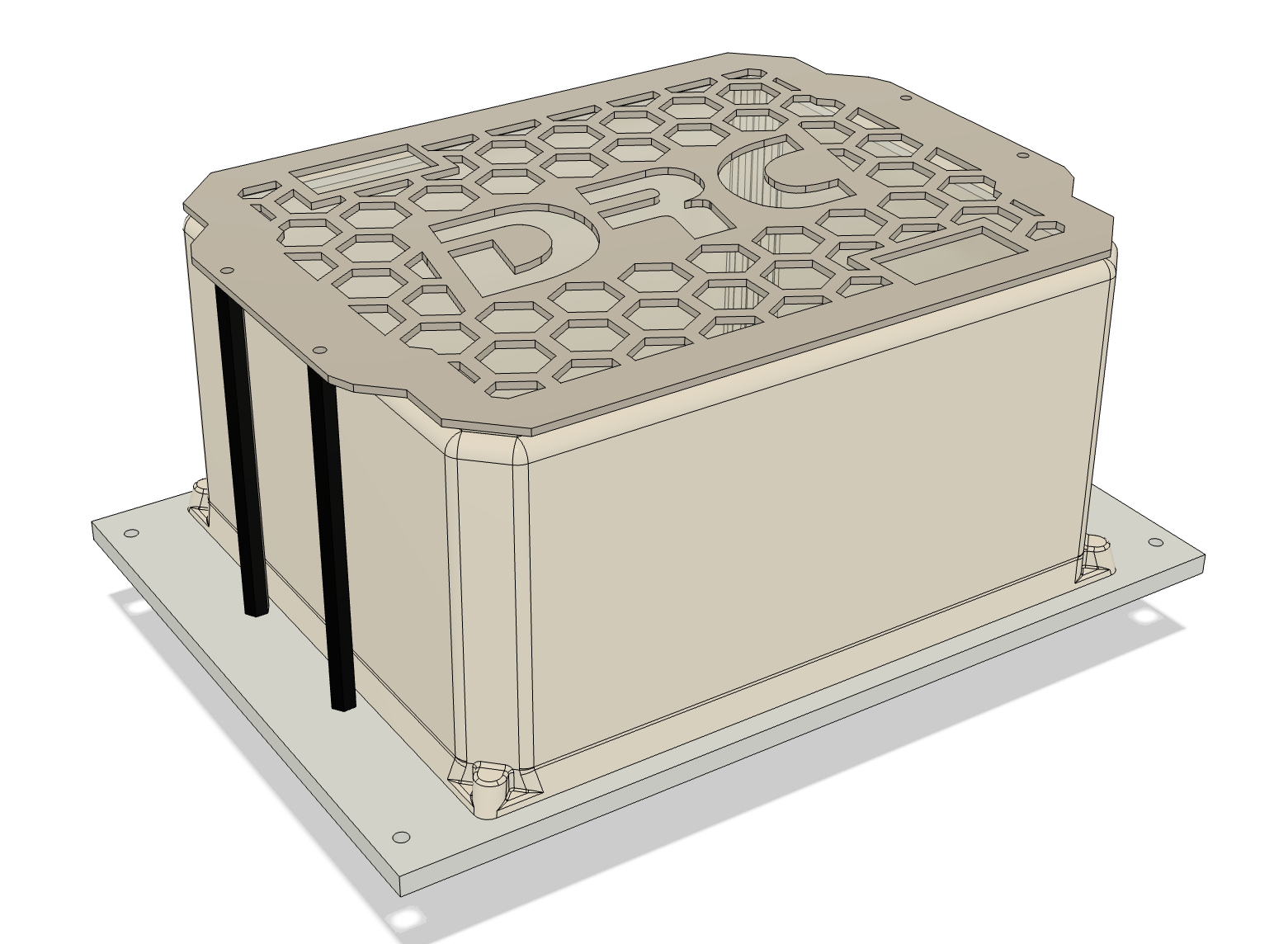}
\caption{New Capsule and Plate Design}
\label{fig:newplatedesign}
\end{figure}

\vspace{-5pt}

The hexagonal design reduces weight while still maintaining rigidity under pressure \cite{b5}. The plate is held by precision-cut hexagonal aluminum standoffs that are connected to Oogway's frame. 

In addition to a significant reduction in leaks, the new system is also simpler to secure and remove as all screws are top-facing. Installation time has been reduced from several minutes to under a minute. 

\subsubsection{Modular Peripheral Components}

\begin{figure}[h!]
    \centering
    \begin{subfigure}[b]{0.45\columnwidth}
        \centering
        \includegraphics[width=\textwidth]{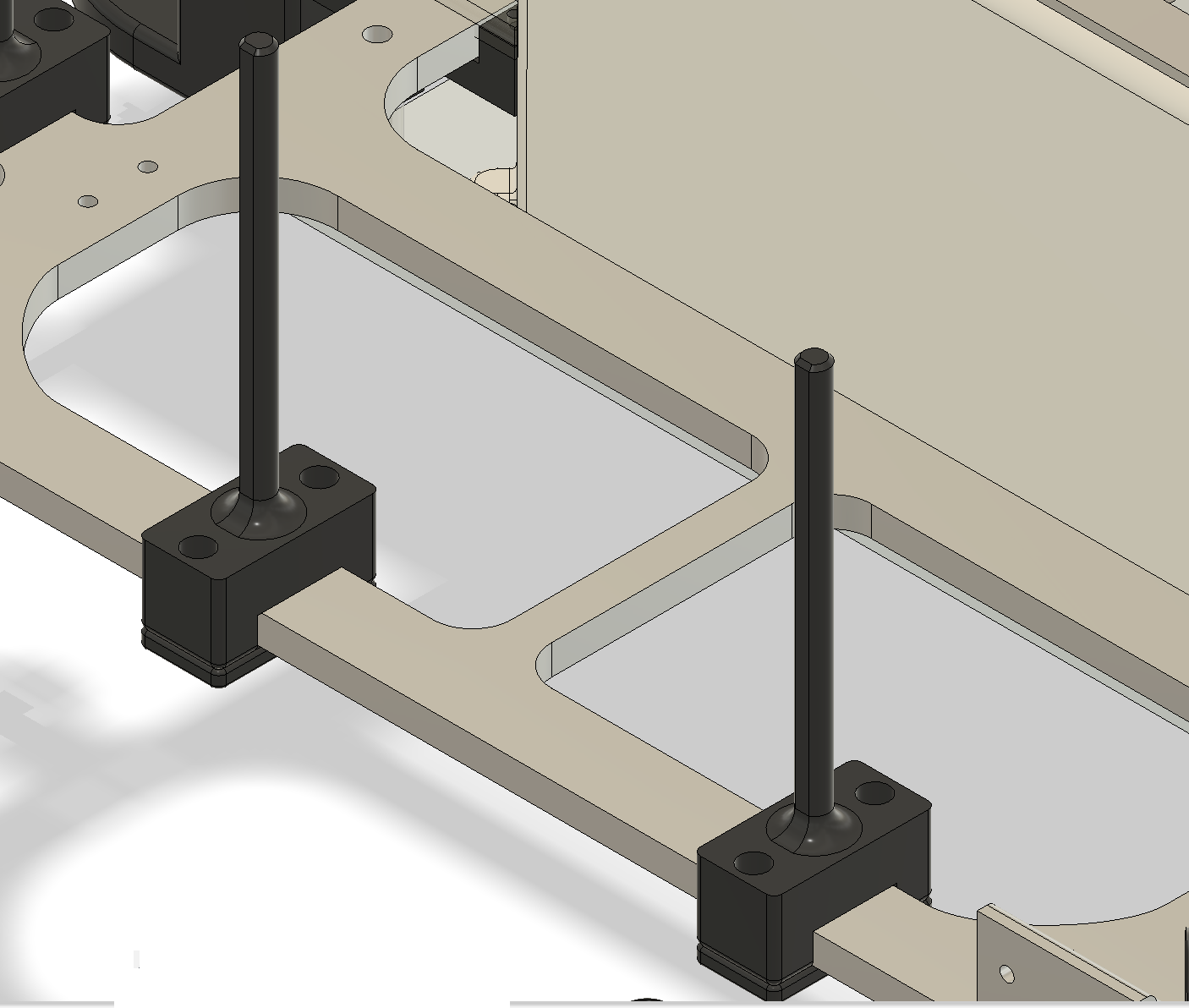}
        \caption{Rod-based Buoyancy System}
        \label{fig:buoyancyrod}
    \end{subfigure}
    \hfill
    \begin{subfigure}[b]{0.45\columnwidth}
        \centering
        \includegraphics[width=\textwidth]{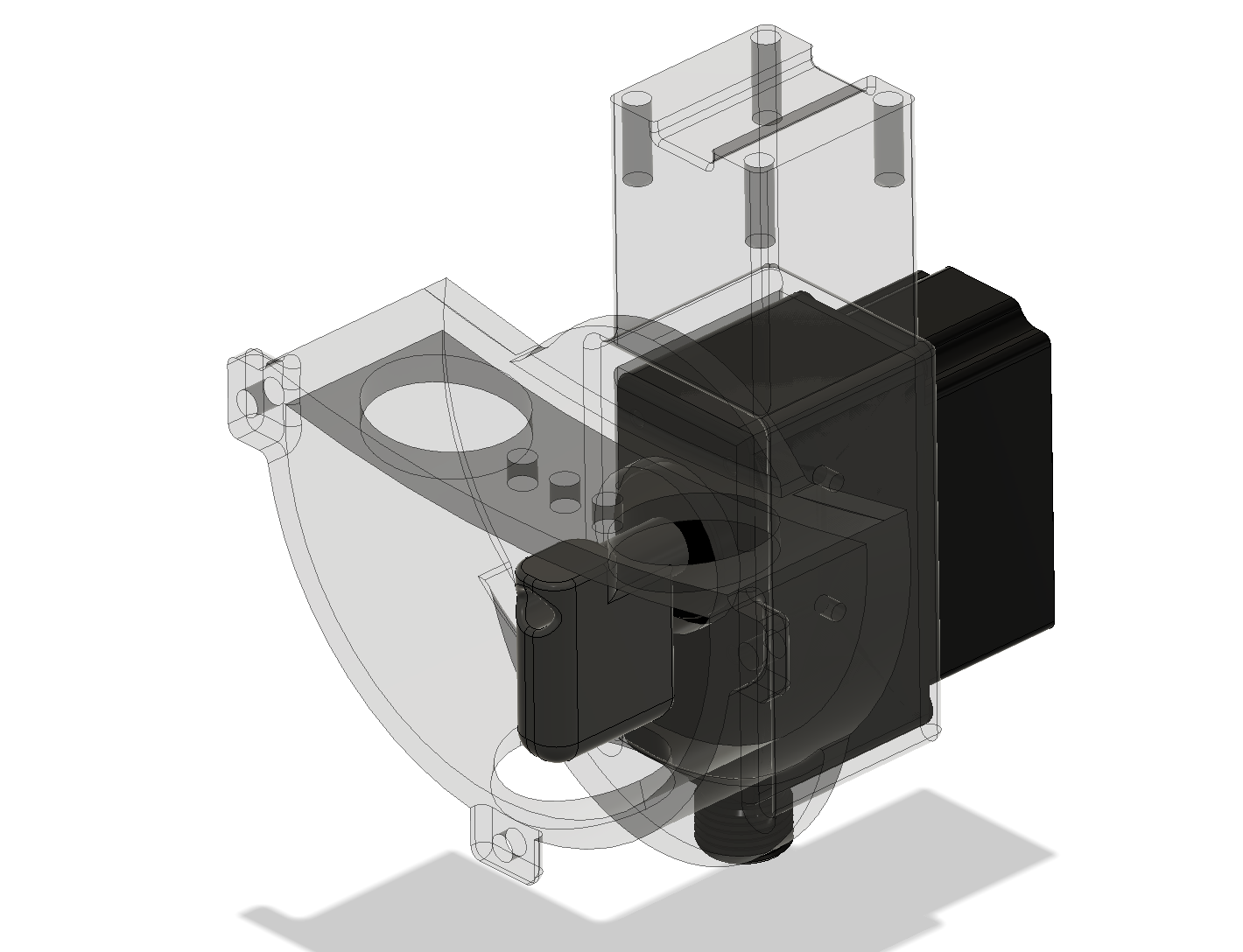}
        \caption{Marker Dropper (body translucent)}
        \label{fig:marker}
    \end{subfigure}
    \caption{New Components Added to Oogway}
    \vspace{-15pt}
    \label{fig:sidebyside}
\end{figure}

We utilized Oogway's modular design to add several new subsystems. The first is our precise adjustable buoyancy system. We transitioned from a rudimentary zip-tie system to a rod-based one, which consists of 3D-printed pillars that the buoyancy blocks slide onto (Fig. \ref{fig:buoyancyrod}). A shaft collar is placed at the top to secure the blocks in place. This mechanism retains the advantages of our previous design--it is light, easy to fabricate, and can be placed anywhere along Oogway's rails. Moreover, the two points of contact hold the buoyancy blocks rigidly in place, preventing changes in buoyancy while the robot moves. 

Another key addition to the robot this year is the marker dropper (Fig. \ref{fig:marker}). It uses a crescent-shaped design that fits two spherical markers, one on each side. The markers naturally roll to the center outlet when the center paddle is rotated by the waterproof servo. There are also several openings at the top of the device: two large holes and three smaller ones. The two larger holes are for quick reloading of the mechanism and are plugged with two rubber stoppers to prevent the balls from falling out during a barrel roll. During testing, we found that air would become trapped in the crescent and slowly leak out, gradually altering the robot's buoyancy throughout a run. The three smaller holes allow air to escape when the robot is submerged. 

\subsection{Electrical Design}

Oogway's new sensing capabilities and ambitious goals for the 2024 RoboSub competition warranted a complete redesign of the electrical stack.

\subsubsection{Electronics Stack Architecture}

\begin{figure}[h!]
\centering
\includegraphics[width=\textwidth]{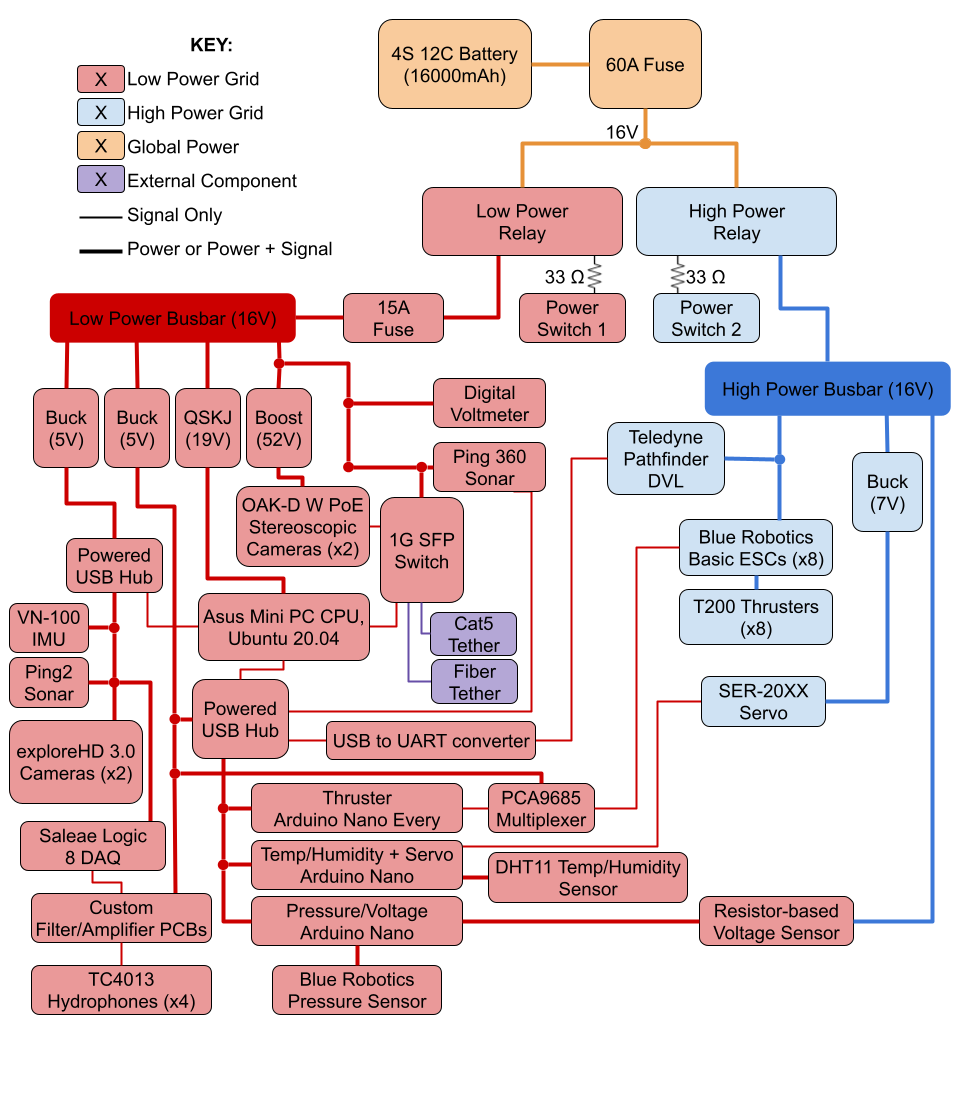}
\caption{Full Electrical System Diagram }
\label{fig:electrical-diagram}
\end{figure}

Our new electrical architecture is based on two independent power grids: one for sensitive low-power and land-safe operations (ex. CPU, cameras), and the other for high draw, underwater-operation-only components (ex. thrusters, DVL). Both grids are fused respective to their own nominal loads, providing added electrical protection to all component failure modes. See Figure \ref{fig:electrical-diagram} for a full system diagram.

The new stack is accessible and adaptable to future components. Expansive busbars and CPU serial busses can accommodate future upgrades. This year alone, the stack has supported three new cameras, an additional sonar, fiber-optic integration, three new sensor types, and two new microcontrollers.

With Oogway's new control system, we found our ESCs susceptible to $\mu$s level inaccuracies in PWM timings. We extensively tested these hardware limitations and devised two means of correction, which we used to advise other RoboSub teams.

We determined that these hardware defects were manufacturing-based and we wrote firmware configurable by what version of the ESC was on the robot. This was utilized when we eventually replaced our ESCs and found that the defects differed. As such, we are adaptable to new and even faulty hardware.

Prior to this year, all of Oogway's sensory components focused on external sensing (e.g., robot acceleration, vision systems, orientation). This year, we have made substantial upgrades to Oogway's internal monitoring, part of which we use to inform our ESC control efforts. Specifically, a new internal voltage sensor is used in dynamically controlling thruster allocations based on available power, using a manufactured-specified and in-house verified thruster power curve~\cite{b12}. Additional internal monitoring includes a capsule-side temperature sensor and a humidity sensor for leak detection.

\subsubsection{Acoustics}
A critical component of the acoustics system includes hand-populated 0603 PC boards that offer bandpass filters and amplifiers for the analog hydrophone signals. The data from these boards is pipelined directly into a DAQ where we calculate and report the azimuth of the pinger to a ROS topic. We formalized findings from prior variable test circuits as a custom PCB, which in total is less than half the size of a credit card. 

\begin{figure}[h!]
\centering
\includegraphics[width=50mm]{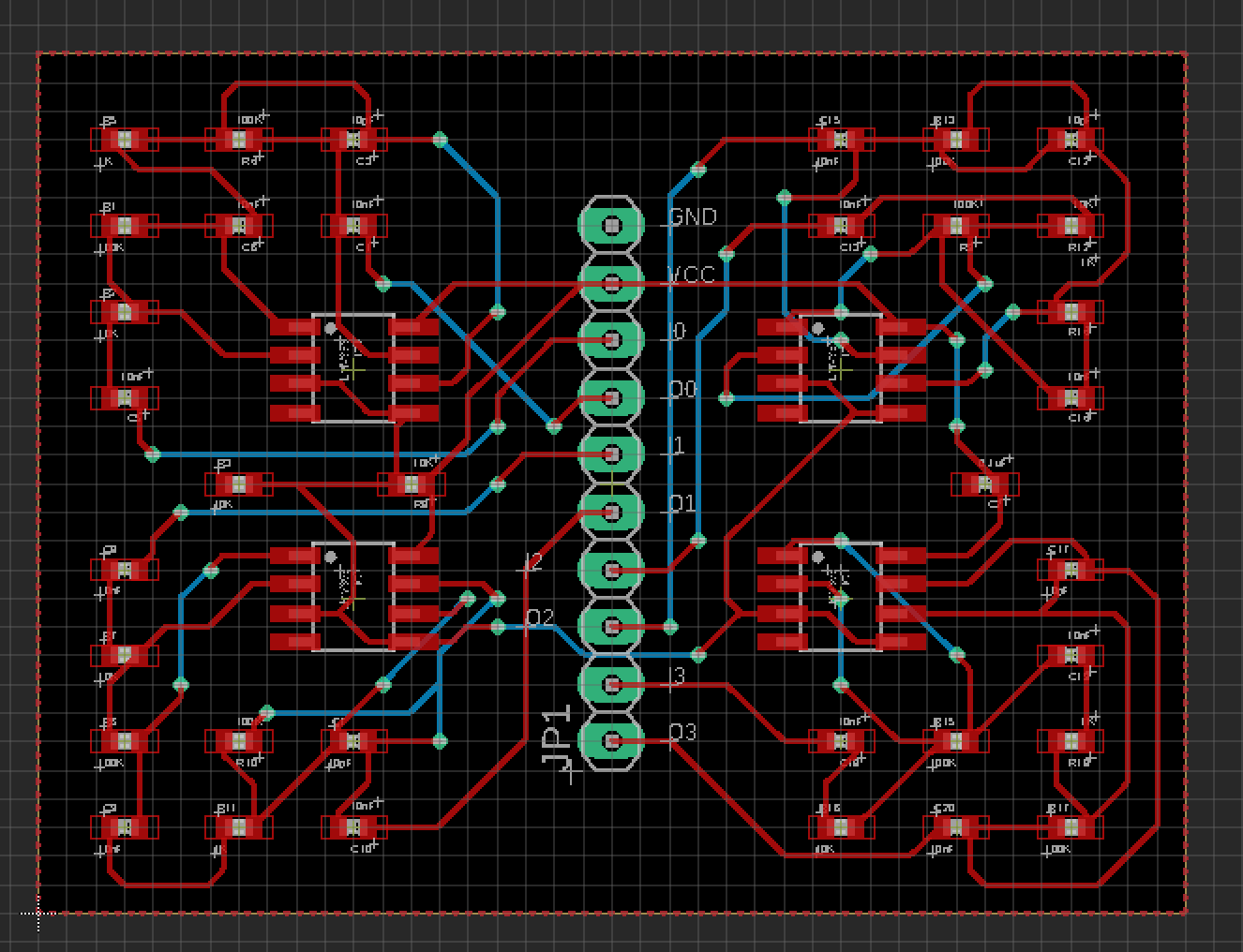}
\caption{Final Acoustic PCB}
\label{fig:pcb}
\vspace{-12pt}
\end{figure}

This hardware was iteratively designed and tested in order to detect pings with maximal accuracy. As part of the testing, we added a high-pass backpack to the boards. The result of our testing yielded filter bands and gain values where the pings were visible in the raw data to the naked eye, even from opposite corners of the lane pool (Fig. \ref{fig:pings}).

Our strategy uses rough guesses collected over time and accurate robot state information to build an aggregate estimate of the absolute position of the pinger. Our hydrophones are located within a few inches of each other to have similar acoustic exposure and use absolute timings to determine the guessed pinger location. Pings are found by first measuring the level of background noise at the desired frequency, and then setting a statistical volume threshold of 4-5 standard deviations from the mean peak levels. Pings can be accurately detected from long distances, with very few false positives. This is done for each hydrophone, using time difference of arrival (TDOA) to approximate the angle at which the source came from~\cite{b1}. We turned this problem into a gradient descent problem as well, classifying the ping timing offsets into 45-degree octants. After verifying feasibility in Python, we wrote an optimized library in Rust.

\subsection{Software Design}
\begin{figure}[h!]
\centering
\includegraphics[width=0.95\textwidth]{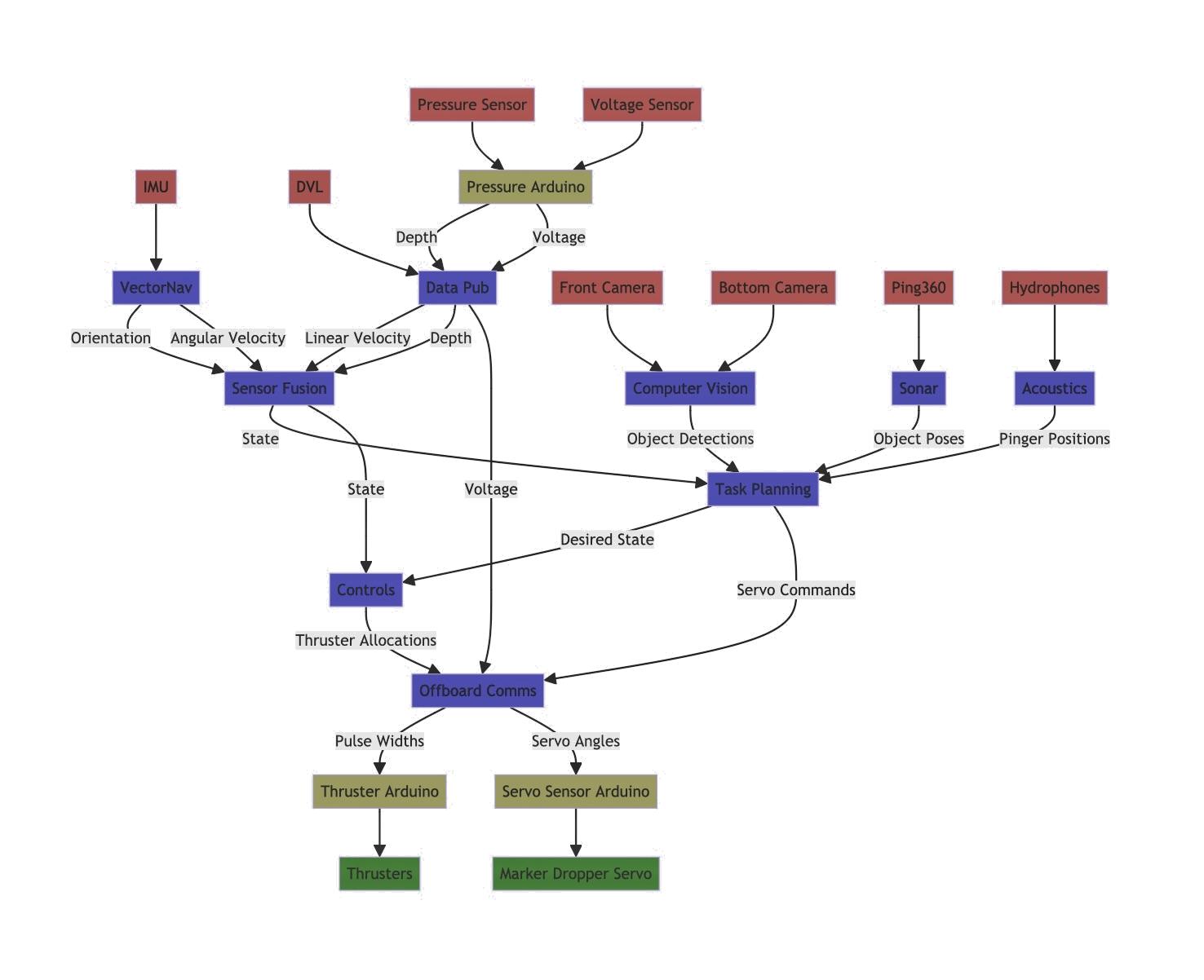}
\caption{System Control Flow Diagram}
\label{fig:flow}
\end{figure}

This year, we have rewritten or refactored every part of our software. This includes controls,  graphical user interface (GUI), computer vision (CV), sonar, and task planning. All of our software is open-source and publicly available on GitHub~\cite{b10}, supporting the AUV community. See Figure \ref{fig:flow} for Oogway's overall software control flow.

\subsubsection{Controls}

\begin{figure}[h!]
\centering
\includegraphics[width=0.95\textwidth]{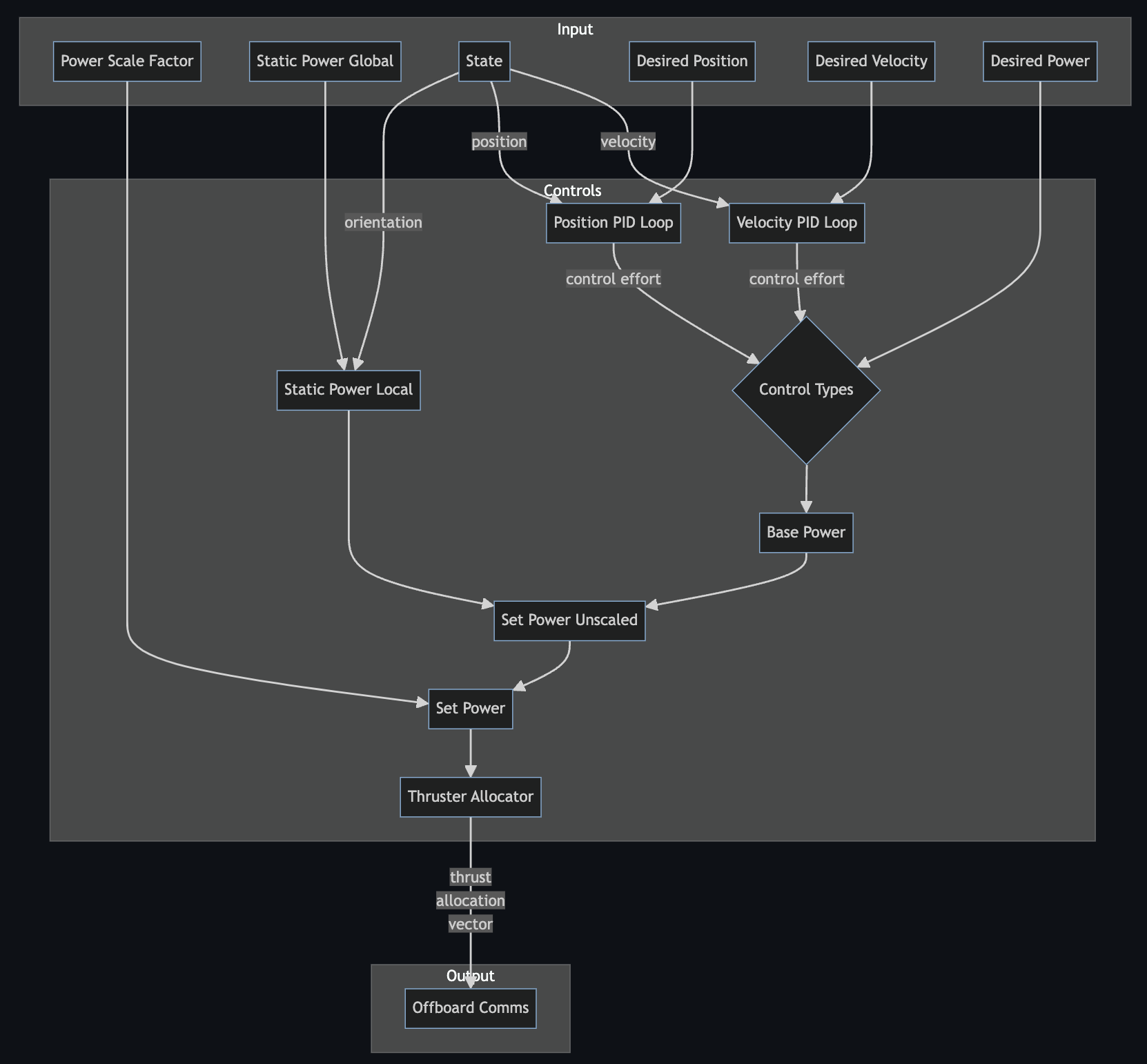}
\caption{New Controls System Diagram}
\label{fig:controls}
\end{figure}

We rewrote our controls system from the ground up this year. We wrote it entirely in C++ for maximum efficiency. (See Figure ~\ref{fig:controls} for a system flowchart.)

The system centers on a set of six Proportional-Integral-Derivative (PID) control loops, one for each axis of movement. We wrote the control loops from scratch, customized for controlling Oogway. They include a second-order Butterworth filter~\cite{b14} to smooth the derivatives for smoother movement. The PID gains and other important parameters can be changed on-the-fly, without re-launching the software, making tuning easy and fast.

The control efforts output by PID are summed with a dynamic offset that neutralizes the robot's buoyancy, enabling it to remain submerged with ease. This is fed into the thrust allocator, which computes the speed each thruster needs to spin at to achieve the desired control efforts without exceeding the thruster power limit. This constrained optimization problem is solved using quadratic programming. 

The system also supports combining multiple control types – position, velocity, and power – for more versatile and precise control over the robot.

\subsubsection{GUI}

This year, we established a new GUI subteam tasked with redesigning our visualization and controls platform using a new, modern interface and architecture. For the past several years, we've used RQt, which is outdated and incapable of holding a real-time connection with Oogway.

In order to facilitate efficient testing, we decided to switch to Foxglove Studio~\cite{b8}, a new robotics visualization platform with plug-and-play ROS integration. Foxglove has a set of built-in extensions (e.g., image and plot panels) which we supplement with custom extensions for our specific use cases. Our new GUI architecture addresses the core limitations of RQt by lowering the bandwidth required by the system, and using TypeScript to develop custom extensions. (See Appendix C).

\subsubsection{Computer Vision (CV)}
This year, the primary focus of our CV subteam was to develop a synthetic data model for machine learning training. In prior years, the process of manually collecting, annotating, and reviewing real images took hundreds of hours of manual labor and was a significant burden. By using Unity Perception~\cite{b9}, we developed a CV simulation which can generate a large labeled dataset of 50,000 images in only a few minutes.

The primary design consideration during the development of the simulation was realism: synthetic images must be captured in a competition-like environment with appropriate randomizations.

\begin{figure}[h!]
    \centering
    \includegraphics[width=0.62\columnwidth]{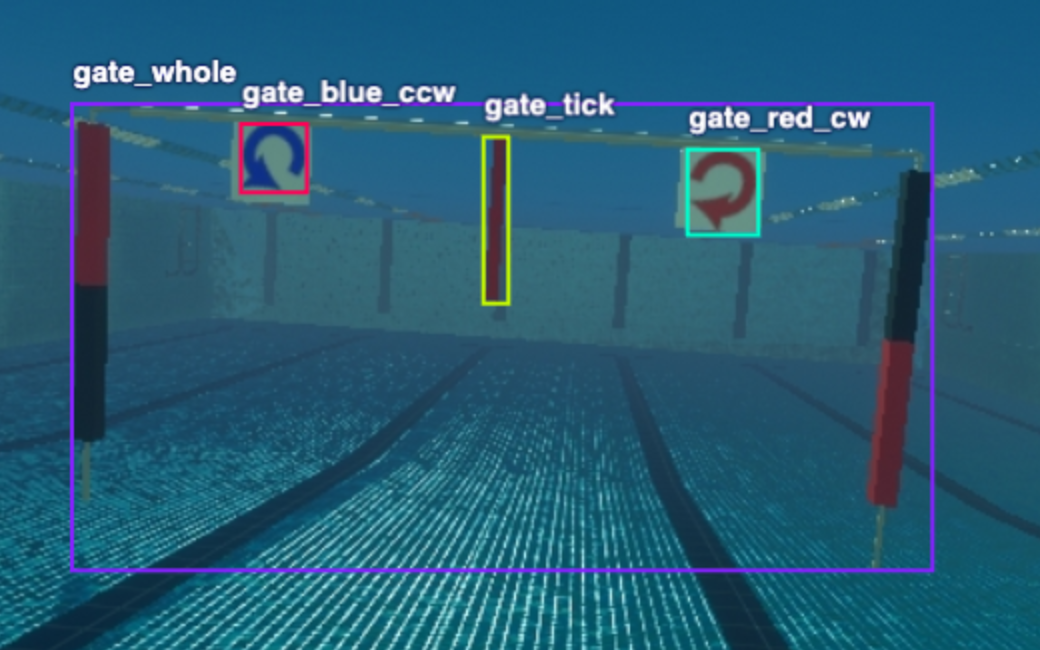}
    \caption{Simulated Gate in 3D Modeled Pool}
    \vspace{-4pt}
    \label{fig:syn_gate}
\end{figure}

These images were then further modified with color, noise, and blur augmentations to create the final model. Like last year, we used the YOLOv7-tiny model~\cite{b2} which can process camera data at 20Hz with high accuracy.

 To detect simple, monochromatic shapes such as the buoy (red circle) and path markers (orange obround), a complex YOLO model would be overkill and incur a lot of computational overhead. Instead, we detect them using  HSV filtering and contour matching. For each camera frame, we mask out the Region of Interest (ROI) of all pixels that are close to the target color in HSV space. We then find the contour of the ROI to classify the object. This proved to be very reliable, as we successfully completed the pre-qualification task using HSV filtering to track the lane markers.

\subsubsection{Sonar}
This year, we rewrote the sonar code to improve its reliability and to have multiple methods of completing tasks. The sonar now avoids false positives caused by acoustic reflections and can detect when the target object is missing.

We overhauled the system with a few important changes. First, we physically mounted the sonar directly below the camera to better align with CV data. Second, we apply a filter to get rid of all low-intensity values and use the DBSCAN~\cite{b15} algorithm to find all clusters in the image. We select the biggest cluster, which represents the buoy, and find its center of mass. In our testing, this method proved to be reliable and ensures the sonar  only recognizes one object. It also provides the the normal angle to the buoy which helps to align Oogway in the correct orientation.

\vspace{-5pt}

\begin{figure}[h!]
\centering
\includegraphics[width=85mm]{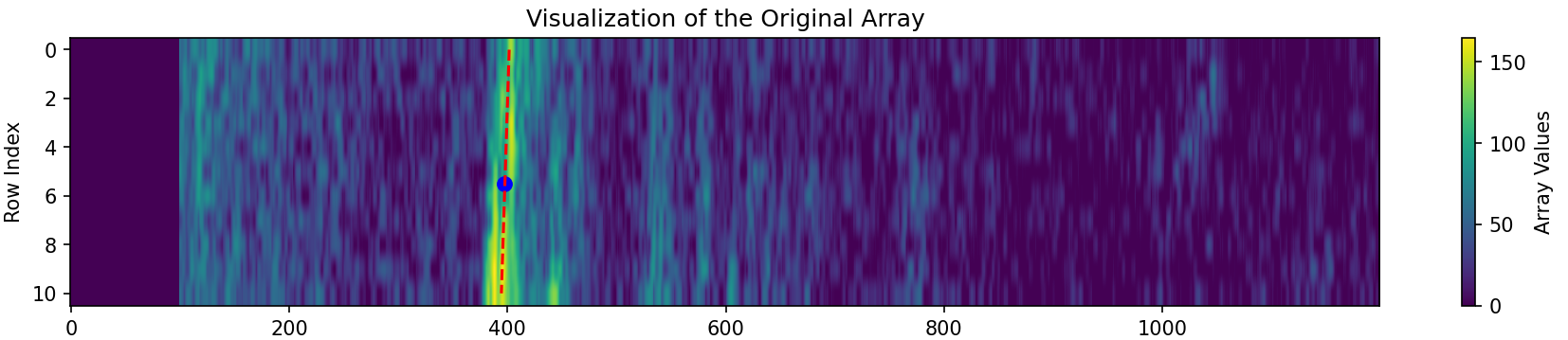}
\caption{Buoy Detected with Normal Line}
\vspace{-5pt}

\label{fig:sonar_scan}
\end{figure}

We also implemented a new system to detect and identify objects within long range sonar scans. This serves as a fallback method for multiple tasks by finding a rough position of all objects in the pool. We find clusters in the image with DBSCAN and filter based on area and circularity. Using these two measurements, we classify objects into \textit{gate}, \textit{buoy}, and \textit{unknown}. We avoided false positives from buoy reflections by filtering out neighboring clusters with similar size and circularity ~\cite{b3}. See Appendix E for the full sonar pipeline.

\subsubsection{Task Planning}

The task planning system receives information from all the robot's sensors, decides the best action to complete the task, and commands the controls system to carry out that action. Each task is written using a Python coroutine. Thus, each child task frequently yields control to its parent task, allowing the parent task to frequently reevaluate the robot's environment and switch between different task pathways. The system is also very modular; small actions are combined to accomplish complex tasks. By adding a simple Python decorator, each task is automatically included in the real-time task monitoring system.

\section{Testing Strategy}

\subsection{Test Plan}
The most important lesson from last year was that testing is the key to success. Ensuring every component and subsystem is reliable will ensure a successful competition run. Oogway's testing suite comes from a combination of simulated and real-world environments. Our test plan is as follows:
\begin{enumerate}[leftmargin=*]
    \item Test all compatible components in simulation. This ensures that each component works properly and connects to the robot without negatively impacting other systems.
    \item Test each component physically on the robot to ensure all components fit properly. We run component specific tests in the pool and in the sink when possible.
    \item Once all components work separately, test the integration of components in the pool to mimic a real competition environment. Pool testing allows us to test the robot on completion tasks and ensure we could complete each task reliably.
\end{enumerate}

Following this strategy ensured that every component was individually reliable and the robot was able to complete all \textit{core tasks}.

\subsection{Simulated Testing}

\subsubsection{Acoustics}
While developing the algorithm underlying the acoustics stack, we created a simulated acoustics program. This involved simulating a pinger source and calculating the induced pressure waves. This, combined with encoded hydrophone geometries and artificial noise, provided realistic data to test the algorithm with.

\subsubsection{Software}
We utilized our custom sim environment in CoppeliaSim--which includes models of Oogway, the gate,  buoys, and the octagon--to test controls and task algorithms. Before empirical testing, we ran each of our \textit{core tasks} in the sim to ensure our code completed the task correctly.

\subsubsection{CV}
As described above, we use Unity to generate images and test our CV model.

\vspace{-5pt}

\subsection{Empirical Testing}
\subsubsection{Mechanical}
All water-facing parts underwent extensive functionality and water-resistance testing. This often involved long-duration submersion tests in a sink. Once passed, the testing would escalate to a diving pool. This comprehensive approach guarantees all vital components will remain watertight during operation. Non-water-facing components go through the standard prototyping process, which involves testing and iteration on dry land.

\subsubsection{Electrical}
Each electrical component was initially ran for extended periods of time to test long-term performance. All subcon and epoxy connections, as well as waterproof components, underwent 24 hour submersion testing. Once all of the components were tested individually, we performed full system integration tests to ensure accurate data.

\subsubsection{Software}
As mentioned above, we tested our new controls and task-planning systems through our Foxglove GUI which allows us to set PID constants and observe setpoints. We visualized the task flow of the robot to verify the robot's state and actions.

\subsubsection{CV}
To test the object detection algorithm, we set up the gate and manually moved the robot around to test all vision angles. We also tested CV-sonar integration, by matching the two data streams. We then tested movement autonomously, verifying the reliability of our gate task.

\subsection{Testing Results and Integration Testing}

Towards the end of our season, we tested the robot with each of our \textit{core tasks} to verify that we could compete in a competition environment.

In total, we spent over 500 hours in simulated and empirical testing environments, resulting in Oogway prequalifing for the 2024 RoboSub competition.

\section{Acknowledgement}

Duke Robotics Club is primarily sponsored by Duke University’s Pratt School of Engineering. We are grateful to Ali Stocks and the Foundry staff for housing and supporting us. We are also indebted to Director of Undergraduate Student Affairs, Tarina Argese, for helping us plan events, our advisors, Professor Michael Zavlanos and Dr. Boyuan Chen, for technical assistance, and the Engineering Alumni Council, for overall club advice. Furthermore, we owe our success to our longtime sponsors: The Lord Foundation, Duke Student Government, General Motors, and SolidWorks. Finally, we thank RoboNation, whose commitment to RoboSub empowers student engineers like ourselves to explore our passion for robotics.

\onecolumn
\clearpage

\section*{Appendix A: Component List}

\begin{table}[H]
\centering
\begin{tblr}{
  width = \linewidth,
  colspec = {Q[198]Q[96]Q[185]Q[137]Q[135]Q[69]Q[115]},
}
\hline
\textbf{Component}               & \textbf{Vendor} & \textbf{Model/Type}          & \textbf{Specs}          & \textbf{Custom/Purchased} & \textbf{Cost} & \textbf{Year of Purchase} \\ \hline
Buoyancy Control                 & Blue Robotics   & Subsea Buoyancy Foam         & R-3312                  & Purchased                 & 200.00        & 2023                     \\
Frame                            &                 & Aluminum                     & 821                     & Custom                    & 2,000.00      & 2023                     \\
Waterproof Housing               &                 & Polycarbonate                & -                       & Custom                    & 345.00        & 2023                     \\
Waterproof Casing                & Blue Robotics   & Watertight Enclosure         & 4" Cast Acrylic Plastic & Purchased                 & 124.00        & 2023                     \\
Waterproof Connectors            & MacArtney       & Subconn/Seaconn Connectors   & -                       & Purchased                 & 1600.00       & 2019                     \\
Waterproof Connectors            & Blue Robotics   & WetLink Penetrators          & 4.5MM-HC                & Purchased                 & 400.00        & 2023                     \\
Propulsion                       & Blue Robotics   & T200 Thruster                & -                       & Purchased                 & 1,790.00      & 2023                     \\
Power System                     & Turnigy         & 16000mAh                     & 4S 12C                  & Purchased                 & 210.00        & 2023                     \\
Motor Controls                   & Blue Robotics   & Basic ESC                    & -~                      & Purchased                 & 300.00        & 2023                     \\
High Level Control               & Arduino         & Arduino Nano Every (x2), Arduino Nano (x1)           & -                       & Purchased                 & 40.50         & 2023                     \\
CPU                              & ASUS            & MiniPC PN80                  & -                       & Purchased                 & 1,300.00      & 2022                     \\
Internal Fiber Comm Network            & Mini Industrial         & 4 Rj45 + 1 SFP Switch                & -                       & Purchased                 & 56.00         & 2024                     \\
External Comm Interface          & NetGear         & 8-Port 1G/10G Gigabit Ethernet Unmanaged SFP Switch              & -                       & Purchased                 & 100.00        & 2024                     \\
Intertial Measurement
Unit (IMU) & VectorNav       & VN-100 Rugged IMU            & -                       & Purchased                 & 1,335.00      & 2020                     \\
Doppler Velocity Logger
(DVL)    & Teledyne        & Pathfinder 600               & -                       & Purchased                 & 10,000.00              & 2023                     \\
Stereo Cameras (x2)                        & Luxonis         & OpenCV AI Kit: OAK-D W PoE     & -                       & Purchased                 & 1,000.00           & 2023                     \\
Hydrophones                      & Aquarian        & H1a Hydrophones              & -                       & Purchased                    & 600.00        & 2021 \\

Imaging Sonar                & Blue Robotics   & Ping360 Scanning Imaging Sonar  & -                 & Purchased                 & 2,650.00        & 2023 \\

Waterproof Servo                 & Blue Trail Engineering   & SER-2020         & -                 & Purchased                 & 495.00        & 2024 \\

Mono Camera (x2)                 & DWE   & exploreHD 2.0 
 & -                 & Purchased                 & 630.00        & 2023 \\

Single-Beam Echosounder                & Blue Robotics   & Ping2 Sonar  & -                 & Purchased                 & 435.00        & 2024\\ \hline
Vision~ ~                        & OpenCV, DepthAI          &                              &                         &                           &               &                          \\
Localization Mapping         &                 & Extended Kalman Filter, SLAM &                         & Custom                    &               &                          \\
Autonomy                         &                 & Custom task planner          &                         & Custom                    &               &                          \\ \hline
Open-Source Software             &                 & Docker, ROS, Ubuntu               &                         &                           &               &                          \\
Programming Language(s)          &                 & C++, Python, Rust             &                         &                           &               &     
\end{tblr}
\end{table}

\clearpage
\twocolumn

\section*{Appendix B: Test Plan Results}

\subsection{Mechanical}
Mechanical testing is arguably the most critical aspect of verifying Oogway's performance. A failure here would disable all other subsystems. As described previously, water-facing components were put through multiple rounds of testing.

\subsubsection{Waterproofing the Stack} 
\begin{figure}[h!]
\centering
\includegraphics[width=69mm]{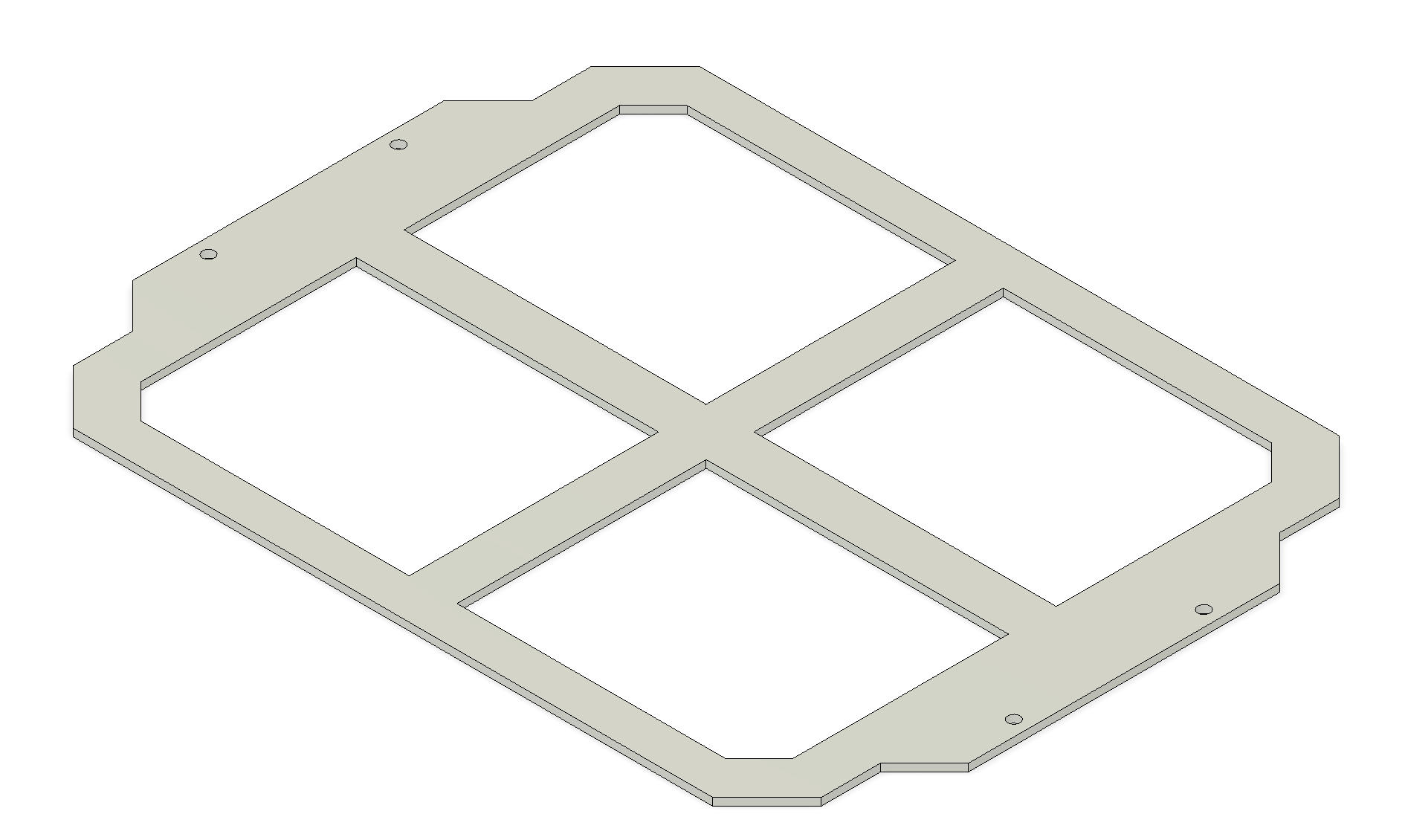}
\caption{First Plate Prototype}
\label{fig:firstplate}
\end{figure}

The titanium plate discussed in the body, for instance, underwent several prototypes, each subject to rigorous testing. The goal of the first prototype was to prove the feasibility of the concept and used leftover 0.125" 6061 aluminum stock (Fig. \ref{fig:firstplate}). Testing on dry land showed that the thin aluminum was far too flexible and bent severely while under pressure. Part of this was also due to the open window design, which future designs improved on. 

\begin{figure}[h!]
\centering
\includegraphics[width=69mm]{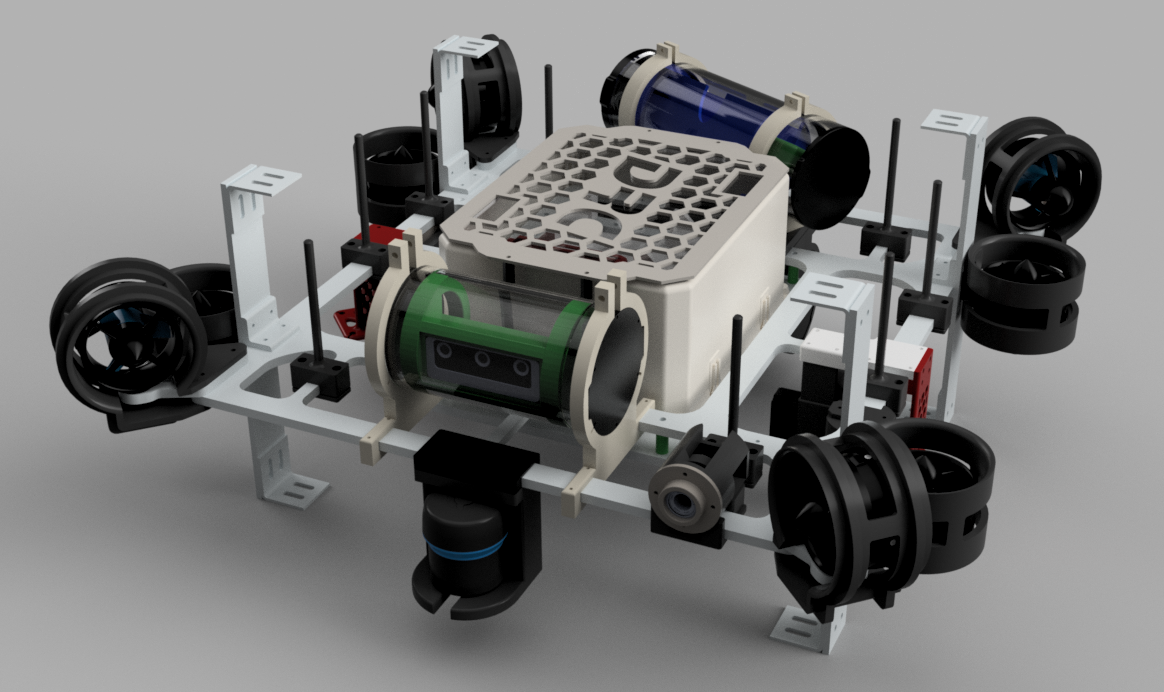}
\caption{Full Assembly Render of Oogway}
\label{fig:render}
\vspace{-0.5cm}
\end{figure}

\begin{figure}[h!]
\centering
\includegraphics[width=65mm]{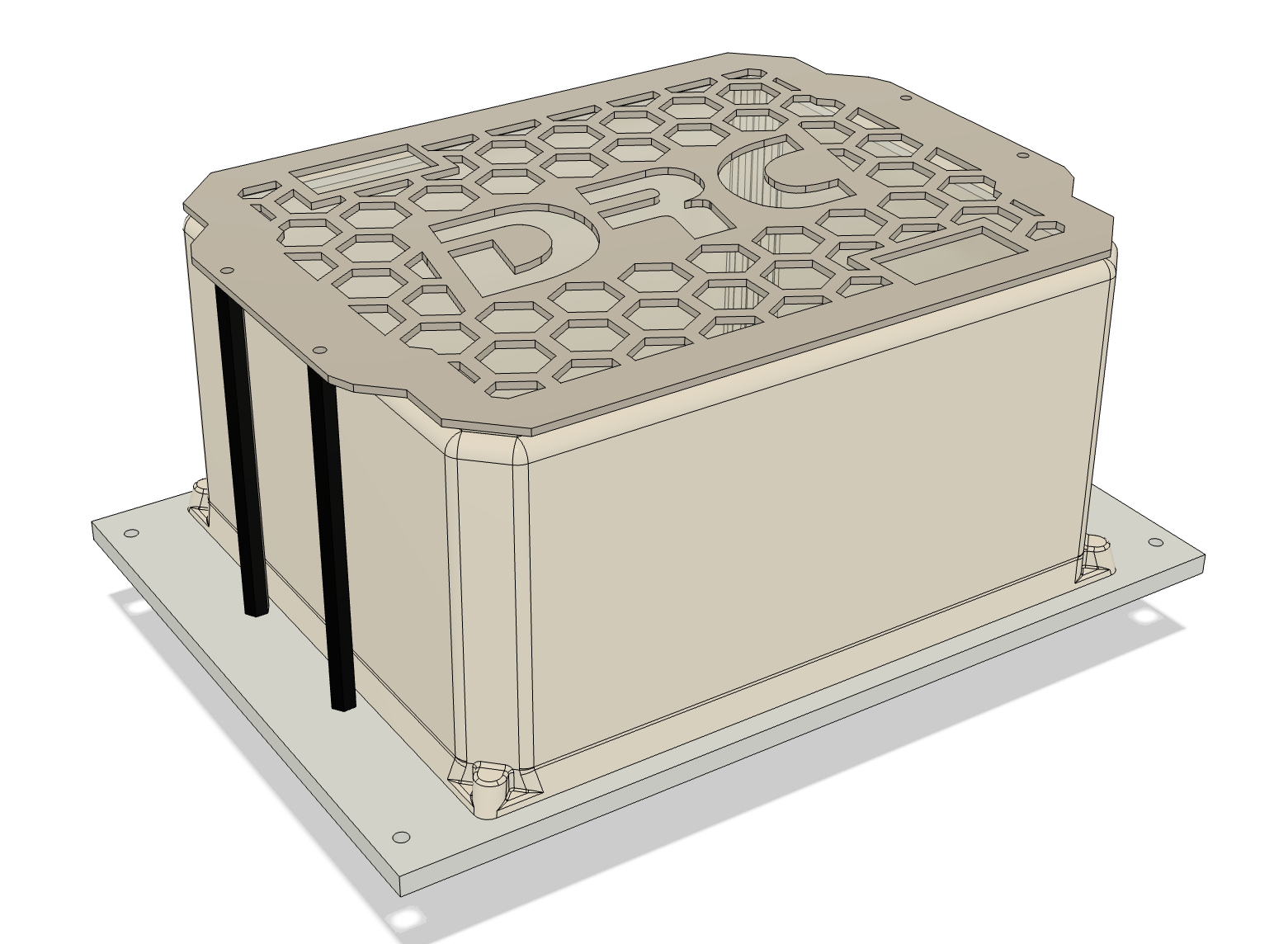}
\caption{New Capsule Plate Design}
\label{fig:newplatedesign}
\end{figure}

\begin{figure}[h!]
\centering
\includegraphics[width=69mm]{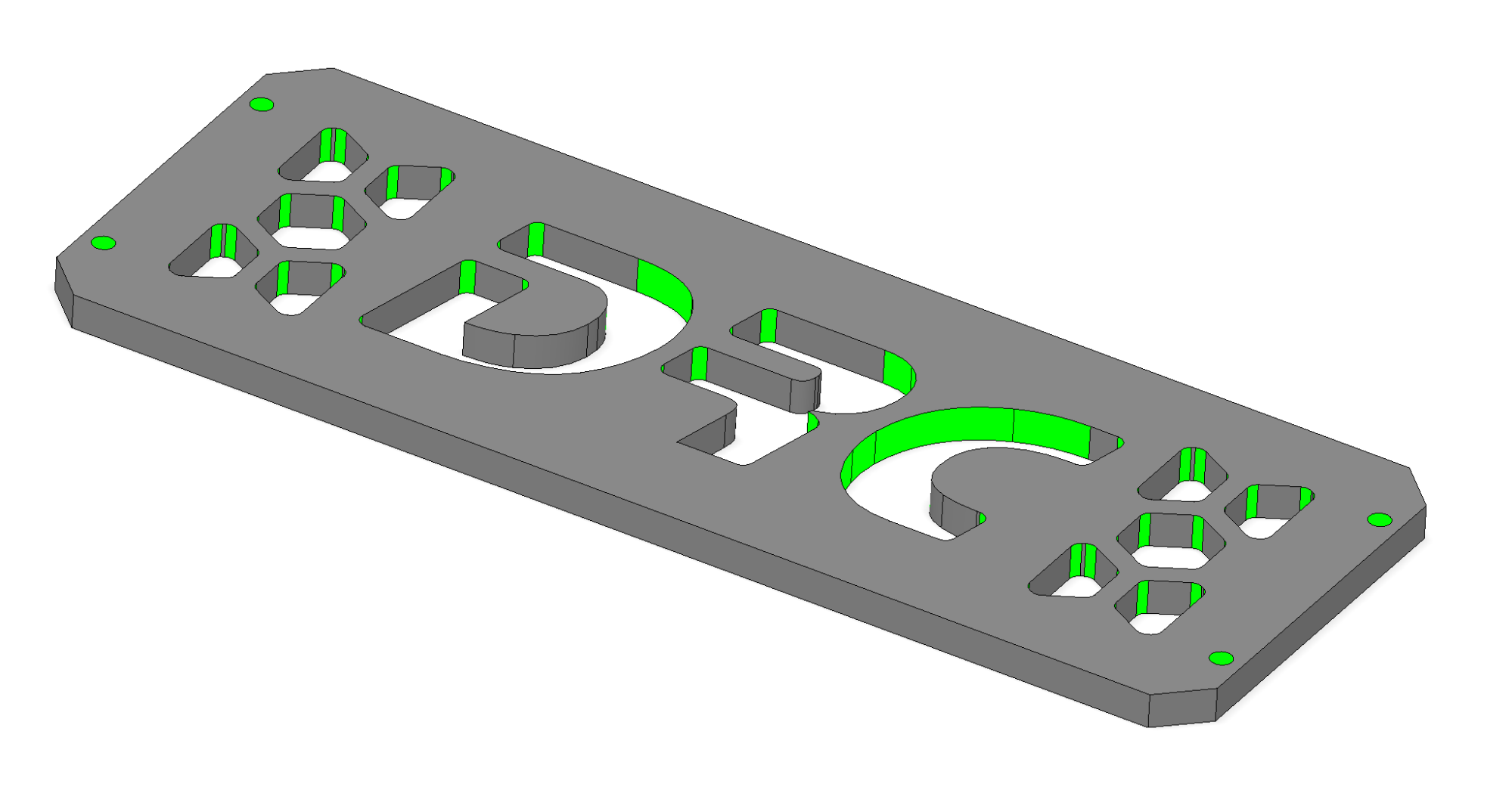}
\caption{Second Plate Prototype}
\label{fig:secondplate}
\end{figure}

The second prototype used 0.4" 6061 aluminum stock, also leftover from a separate project (Fig. \ref{fig:secondplate}). To compensate for the weight added by the increased thickness, the plate only stretched across the center. This prototype performed much better than the previous: there was no bend in the aluminum when fully tightened, but it did weigh significantly more. Moreover, because it only covered the center, pressure was not applied evenly across the entire o-ring. \\
Even so, we were able to trial this design in the pool with some added precautions. In addition to screwing down the four screws that keep the plate in place, we also screwed in the four corner screws of the capsule as extra support. We also put the robot in slowly and monitored closely for bubbles.

\begin{figure}[h!]
\centering
\includegraphics[width=69mm]{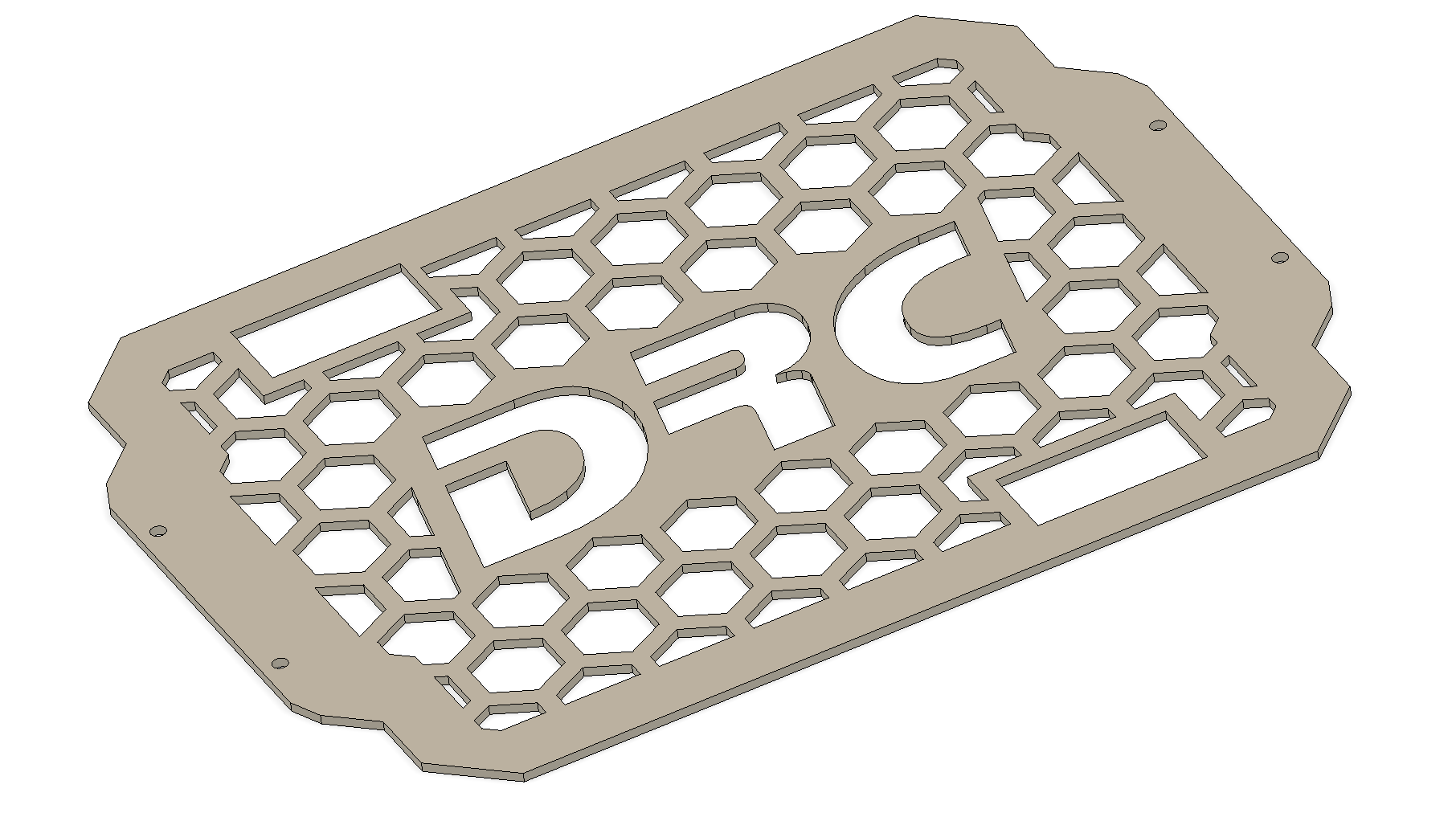}
\caption{Final Plate Design}
\label{fig:finalplate}
\end{figure}

The final prototype combined the benefits from both of our previous prototypes: the design stretches across the entire capsule to provide even pressure, and the chosen material, Grade V titanium, is both light and rigid (Fig. \ref{fig:finalplate}) \cite{b4}. 

This new titanium plate design initially showed a substantial increase in the watertightness of the capsule's seal. While we could not trial this design in the sink, we took steps to ensure that the electrical stack would not be damaged during the first few tests. Like with the previous prototype, this involved the use of the four corner screws in addition to the four screws used to secure the top plate. Later, a humidity sensor was added to the bottom of the stack such that leaks could be recognized instantly. While this solution has performed much more consistently than the previous one, some minor leaks still occasionally occur, likely due to wear-and-tear on the capsule, wear-and-tear on the o-ring, or loose ports in the stack plate. However, the humidity sensor allows these leaks to be identified before they become catastrophic. 

There are still areas for improvement. As discussed in the body, moving to more standard parts or utilizing in-house fabrication would lessen our reliance on distributors and allow us to replace worn parts easily. 

\begin{figure}[h!]
\centering
\includegraphics[width=69mm]{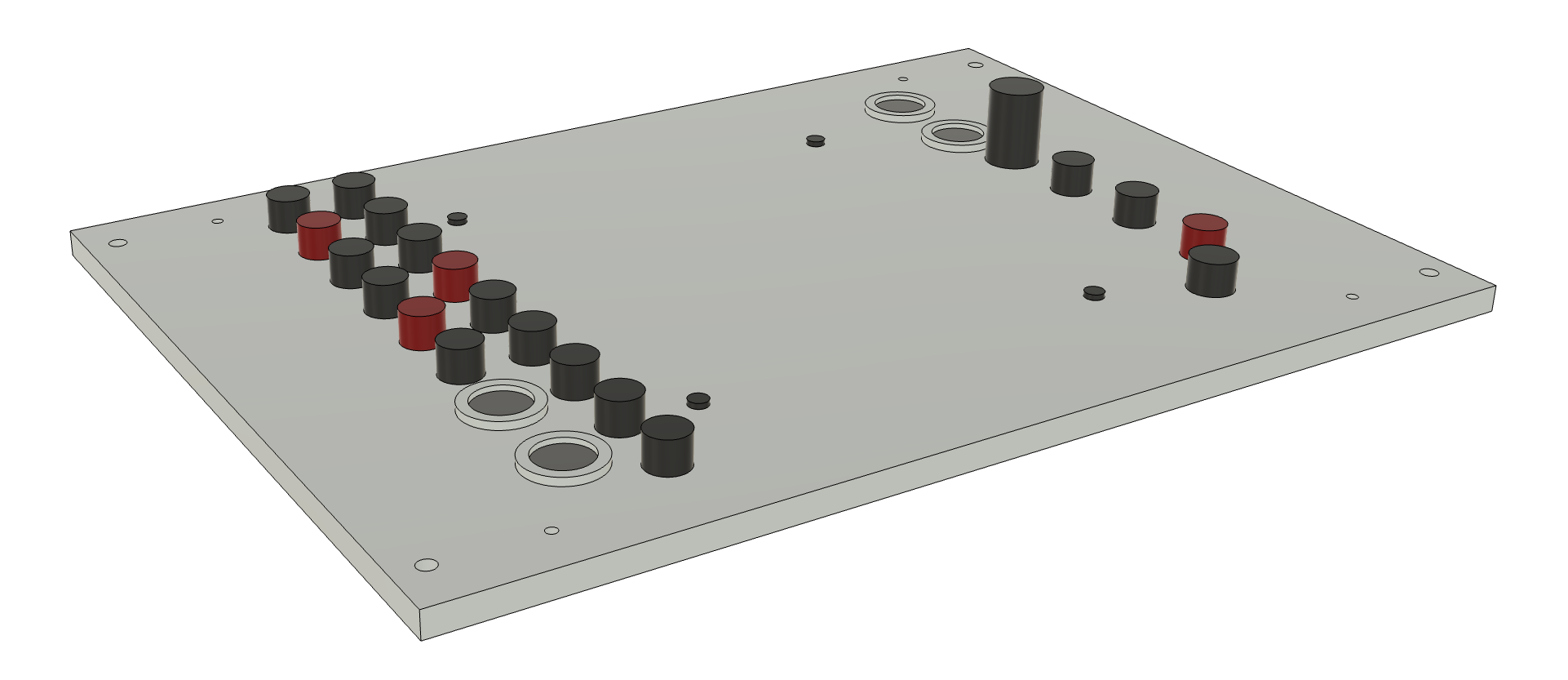}
\caption{Aluminium base plate built for adaptability}
\label{fig:baseplate}
\end{figure}

The aluminum plate at the bottom of the stack likewise went through rigorous waterproof testing. It was built to incorporate many sensors and electrical attachments and contains extra holes to be adaptable to additional sensors that need to be attached (Fig. \ref{fig:baseplate}). We tested the watertightness of the plate using blanks for all the plugs and screwing in a capsule on the four corners. Dye packets were placed inside the capsule and it was then submerged in the sink for 24 hours to stress test the system. No leaks were detected. Moreover, we anodized the plate following a professional guide to prevent corrosion \cite{b6}. 

\subsubsection{Waterproofing 3D prints} Due to the large number of 3D printed parts on the robot, one problem we noticed during pool tests was that our PLA 3D prints would fill with water \cite{b7}. Although usually not a problem with PLA plastic, the low infill density (15\%) and gaps between the layer lines meant water could easily fill the large cavities within the print. This resulted in the robot losing its finely tuned buoyancy gradually, making robot movement impossible to correctly tune. \\

To fix this, we tried several different solutions that either provided a seal around the outer layer of the 3D print or fused the layer lines. For each of the solutions, we took data at regular intervals over an hour. The results are given in Table \ref{tab:data}. 

\begin{table}[ht]
    \centering
    \caption{3D Print Water Absorption Data}
    \begin{tabular}{p{28pt}p{23pt}p{23pt}p{23pt}p{23pt}p{23pt}p{23pt}}
        \toprule
        Time & Control White & Control Black & Wood Glue & Clear Nail & Blue Nail & Clear Sealer Gloss \\
        \midrule
        0   & 66 & 56 & 72 & 62 & 28 & 66 \\
        5   & 68 & 58 & 74 & 62 & 28 & 68 \\
        10  & 68 & 58 & 76 & 62 & 28 & 70 \\
        15  & 68 & 58 & 74 & 62 & 28 & 68 \\
        20  & 68 & 60 & 76 & 62 & 30 & 70 \\
        25  & 68 & 60 & 76 & 62 & 30 & 68 \\
        30  & 68 & 58 & 74 & 62 & 28 & 68 \\
        45  & 68 & 58 & 74 & 62 & 28 & 68 \\
        60  & 68 & 58 & 76 & 62 & 30 & 68 \\
        \midrule
        \% gain & 3.03 & 3.57 & 5.56 & 0 & 7.14 & 3.03 \\
        \bottomrule
    \end{tabular}
    \label{tab:data}
\end{table}

The clear nail polish performed best: the acetone in the nail polish fuses the layer lines and can be thoroughly applied. It was unclear why the blue nail polish performed much worse, but may have to do with the dye used or variation in the quality of the print that it coated. 

During pool tests, the nail-polished prints have been successful in minimizing water intake. In previous pool tests, bubbles were noticeable coming out of 3D prints as the robot was dropped in water, indicating the print filling with water, but this no longer occurs. Moreover, buoyancy has been found to remain stable across the entire period. 

\subsubsection{Buoyancy} One of the main priorities for the mechanical team is to ensure that the robot's buoyancy is level and can be easily adjusted as new parts are added. Thus, we tested different buoyancy mount designs before settling on the final design. One such design was the capsule-based system pictured below. 

\begin{figure}[h!]
\centering
\includegraphics[width=69mm]{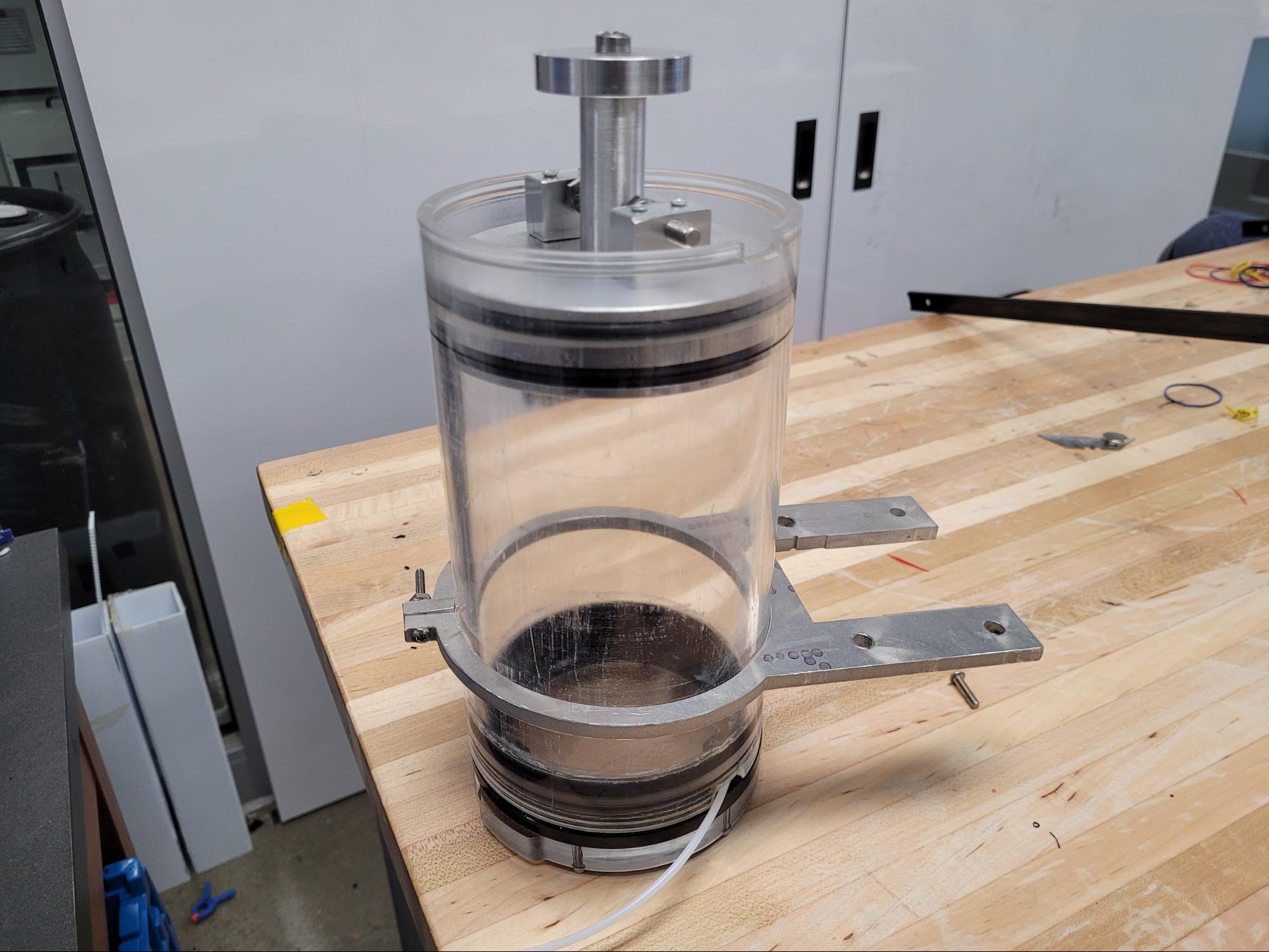}
\caption{Buoyancy Piston Prototype}
\label{fig:piston}
\end{figure}

We decided to retrofit an old, damaged acrylic capsule into a piston-like mechanism that allows its volume to be adjusted when needed, providing higher granularity than the discrete buoyancy blocks we previously relied on (Fig. \ref{fig:piston}). Moreover, a future design could leverage actuators to automatically compress and depress the piston based on a feedback loop. 

However, this system posed a few disadvantages over the old design. It was much larger and heavier than the simple 3D-printed mounts, meaning there were only a few spots where it could be mounted on the robot. While having high granularity was important, testing showed that positional flexibility was much more important in ensuring a neutrally buoyant robot. Therefore, we moved to the rod-based system discussed in the body.

\subsubsection{Mounting Brackets} 
\begin{figure}[h!]
\centering
\includegraphics[width=69mm]{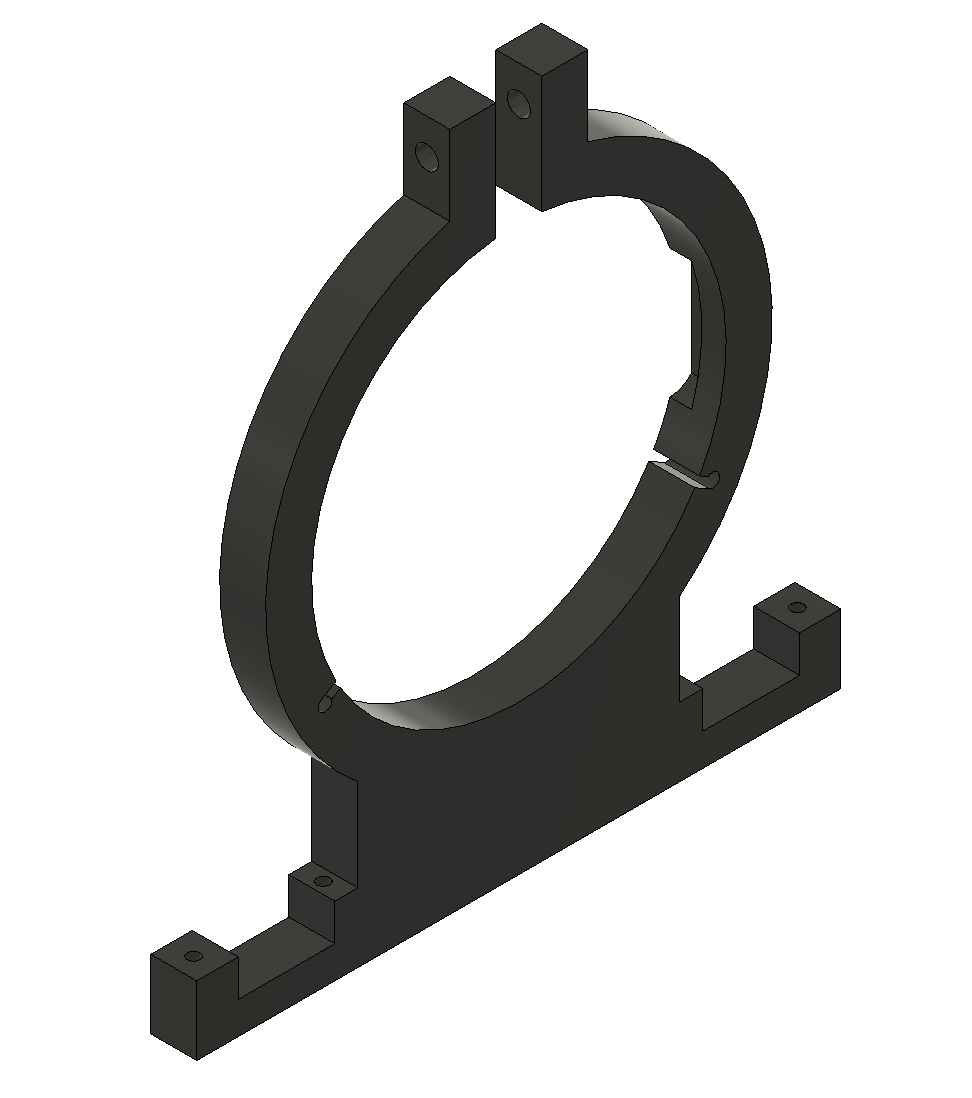}
\caption{New Unibody Camera Capsule Mount}
\label{fig:camera}
\end{figure}

\begin{figure}[h!]
\centering
\includegraphics[width=69mm]{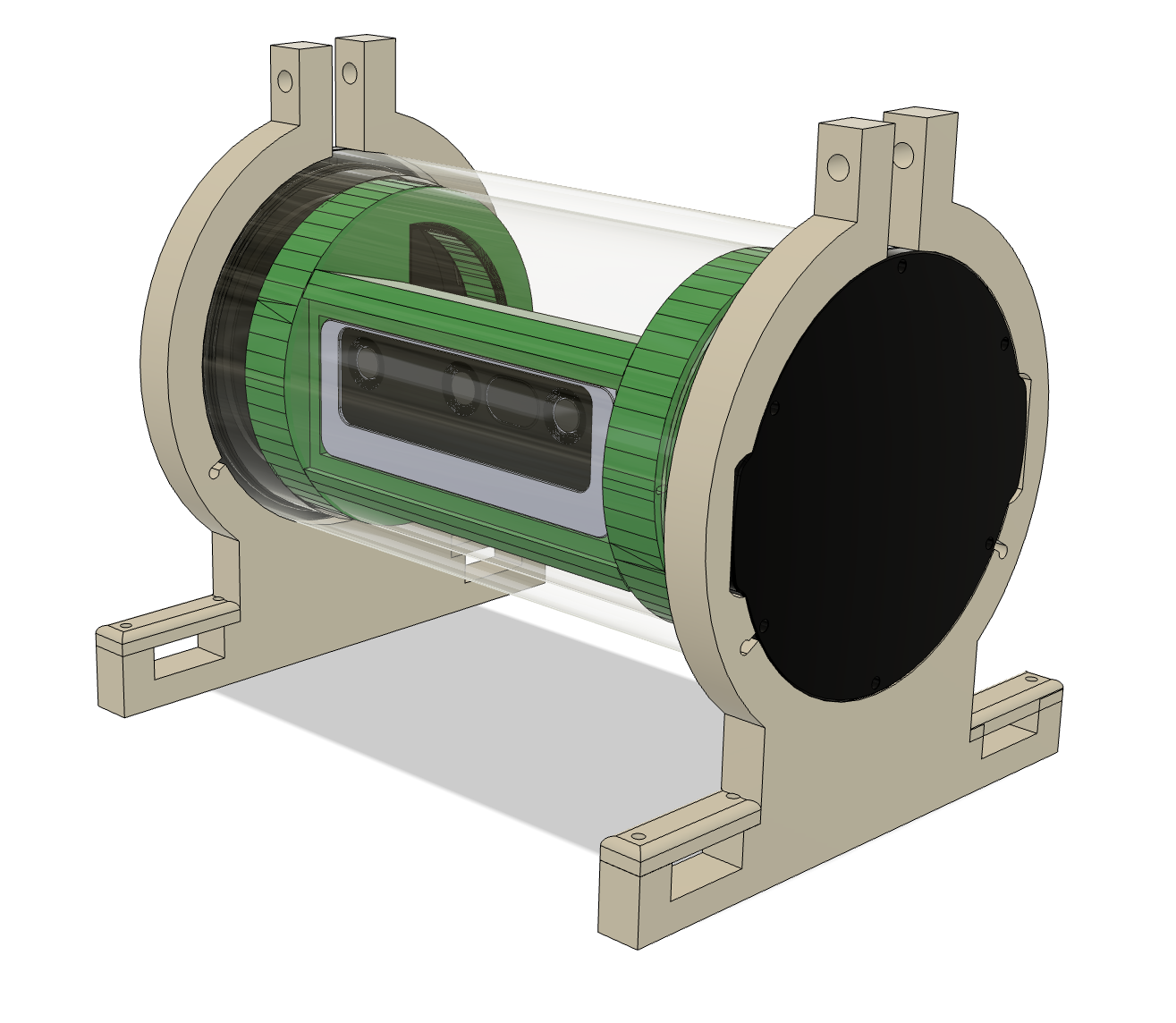}
\caption{Camera Capsule Subassembly}
\label{fig:camerasubassm}
\end{figure}

This year also saw the update of all of our old mounting solutions. During the competition last year, several 3D-printed mounts failed under stress. However, we quickly identified the issue for the camera capsule and battery capsule mounts: the two-part mechanism was weak at the interface, causing the plastic to strip out under pressure. Therefore, we adopted a unibody mount for the camera capsule, eliminating the weak point (Fig. \ref{fig:camera}, Fig. \ref{fig:camerasubassm}). This system was stress-tested during pool tests and we noted no failures this year. 


\begin{figure}[h!]
\centering
\includegraphics[width=69mm]{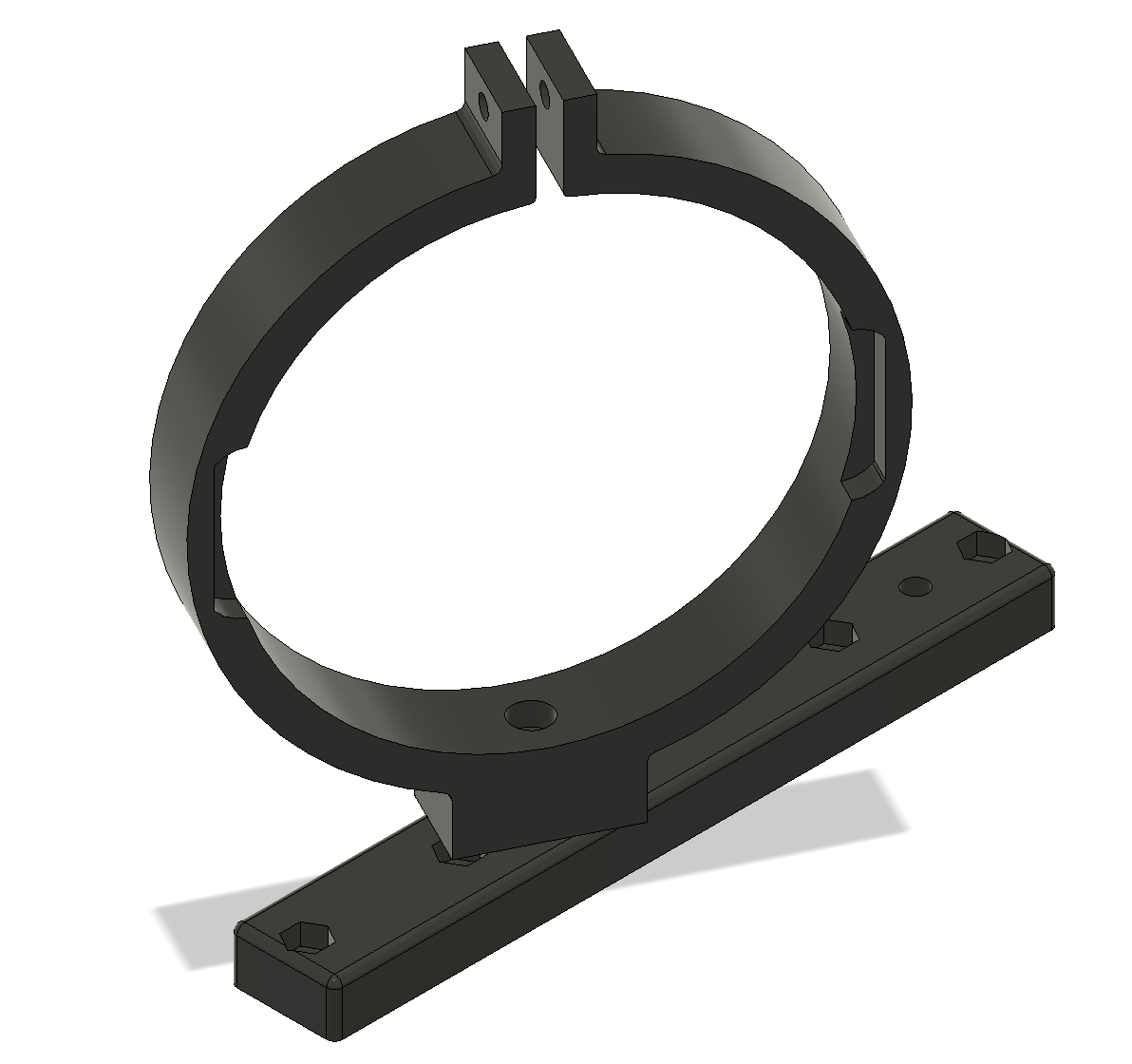}
\caption{New Battery Capsule Mount}
\label{fig:bcm}
\end{figure}

\begin{figure}[h!]
\centering
\includegraphics[width=69mm]{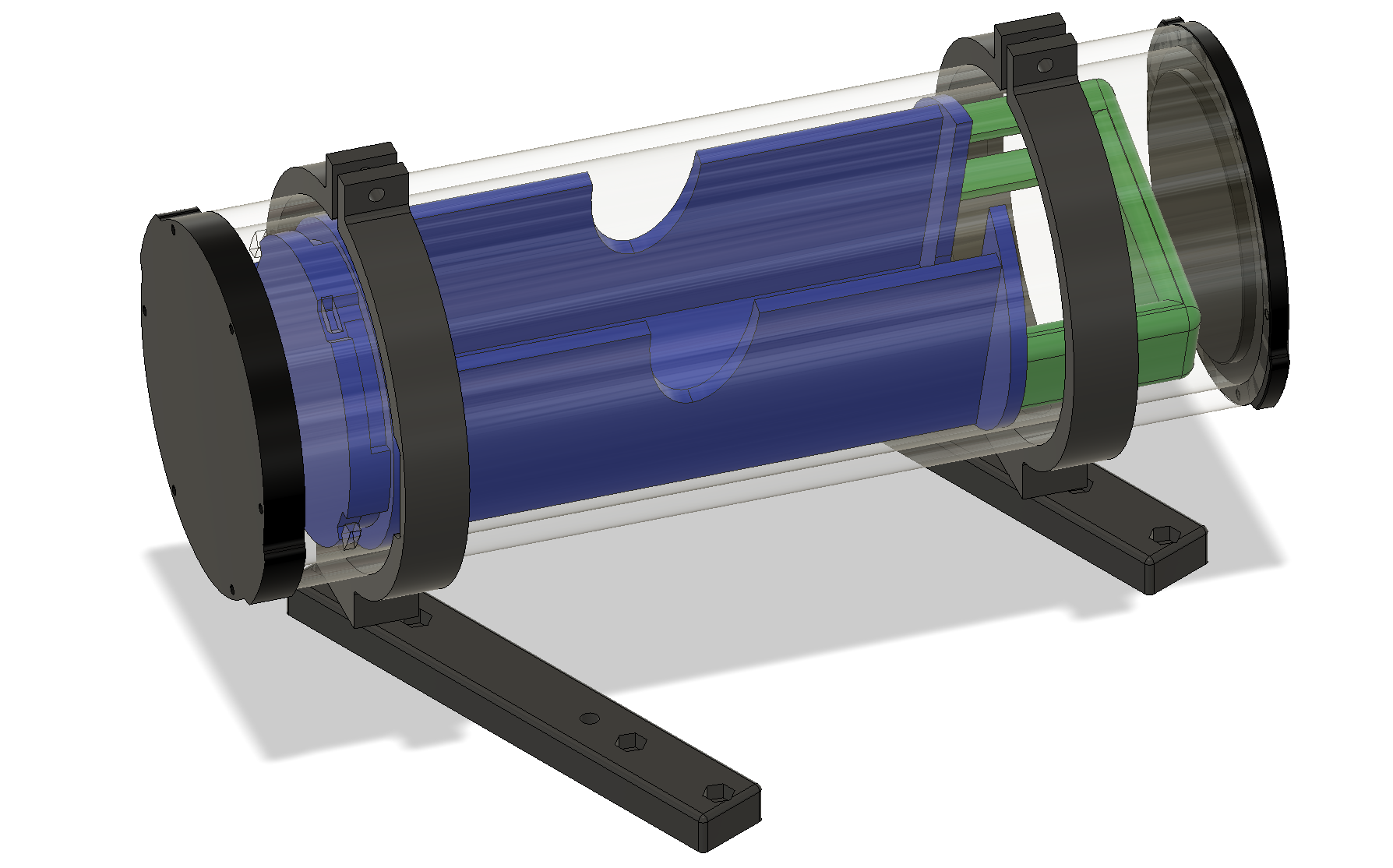}
\caption{Battery Capsule Subassembly}
\label{fig:bcs}
\end{figure}

However, we could not do the same for the battery capsule, as the ring needs to rotate separately from the body to allow the battery to be quickly and easily replaced. Therefore, we opted to instead increase the width, the infill density, and the wall thickness on the 3D print (Fig. \ref{fig:bcm}, Fig. \ref{fig:bcs}). This kept the design light and easy-to-fabricate, while also increasing the overall robustness of the mount. While much less prone to break than the old design, we still have occasional fractures of the mount at the same interface - the 3D-printed mount comes under stress when loading and unloading the heavy battery. An ideal design would eliminate the two-piece bracket (similar to the new camera mount) and is a priority next year. 


\subsubsection{Flexible Mounting} 
\begin{figure}[h!]
\centering
\includegraphics[width=69mm]{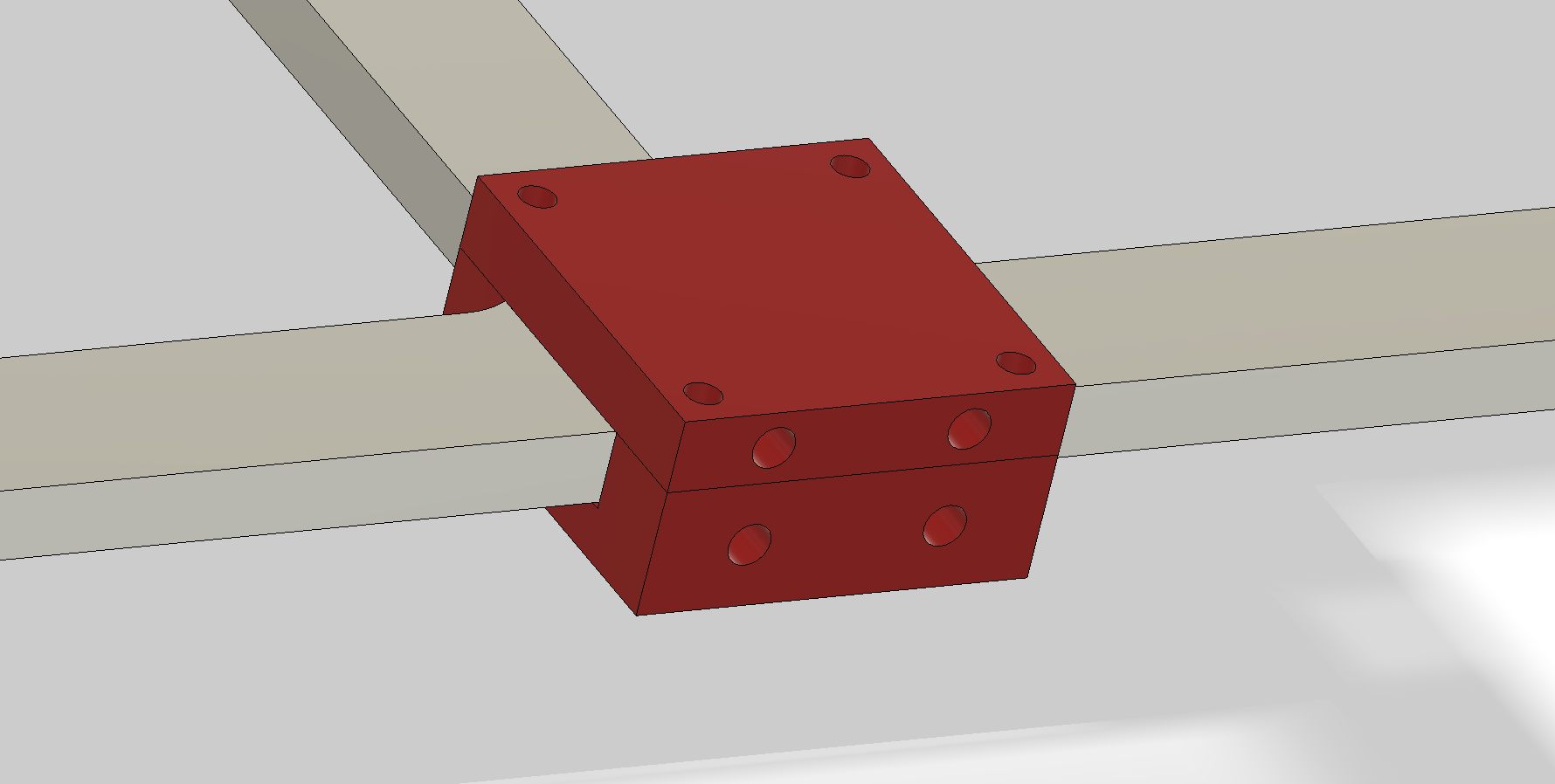}
\caption{Modular Mounting Block}
\label{fig:mmb}
\end{figure}

Per our goal of maximizing the modularity of the robot, one design element we trialed this year was the use of standardized mounting blocks (Fig. \ref{fig:mmb}). To ensure minimal interference with already working parts, we only tested this system on new components added this year. These 3D-printed blocks clamp around the aluminum frame and are fitted with brass inserts to allow components to be easily face-mounted. 

\begin{figure}[h!]
\centering
\includegraphics[width=69mm]{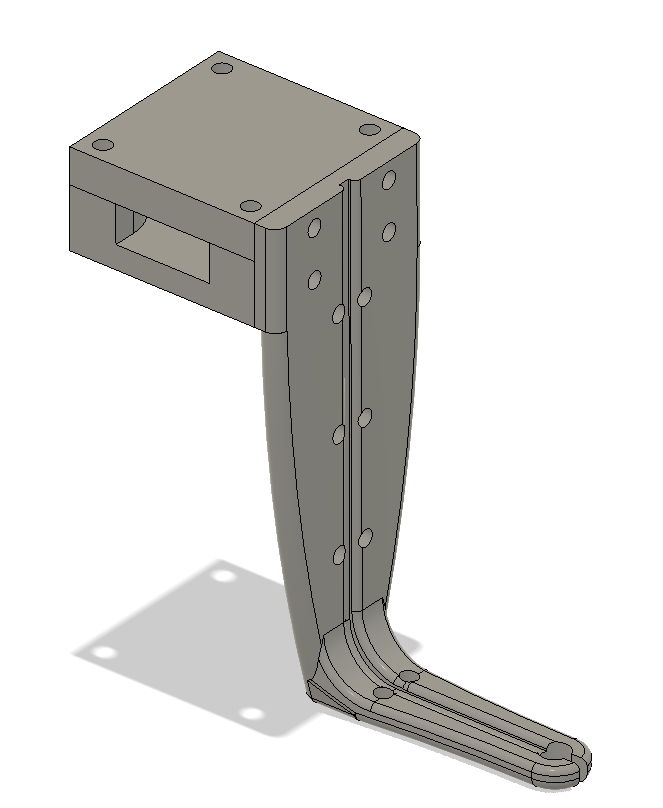}
\caption{Old Hydrohone Mount. Required four different mounting locations.}
\label{fig:oldhp}
\end{figure}

\begin{figure}[h!]
\centering
\includegraphics[width=69mm]{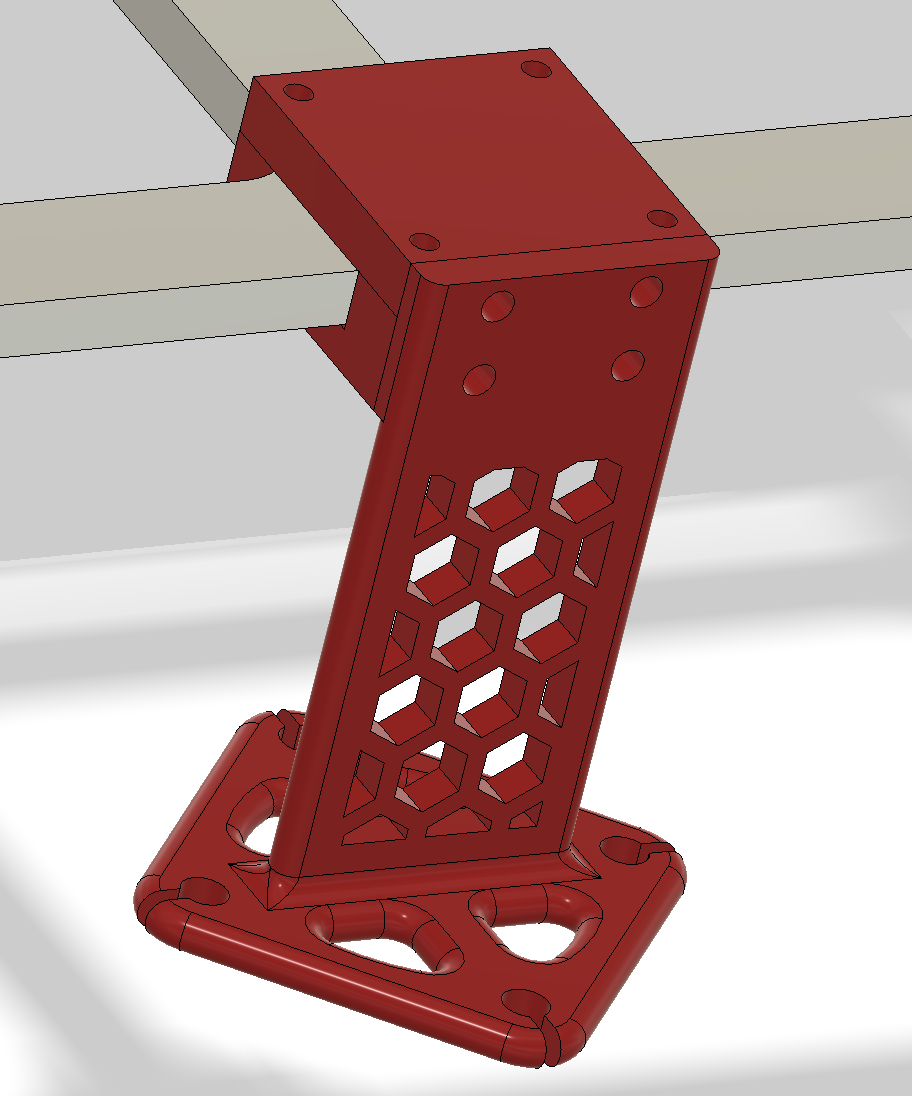}
\caption{New Hydrophone Mount. All hydrophones grouped together in one spot.}
\label{fig:newhp}
\end{figure}

We trialed this system on new components that were in flux, such as the hydrophone mounts (Fig. \ref{fig:oldhp}, Fig. \ref{fig:newhp}). As the acoustics subteam was testing different mounting locations and styles, the modular mounting block system allowed the hydrophone orientation to be modified easily. 

\begin{figure}[h!]
\centering
\includegraphics[width=69mm]{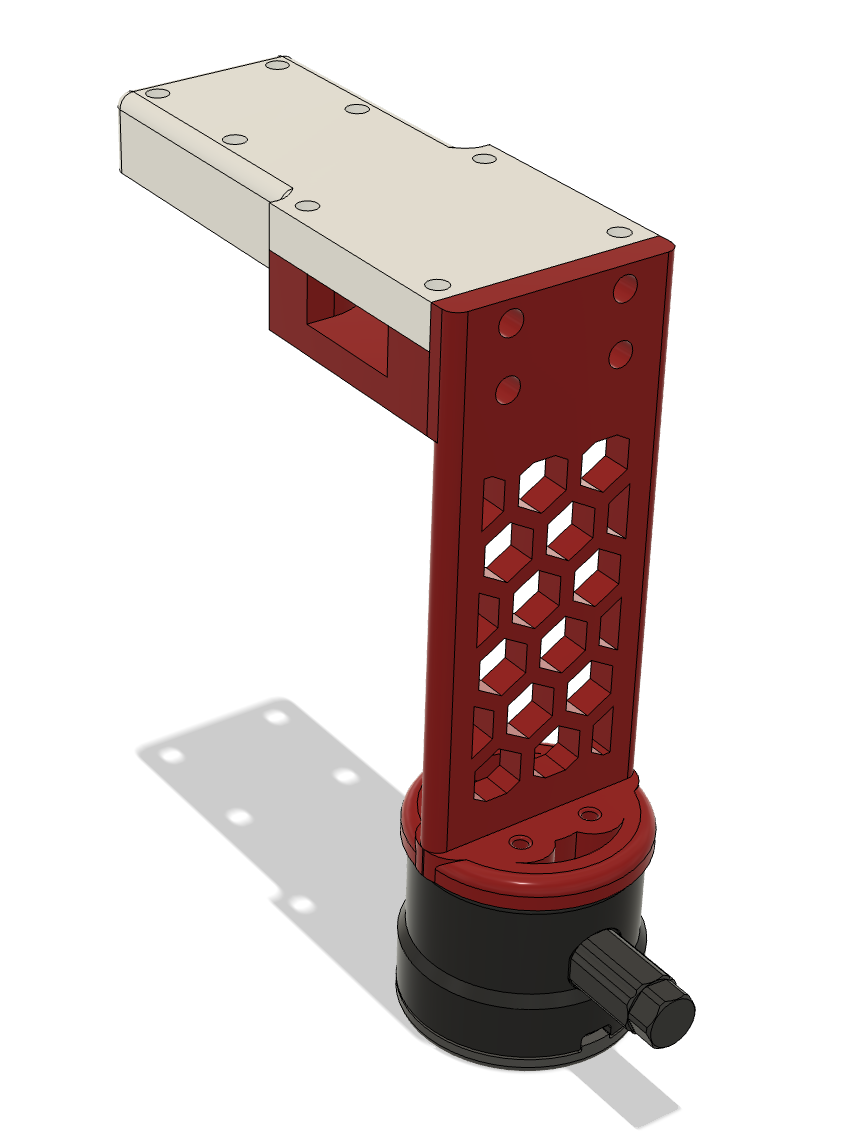}
\caption{1D Sonar Mount. This traded spots with the hydrophone mount a few times when finding the optimum location.}
\label{fig:minisonar}
\end{figure}

We utilized the same system for the 1D sonar, another new addition to the robot (Fig. \ref{fig:minisonar}). By having an easily adjustable location, we were able to find and mount it in the most effective spot on the robot. 

The trial was a clear success for the new standardized mounting solution: it is compact, and parts are easy to attach and detach. Old mounts will be adapted to this new design next year. 

\subsubsection{Torpedo Mechanism} One of the components that underwent the most iterations this year was the torpedo mechanism. The testing process was three-fold: first, a prototype would be tested on dry land to ensure the torpedoes could be safely discharged without causing harm. We would then use the sink to test underwater performance. While the length of the sink was short (only around two feet), we could identify any obvious problems with the prototype quickly without needing to go to the pool. Finally, the prototype would be tested in the pool to gauge its accuracy and distance in a realistic environment. 

\begin{figure}[h!]
    \centering
    \begin{subfigure}[b]{0.45\columnwidth}
        \centering
        \includegraphics[width=32mm]{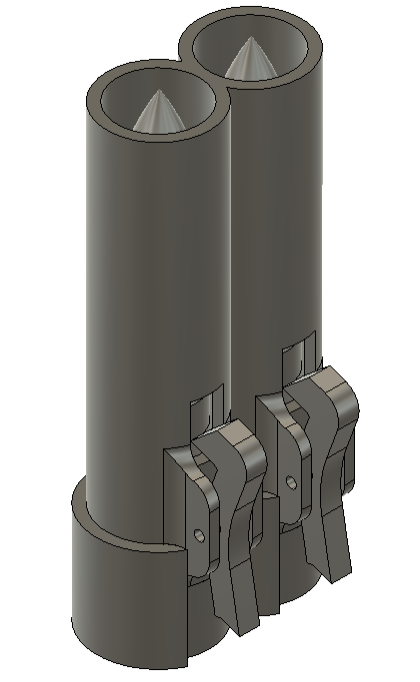}
        \caption{First Torpedo Tube Prototype}
        \label{fig:tube1}
    \end{subfigure}
    \hfill
    \begin{subfigure}[b]{0.45\columnwidth}
        \centering
        \includegraphics[width=48mm]{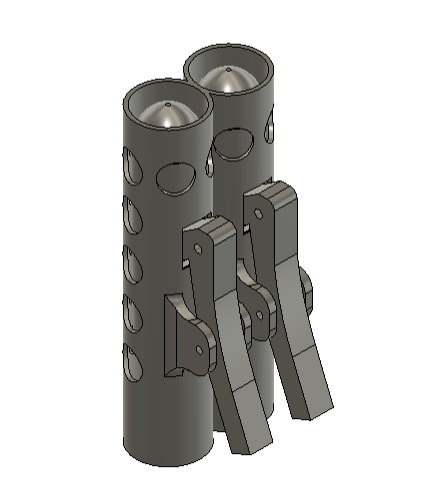}
        \caption{Current Torpedo Tube Prototype}
        \label{fig:tube2}
    \end{subfigure}
    \caption{Torpedo Tube Design Evolution}
    \label{fig:torpedotube}
\end{figure}

This testing plan has allowed us to fine-tune the torpedo tube design: we were able to pinpoint the minimum safe compression length for the spring using the dry land test, and the sink test showed that we needed holes to allow the torpedo to push water out of the way as it exits the chamber (Fig. \ref{fig:torpedotube}).  

\vspace{-2pt}

\vspace{-5pt}

\begin{figure}[h!]
    \centering
    \begin{subfigure}[b]{0.45\columnwidth}
        \centering
        \includegraphics[width=40mm]{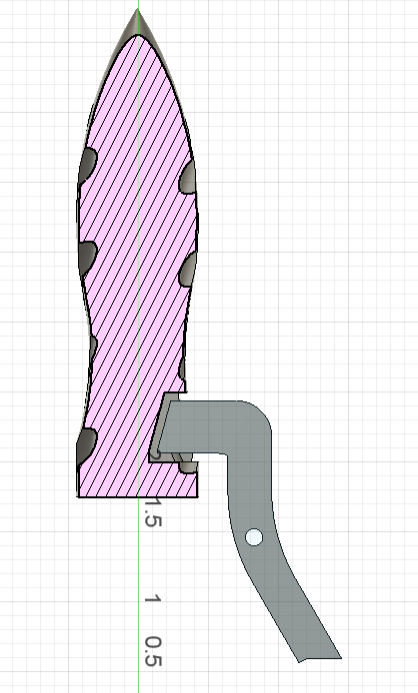}
        \caption{First Trigger Mechanism Prototype}
        \label{fig:trigger1}
    \end{subfigure}
    \hfill
    \begin{subfigure}[b]{0.45\columnwidth}
        \centering
        \includegraphics[width=40mm]{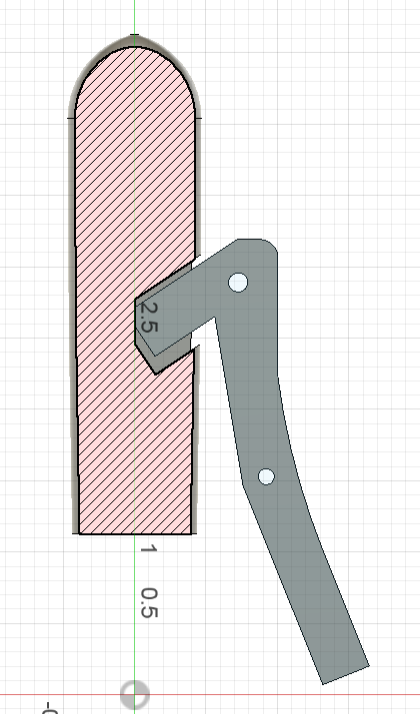}
        \caption{Current Trigger Mechanism Prototype}
        \label{fig:trigger2}
    \end{subfigure}
    \caption{Trigger Mechanism Design Evolution}
    \label{fig:trigger}
\end{figure}

The dry land test also allowed us to optimize the design of the firing mechanism. Original prototypes were initially too weak due to their PLA construction and the shallow angle meant that misfires were common (Fig. \ref{fig:trigger1}). We corrected this by giving the trigger a larger surface area and steeper angle, ensuring it could keep the spring compressed and release it correctly (Fig. \ref{fig:trigger2}). While this mechanism works well, there is still work to be done integrating it with a servo motor. We are currently designing a system with rods offset 180° from each other on the servo horn. These will individually push down on each trigger, allowing the torpedoes to be fired independently. 

\vspace{-2pt}
 
\begin{figure}[h!]
    \centering
    \begin{subfigure}[b]{0.45\columnwidth}
        \centering
        \includegraphics[width=20mm]{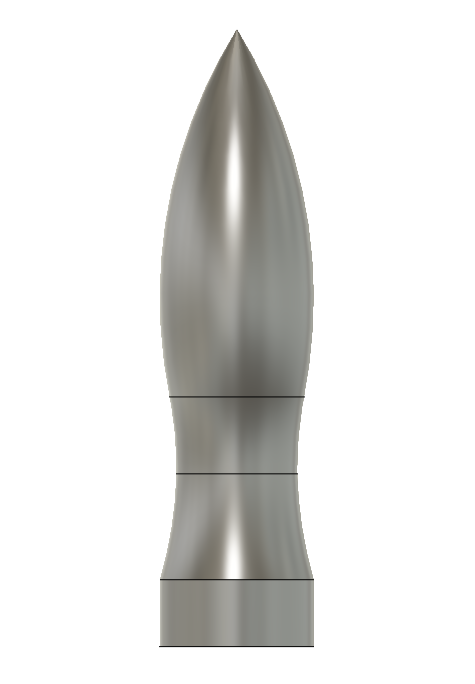}
        \caption{First Torpedo Prototype}
        \label{fig:t1}
    \end{subfigure}
    \hfill
    \begin{subfigure}[b]{0.45\columnwidth}
        \centering
        \includegraphics[width=15mm]{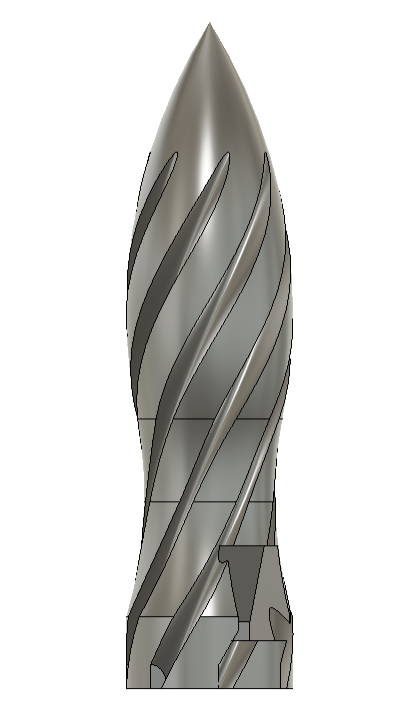}
        \caption{Second Torpedo Prototype}
        \label{fig:t2}
    \end{subfigure}
    \hfill
    \begin{subfigure}[b]{0.45\columnwidth}
        \centering
        \includegraphics[width=20mm]{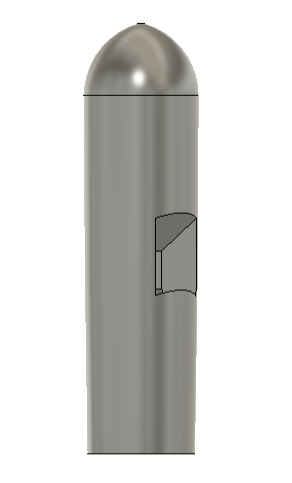}
        \caption{Current Torpedo Prototype}
        \label{fig:t3}
    \end{subfigure}
    \caption{Torpedo Design Evolution}
    \label{fig:torpedo}
\end{figure}

The sink test also allowed us to improve on the design of the actual torpedo. We initially experimented with fins to maintain a straight line, but this failed to show results in the sink (Fig. \ref{fig:t1}). We then introduced a spiral pattern to help maintain their path, which once again failed to show results (Fig. \ref{fig:t2}). Eventually, we designed the torpedoes as cylinders with spherical tops tapering towards the bottom (Fig. \ref{fig:t3}). While the current prototype occasionally travels in a straight line, it is not consistent. Moreover, the 3D-printed torpedoes tend to float, giving them an undesirable upward trajectory. Future work consists of running fluid simulations to optimize torpedo shape and determine why the torpedo sometimes varies from its path. 

\vspace{-2pt}

\subsubsection{Robot Assembly} 
\begin{figure}[h!]
\centering
\includegraphics[width=69mm]{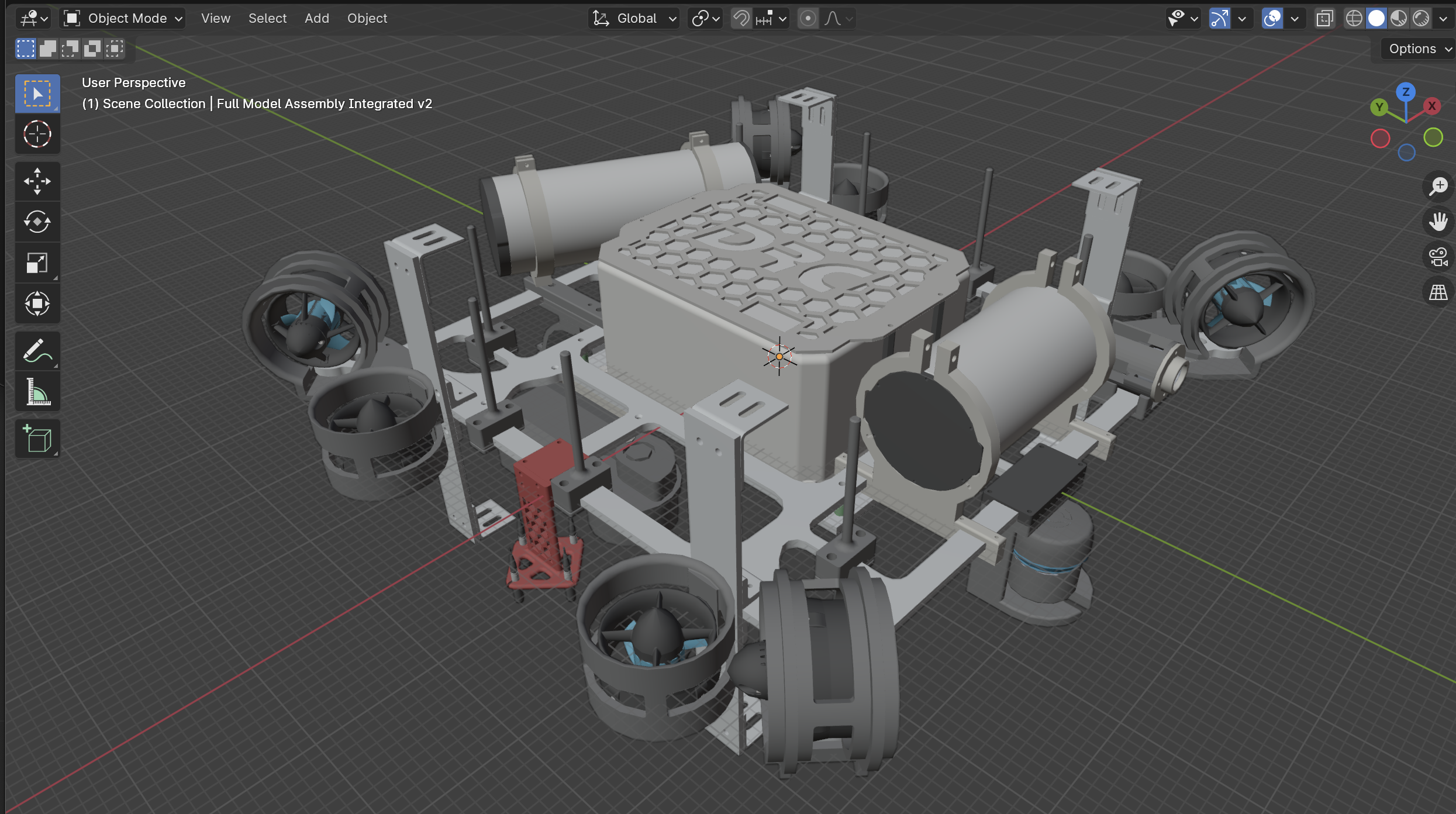}
\caption{Full Robot Assembly in Blender.}
\label{fig:blender}
\end{figure}

With all the new components this year, much more focus was put on maintaining an up-to-date full robot assembly. This has become an integral part of our prototyping process and has allowed us to better understand how components fit together and identify conflicts before fabrication. Moreover, as we further develop the task planning simulation next year, the accurate assembly will provide a starting point to roughly tune movement constants before putting the robot in the pool. \\
This assembly has also allowed our team to construct higher-quality media, such as all the mechanical diagrams found in this paper and the render in Figure \ref{fig:render}. For the first time this year, we also leveraged Blender to convert the assembly into the compact glb file format, allowing us to embed a 3D view of the robot directly on our website (Fig. \ref{fig:blender}).

\subsection{Electrical}

Rigorous testing is the foundation of ensuring electrical reliability. Our testing surrounds the two principal responsibilities of the electrical subsystem: power distribution and data acquisition, in addition to research and development surrounding our acoustics system.

\subsubsection{Power}

The electrical team must assume that the entire electrical stack will undergo extensive use under very demanding circumstances. This includes significant power, high heat, and surviving mechanical imperfections such as vibrations and controlled water ingress.

Due to the limited battery life of the batteries, the team constructed a dedicated testing rig to stress test the entire system for durations far exceeding those on a single battery charge. This rig involved a modified power supply serving as a battery emulator, which we configured to have the same power characteristics as our 4S 12C lithium battery. During our 6-hour testing, we logged many system vitals, such as capsule temperature and sensor drift.

The results of our testing showed a dramatic accumulation of heat, with internal temperatures nearing 70ºC. After inspection, we deduced that the main cause was the POE network switch, which quite inefficiently drew excess power that it did not deliver over unconnected ports. We took the opportunity to replace it with a lower-voltage (12V versus the original 53V) switch and saw notable improvements as temperatures stabilized at around 55ºC. We also took the opportunity to upgrade to a 1G SFP switch, which supported the team goal of refactoring to the improved water-resistance of a fiber optic tether with added network speeds required by an advanced GUI framework with additional camera feeds.

The migration to this new switch did prompt an engineering trade-off decision. Specifically, since we have two POE cameras on Oogway, we then had to create our own POE injectors, per the IEEE 802.3a Class A standard. In making this decision, we consulted with a former Duke Robotics member who currently works professionally as an electrical engineer on POE-based networking, and concluded that this approach would simplify the stack and improve thermal performance.

Further, this long-duration power test was repeated with the robot in various orientations, notably completely inverted. The main usage of this test was to determine if mechanical attachments and adhesives would fail with heat accumulation. We also tested the robot while undergoing intense vibrations through use of a dolly. This allowed the team to pinpoint mechanical attachments that would eventually fail. This included nuts that were replaced with lock nuts, components that would slide out of 3D printed mounts, and connectors that would expose their conductors below their shielding.

The battery testing rig that we developed now serves an additional purpose of running the robot during typical lab testing and development without diminishing the capacity of our existing batteries.

\subsubsection{Data}

As described earlier, the team noticed hardware defects within our ESCs. Table \ref{tab:esc-data} outlines our first-difference based calculations for various PWM signals, showing both our input efforts, their corresponding PWM durations, and the measured offsets calculated with an oscilloscope's trigger. The result of these calculations for this specific ESC yielded a very consistent offset of 31$\mu$s, which we then utilized in our firmware configuration. This process was applied for each ESC.

\begin{table*}[ht]
\centering
\begin{tabular}{|r|r|r|r|r|r|r|r|}
\hline
int8 & PWM Sent & PWM Measured & Sent - Measured & Measured Diff. & Test Diff. & PWM Test & Test Diff. \\ \hline
-128 & 1131 & 1100 & 31 &  &  & 1100 & 31 \\ \hline
-120 & 1156 & 1120 & 36 & 20 & 25 & 1125 & 31 \\ \hline
-110 & 1187.25 & 1160 & 27.25 & 40 & 31.25 & 1156.25 & 31 \\ \hline
-100 & 1218.5 & 1180 & 38.5 & 20 & 31.25 & 1187.5 & 31 \\ \hline
-90 & 1249.75 & 1220 & 29.75 & 40 & 31.25 & 1218.75 & 31 \\ \hline
-80 & 1281 & 1240 & 41 & 20 & 31.25 & 1250 & 31 \\ \hline
-70 & 1312.25 & 1280 & 32.25 & 40 & 31.25 & 1281.25 & 31 \\ \hline
-60 & 1343.5 & 1300 & 43.5 & 20 & 31.25 & 1312.5 & 31 \\ \hline
-50 & 1374.75 & 1341 & 33.75 & 41 & 31.25 & 1343.75 & 31 \\ \hline
-40 & 1406 & 1360 & 46 & 19 & 31.25 & 1375 & 31 \\ \hline
-30 & 1437.25 & 1400 & 37.25 & 40 & 31.25 & 1406.25 & 31 \\ \hline
-20 & 1468.5 & 1420 & 48.5 & 20 & 31.25 & 1437.5 & 31 \\ \hline
-10 & 1499.75 & 1460 & 39.75 & 40 & 31.25 & 1468.75 & 31 \\ \hline
0 & 1531 & 1480 & 51 & 20 & 31.25 & 1500 & 31 \\ \hline
10 & 1562.25 & 1520 & 42.25 & 40 & 31.25 & 1531.25 & 31 \\ \hline
20 & 1593.5 & 1541 & 52.5 & 21 & 31.25 & 1562.5 & 31 \\ \hline
30 & 1624.75 & 1580 & 44.75 & 39 & 31.25 & 1593.75 & 31 \\ \hline
40 & 1656 & 1608 & 48 & 28 & 31.25 & 1625 & 31 \\ \hline
50 & 1687.25 & 1640 & 47.25 & 32 & 31.25 & 1656.25 & 31 \\ \hline
60 & 1718.5 & 1680 & 38.5 & 40 & 31.25 & 1687.5 & 31 \\ \hline
70 & 1749.75 & 1700 & 49.75 & 20 & 31.25 & 1718.75 & 31 \\ \hline
80 & 1781 & 1740 & 41 & 40 & 31.25 & 1750 & 31 \\ \hline
90 & 1812.25 & 1760 & 52.25 & 20 & 31.25 & 1781.25 & 31 \\ \hline
100 & 1843.5 & 1800 & 43.5 & 40 & 31.25 & 1812.5 & 31 \\ \hline
110 & 1874.75 & 1820 & 54.75 & 20 & 31.25 & 1843.75 & 31 \\ \hline
120 & 1906 & 1860 & 46 & 40 & 31.25 & 1875 & 31 \\ \hline
127 & 1927.875 & 1880 & 47.875 & 20 & 21.875 & 1896.875 & 31 \\ \hline
\end{tabular}
\caption{Measured PWM offsets for a single ESC}
\label{tab:esc-data}
\end{table*}

\begin{figure}[h!]
\centering
\includegraphics[width=60mm]{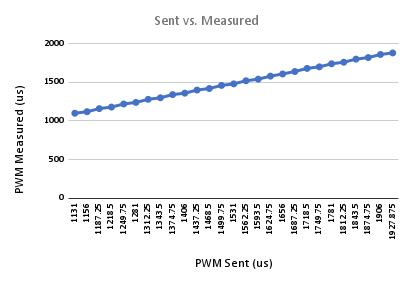}
\caption{Plotted differentials between sent and measured PWM shows linear offset, which we then applied the inverse of in our corrective software.}
\label{fig:sent-measured}
\end{figure}

An additional result of the extended power testing described earlier also unveiled what ended up being a rare, but catastrophic memory leak in our microcontroller firmware bridging ROS and the ESC interface. As indicated by our extensive logging within ROS, we were able to detect these outages before the microcontroller was able to reboot. Upon correction, we found autonomous performance of the robot to improve to a noticeable degree. This included general smoother movement and faster response times. We concluded that, while the microcontroller would only occasionally crash, the memory leak buy would gradually degrade performance on the memory-constrained device, which is why correcting this bug improved performance overall.

Another data reliability issue we encountered during our testing pertains to our POE cameras. We noticed that the streams would drop after many hours of use, and they would only return functionality once rebooted. Despite the outages, the issue only occurred after at least an hour, which is not an expected runtime during official runs. That, combined with double-coverage on most of cameras and sensors, and the determination that the issue is likely a memory leak within the camera's internal firmware, led the team to consider the issue to not impact Oogway's performance.

It is inevitable that trace amounts of water will enter the capsule during an extended pool test. This is due to the various reasons described earlier in the mechanical sections. While not catastrophic, we have been able to validate our humidity sensor data by removing the robot from the pool after a spike in the data and subsequently verifying that trace amounts of water has entered. Specifically, we use humidity sensing because some of the water evaporates upon entering the warm capsule. Likewise, our other internal monitoring sensors, voltage and temperature, have also been empirically verified during dedicated pool test runs. All such sensors have proved very useful and practical.

\subsubsection{Acoustics Hardware}

Our acoustics PCBs were designed the thought that we would only need low-pass filters. However, when mounted to Oogway, low frequency noise from the thrusters and other inductive loads deteriorated the signal-to-noise ratio (SNR). To surmount this, we devised a bandpass filter (BPF). Remarkably, this did not require any modification to the existing circuitry as a backpack was designed to attach directly to the already-made boards.

\begin{figure}[h!]
\centering
\includegraphics[width=69mm]{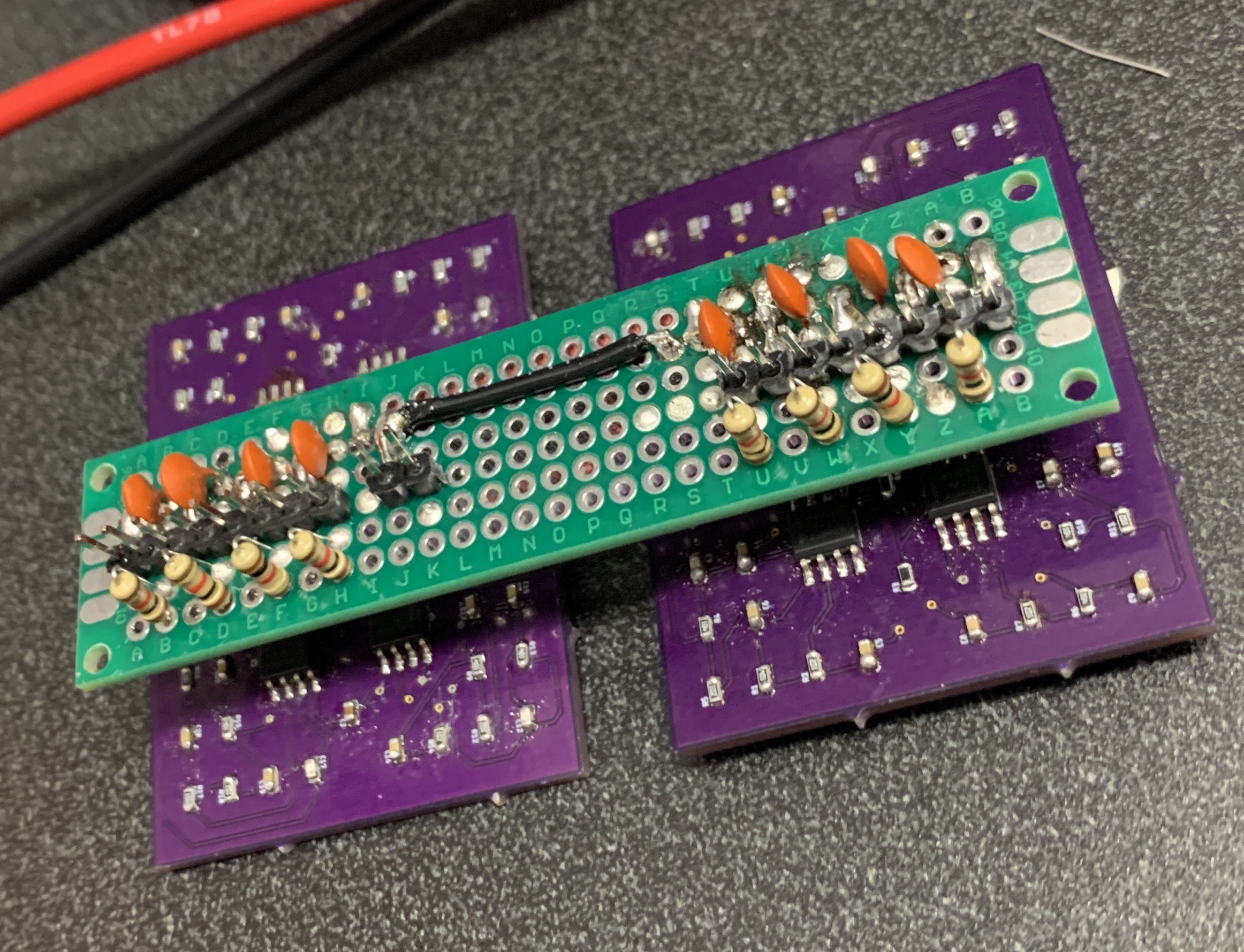}
\caption{Acoustics Filter and Amplifier Boards with BPF Backpack}
\label{fig:method}
\end{figure}

With the improved filters, we constructed a testing setup wherein the gain of the amplifiers could be changed in real time. Through a series of tests in a pool, a gain on the order of magnitude of $10 ^ 1 V/V$ provided visual indication of the ping in the raw data, even from tens of meters away. 

\begin{figure}[h!]
\centering
\includegraphics[width=69mm]{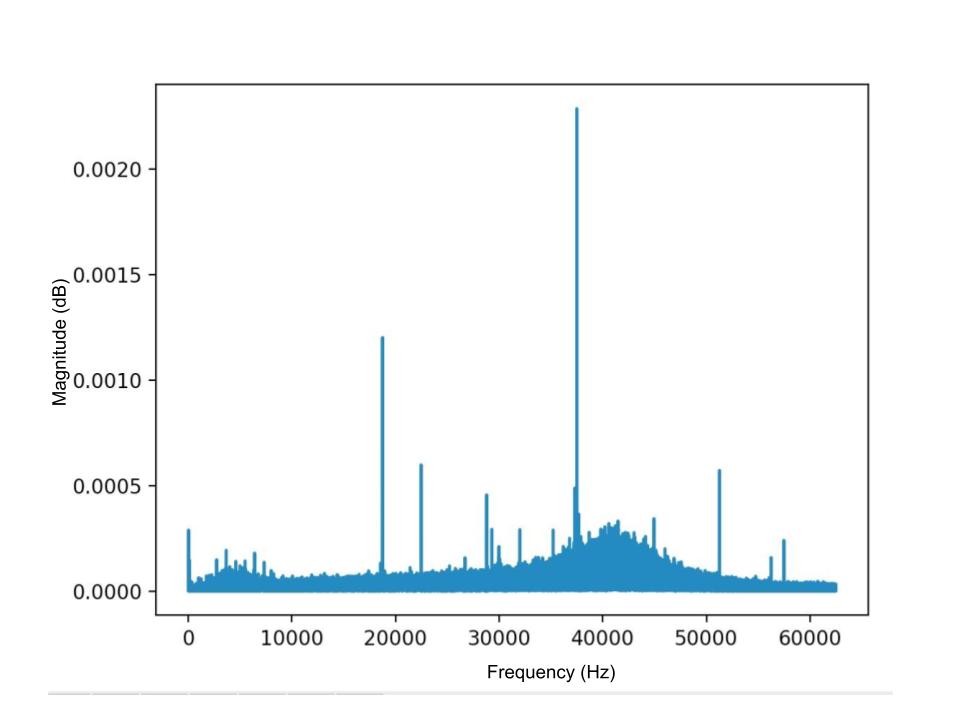}
\caption{Magnitude of frequency response of filters in pinger-less pool. Bandpass 'hump' visible at $\sim$40kHz is about the same as the frequency range of our pinger.}
\label{fig:method}
\end{figure}

\subsubsection{Acoustics Software}

The acoustics software is composed of Python and Rust scripts working together, and the choices made are a reflection of our development process.

Our first attempts included tests with three hydrophones, using a target with a known x and y location in the plane. This required exact measurements within the pool relative to the position and rotation of the robot. Gathering this data was slow so there were only a few data points to work with. Among those few data points received, the data was too noisy to be reliably predicted through ideal mathematical means. A tensorflow neural network trained to predict distance and angle on this data produced wildly varied results, as there weren't enough data points to accurately train the model. This led to the idea of collecting data in discrete angular quantities.

Splitting the angular space into octants allowed the team to gather much more data in the pool, as we could drag the pinger around the pool without worrying about its exact location. Predictions within octants would also be fine enough to steer the robot in the desired direction.

\begin{figure}[h!]
\centering
\includegraphics[width=69mm]{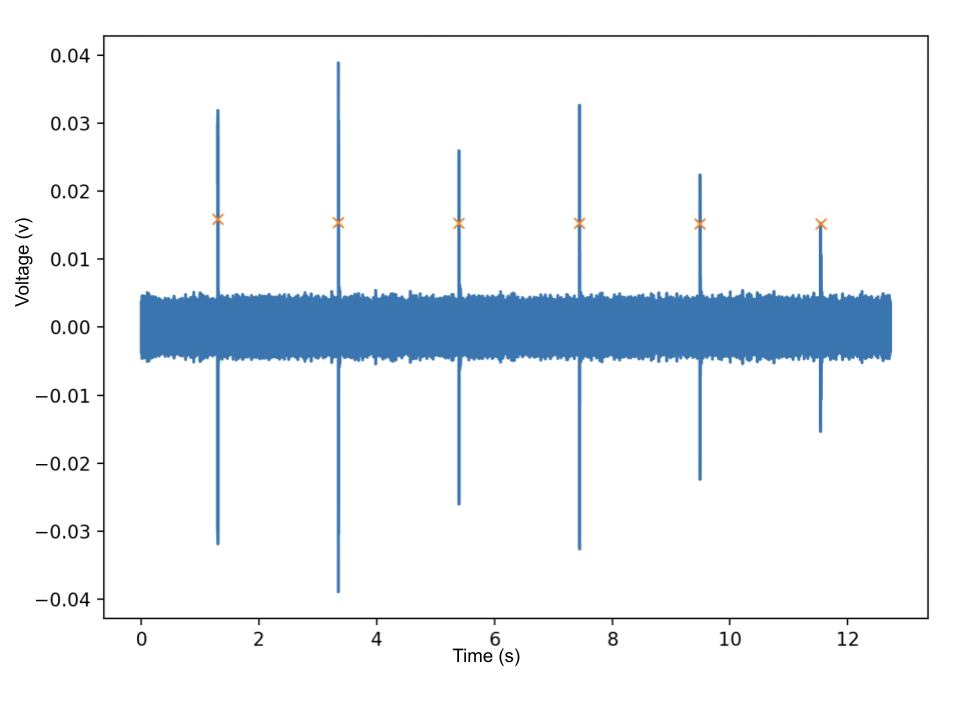}
\caption{Finding pings amidst background noise}
\label{fig:pings}
\end{figure}

The first task in the pipeline is to detect pings from a single hydrophone. Timing is important, so the goal is to find the ping in the audio source when it started. Finding the ping using noise level alone didn't work with different noise levels, and the derivative of the sound wave curves produced too many false positives. Using the envelope of the sound wave increased accuracy, but it was too computationally expensive. In the end, we used a dynamic threshold based on the mean peak levels of the background noise. This made it adaptive to background noise levels, with few if any false positives.

\begin{figure}[h!]
\centering
\includegraphics[height=42mm]{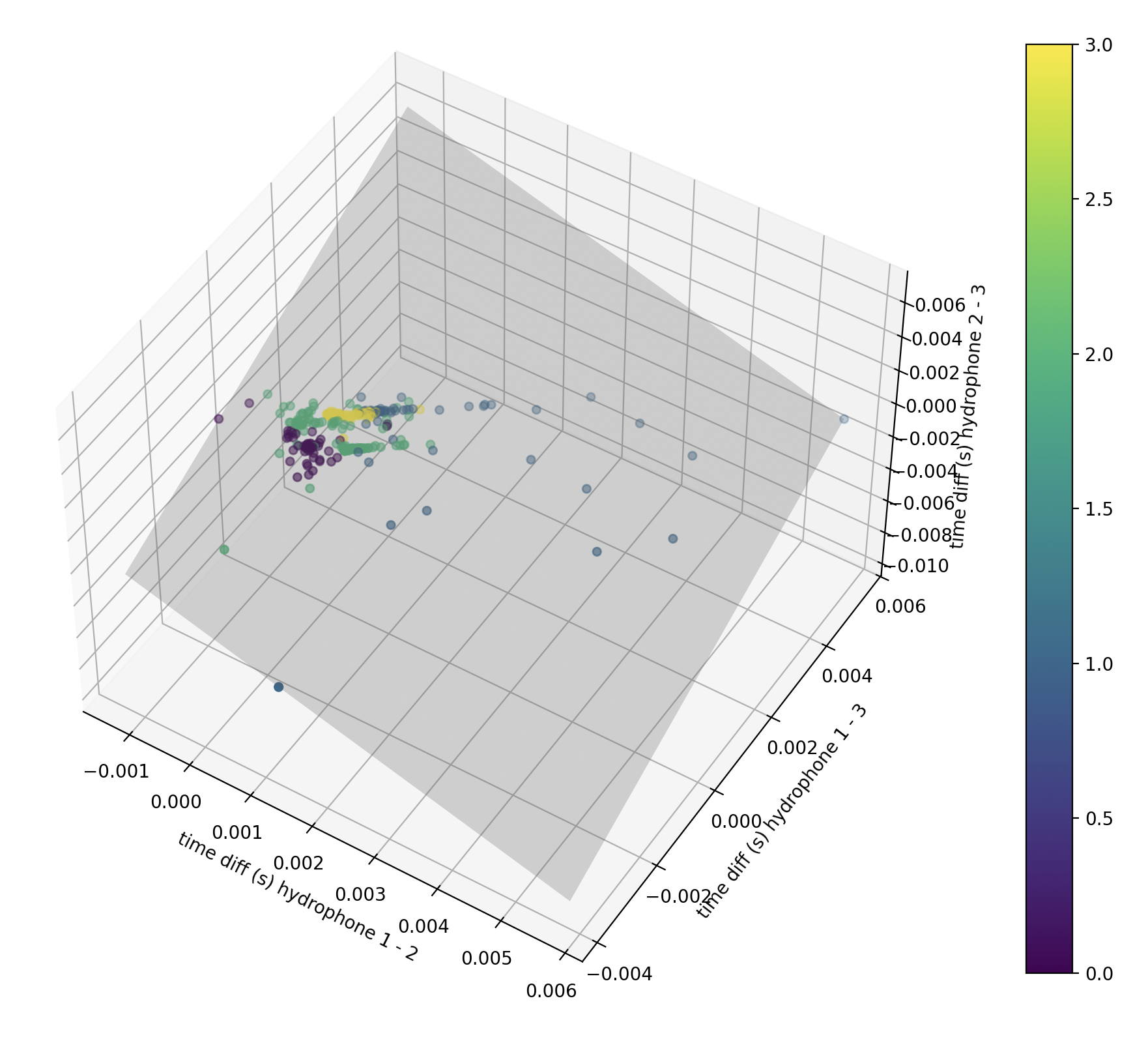}
\\
\text{(a)}
\\
\includegraphics[height=42mm]{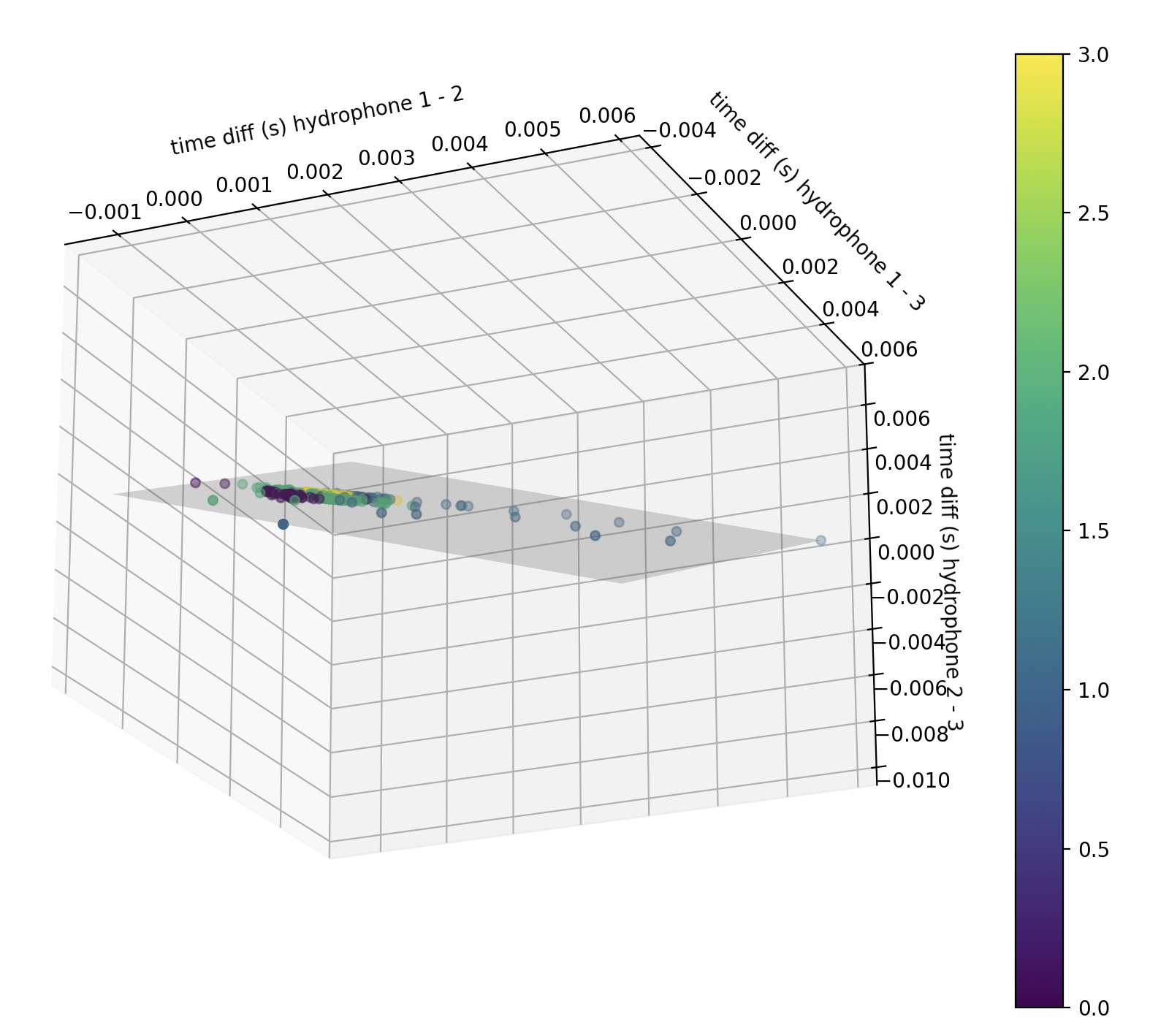}
\\
\text{(b)}
\caption{Classifying the acoustics subspace. Each dot is a ping measurement across three hydrophones. (a) rotation 1 (b) same graph different rotation} 
\label{fig:subspace}
\end{figure}

Once we could successfully detect pings as they started, we started trying to train a classifier into octants. Soon after we started, we noticed the dependence shown in Fig. \ref{fig:subspace}: due to geometric constraints across three hydrophones, the measured relative ping timings are dependent. The three sets of relative measurements across the three hydrophones (1-2, 1-3, and 2-3) is rank-2, meaning that only two of those attributes are needed to contain all the information. Thus, all the data  points collapse onto a plane, shown in gray on the graphs. This is useful, as when training we were able to use reduce the size of each input to only the (1-2, 1-3) measurements.

\begin{figure}[h!]
\centering
\includegraphics[width=69mm]{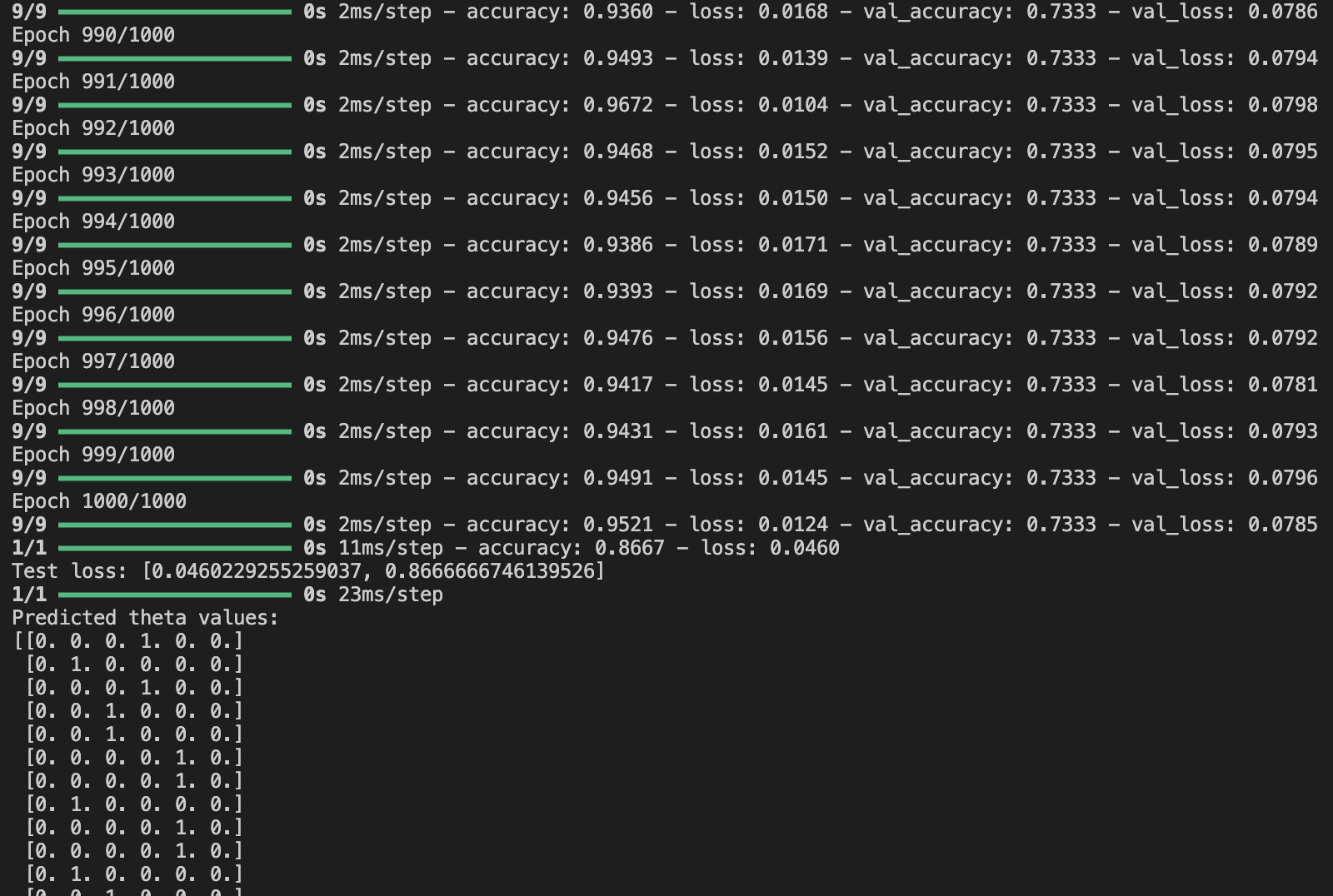}
\caption{Training a tensorflow model on the data}
\label{fig:pings}
\end{figure}

Finally, we trained a neural network on the hundreds of examples we were able to collect from all around the pool. After preprocessing the data to include only what was needed, the network achieved 86\% accuracy on the test dataset. With guesses this accurate, the robot can quickly and effectively find the location of the pinger and the probability of determining the global optimization in pinger location is greatly increased.

Lastly, once this pipeline was finished, it was too slow. Python itself is an interpreted language with large overhead, so the code itself could only be optimized so much. We set out to compile the code closer to the hardware with Rust. Rust is a compiled language that has about the same speed as C. One-by-one, the lines of code were translated from Python to Rust, making use of rust-specific optimizations as well. With the whole setup, the time to process and classify data from the hydrophones was dramatically decreased - from ~25 seconds to ~1.5 seconds. This code was then packaged into a Python library using pyo, allowing it to be integrated with ROS through python.

The acoustics system is currently still in development. This is mostly due to difficulties mounting the hydrophones properly where they are  not blocked by Oogway's other components and do not suffer from thruster RF noise. Additionally, the test data was collected off of the robot on a separate custom rig. So to account for this, we would need to re-record the data on the robot in a possibly different layout. Further, the 

We also considered using the MUSIC algorithm, which uses phase differences to determine the source location of a sound. However this requires extremely precise phase information, which we have not yet been able to produce. Additionally, the hydrophones would need to be moved to within a wavelength of each other. We look to developing our own DAQ with custom hardware processing to combat this issue as a point of future research.

\subsection{Software}
The majority of the individual software components passed component testing without many problems.

\subsubsection{State}
Before the controls rewrite, we ran into verification issues with our robot's state. Like most AUVs, we represent our state in 6 variables: \textit{x}, \textit{y}, \textit{z}, \textit{roll}, \textit{pitch}, \textit{yaw}. When manually moving Oogway forward the \textit{x} and \textit{y} components of the state would increase drastically once the robot stopped moving. When debugging, we mainly looked at the output from our three positional sensors, the IMU, DVL, and pressure sensor. We quickly realized that the data from the IMU and DVL were in different coordinate frames; the IMU is right-handed and the DVL is left-handed. After changing the configurations, our state was more stable. However, we decided to create a testing plan in case of future issues with the state.

Our testing plan for issues with the state is as follows:
\begin{enumerate}[leftmargin=*]
    \item First, restart the robot, all Docker containers, and ensure that the Arduino is connected. Then turn the power DVL switch on.
    \item Verify that all of the DVL, IMU, and pressure sensor are outputting to the correct ROS topics.
    \item Verify that the robot-localization ROS package is running.
    \item Manually move the robot forward, to the left, and up. Verify that the DVL's velocities in each of these directions is positive.
    \item Manually rotate the robot counter-clockwise about the z-axis. Verify that the IMU's yaw value is increasing.
    \item Leave the robot stationary and observe the frequencies of the DVL, IMU, and pressure sensor. Verify that they are all above 20Hz.
    \item Leave the robot stationary and observe the \textit{/state} topic's covariance matrices. Verify they are not increasing over time.
\end{enumerate}

Later in the season, after the controls rewrite, we ran into similar issues with the state. Following our test plan, we realized that the pressure sensor would only publish in bursts of data and not at a constant rate. This causes the state to over-rely on the other two senors for inaccurate depth information. After updating the Arduino code to resolve this issue, we observed that while the robot was stationary, the positional covariance would sometimes increase rapidly. This prompted a deep dive into our sensors configurations. 

After many hours of testing we realized that there were two large issues with our IMU's configuration. The first was that ROS's robot-localization expected all quaternion data to be in ENU (East North Up) format while the IMU was outputting data in NED (North East Down) format. This caused Oogway's state to sometimes have a mismatch between the DVL's coordinate frame and the IMU's coordinate frame leading to drift over time. Another issue we found was with magnetic interference from Oogway's thrusters. The IMU does not have hard/soft iron adaptive filtering enabled by default, and once we enabled it, we observed a large increase in state stability.

\begin{figure}[h!]
\centering
\includegraphics[width=\columnwidth]{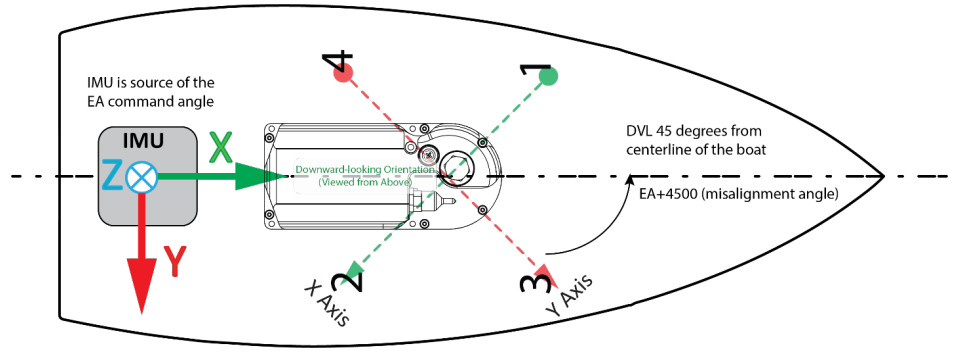}
\caption{Correct allignment of the IMU with the DVL ~\cite{b13}}
\label{fig:teledyne}
\end{figure}

Further, our testing also involved ensuring compliance with the manual for each of the three positional sensors. This included referencing diagrams such as Figure \ref{fig:teledyne}, which imposed mechanical constraints. In the case of the DVL, where we were not able to mount it in the direct center of the robot, we applied a corrective static transform.

\subsubsection{Controls}

The rewrite of the controls system demanded careful yet through testing of the new code. We took a step-by-step approach, correcting bugs along the way, and covered a wide variety of scenarios. To prevent the robot from making any sudden, unintended moves, we disabled the thrusters during these tests.
\begin{enumerate}[leftmargin=*]
    \item We first checked that when controls received a setpoint, it computed the correct error values, which determine where the robot will attempt to move. We fixed a bug that performed an incorrect transform from global to local coordinates.

    \item Once the errors were accurate, they were fed into the PID loops, which computed their integrals and derivatives. We made sure that these were the correct values, and tuned the parameters of the Butterworth filter for optimal smoothing of the derivative.

    \item We then checked that the control efforts output by the PID loops would lead to the robot achieving its desired state.

    \item We checked the dynamic offset added to the PID outputs to counteract the robot's positive buoyancy. We rolled, pitched, and yawed the robot many different directions and ensured that the offset vector always pointed straight up, regardless of the robot's orientation.

    \item We then verified that the thrust allocations obtained from our quadratic programming solver matched our expectations. We found that a few values were unexpectedly negated, which we corrected.

    \item We modified the PID gains and other values on-the-fly, and ensured that they updated the system's behavior as expected and were saved to disk so they could be reused in future runs.

    \item With all parts of the system independently tested and verified to be working as expected, we finally enabled the thrusters and allowed the robot to move using the new controls system. The robot behaved as we expected, and didn't perform any movements that would damage itself.
\end{enumerate}

\subsubsection{Foxglove GUI}
Of particular note this year is our emphasis on tight integration of ROS with our Foxglove GUI developer tools. By automatically converting ROS message schemas to TypeScript types~\cite{b11}, we are able to enforce static type checking to validate the connections between our landside GUI and onboard ROS system. This means that \textit{at development-time}, TypeScript will alert us if any concurrent work done on our ROS stack (e.g., a change to our PID service API) will interfere with our Foxglove GUI stack. This allows us to catch numerous bugs early, before any actual testing has occurred. 

\subsubsection{CV \& Sonar}
The CV object detection worked in the pool on the first attempt. After some initial data collection, the sonar correctly identified the buoys and the gate with no additional testing. However, the integration between CV and sonar required many hours of testing in the pool. We found that the angles of the CV camera and sonar did not always match so the sonar would scan empty pool space. Additionally, we found that the sonar has a minimum range of about a meter so the robot would run through the buoy without stopping and backing up. Both of these are limitations in the hardware of the sonar so we may consider buying a different model for future competitions.

\clearpage 
\section*{Appendix C: Custom Foxglove Extensions \& Layouts}

\begin{figure}[h!]
\centering
\includegraphics[width=80mm]{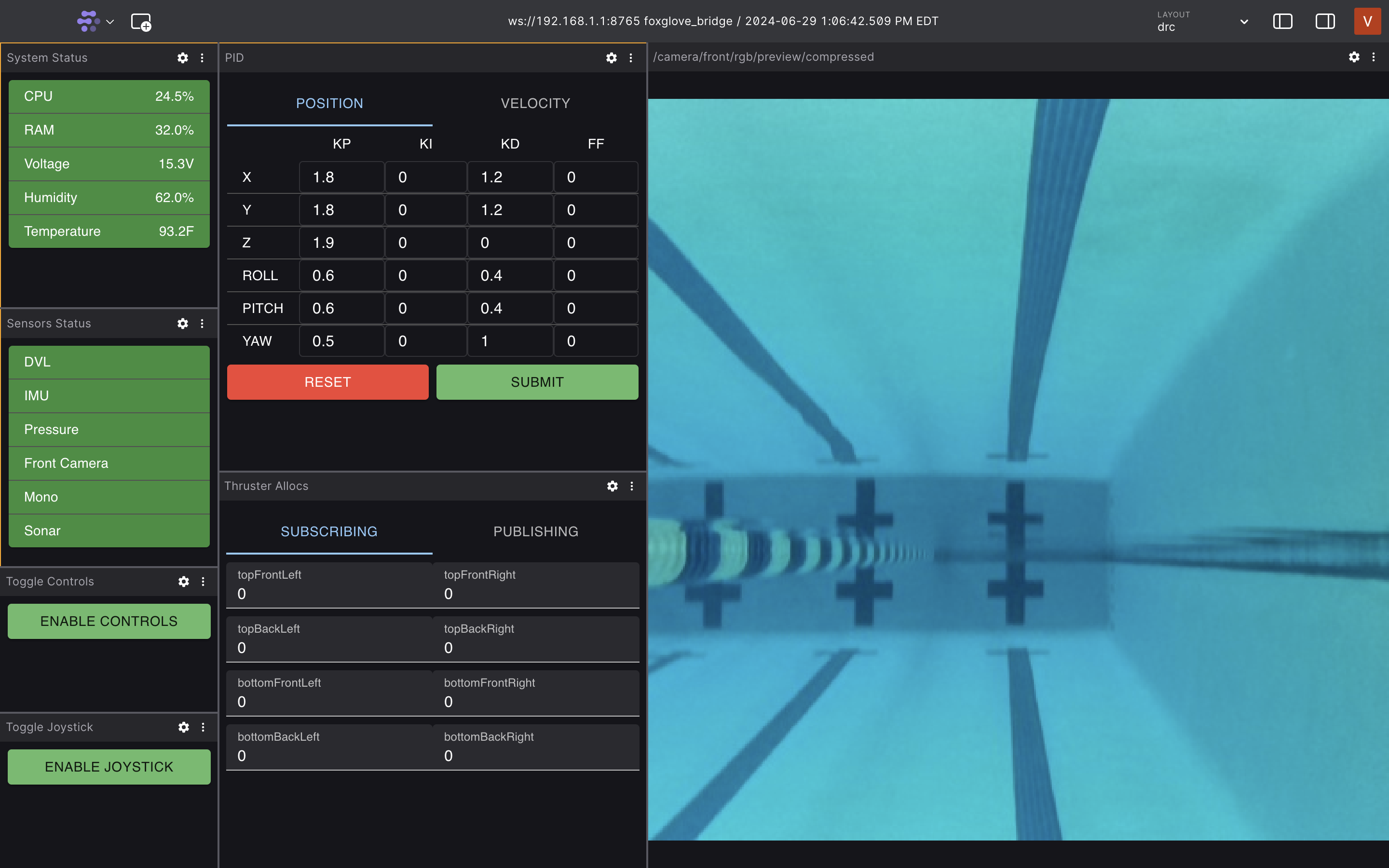}
\caption{Standard Foxglove GUI Layout. Our GUI allows us to easily visualize and control Oogway during pool tests. For example, operators can view live bounding boxes, monitor the status of subsystems, and fine-tune PID constants on the fly.}
\label{fig:gui}
\end{figure}

\vspace{-10pt}

\begin{figure}[h!]
\centering
\includegraphics[width=60mm]{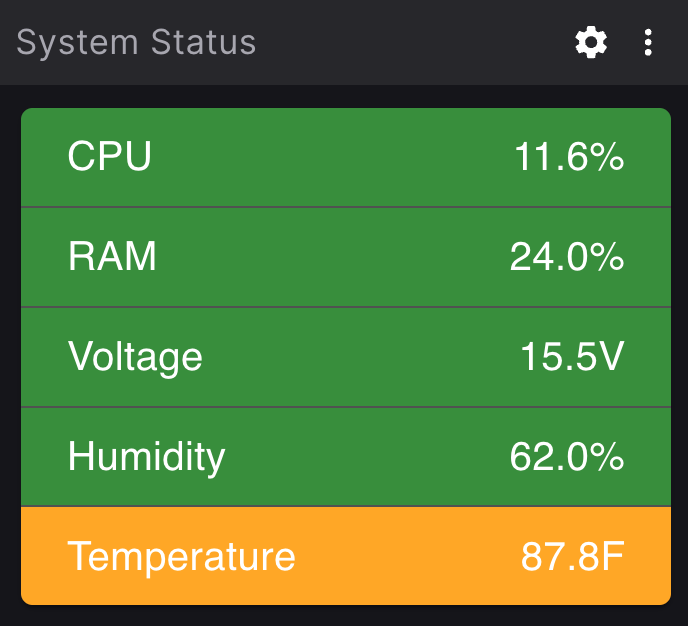}
\caption{The System Status Panel provides real-time information about system components, such as the overall temperature and humidity of the electrical stack. Subsystems operating outside their expected parameters are highlighted, enabling operators to address potential issues proactively.}
\end{figure}

\vspace{-10pt}

\begin{figure}[h!]
\centering
\includegraphics[width=60mm]{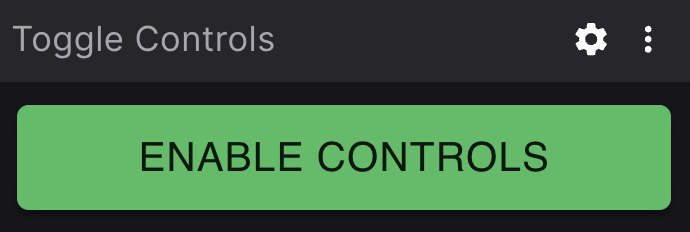}
\caption{The Toggle Controls Panel serves as a software kill-switch for our robots’ control systems, ensuring immediate response to emergency situations or unexpected behavior.}
\end{figure}

\vspace{-10pt}

\begin{figure}[h!]
\centering
\includegraphics[width=60mm]{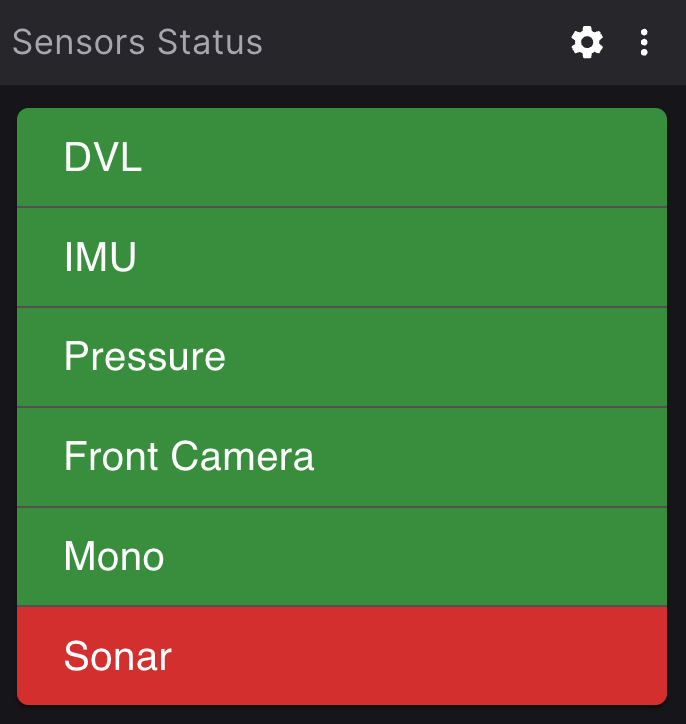}
\caption{The Sensors Status Panel continuously checks when the last message from each sensor was published. Sensors currently active are highlighted in green, while those not publishing appear in red.}
\end{figure}

\vspace{-10pt}

\begin{figure}[h!]
\centering
\includegraphics[width=60mm]{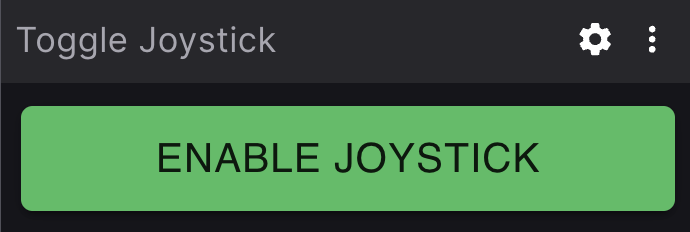}
\caption{The Toggle Joystick Panel maps inputs from a Thrustmaster T.Flight HOTAS X joystick to a desired power message, allowing operators to move Oogway in six degrees of freedom.}
\end{figure}

\begin{figure}[h!]
\centering
\includegraphics[width=60mm]{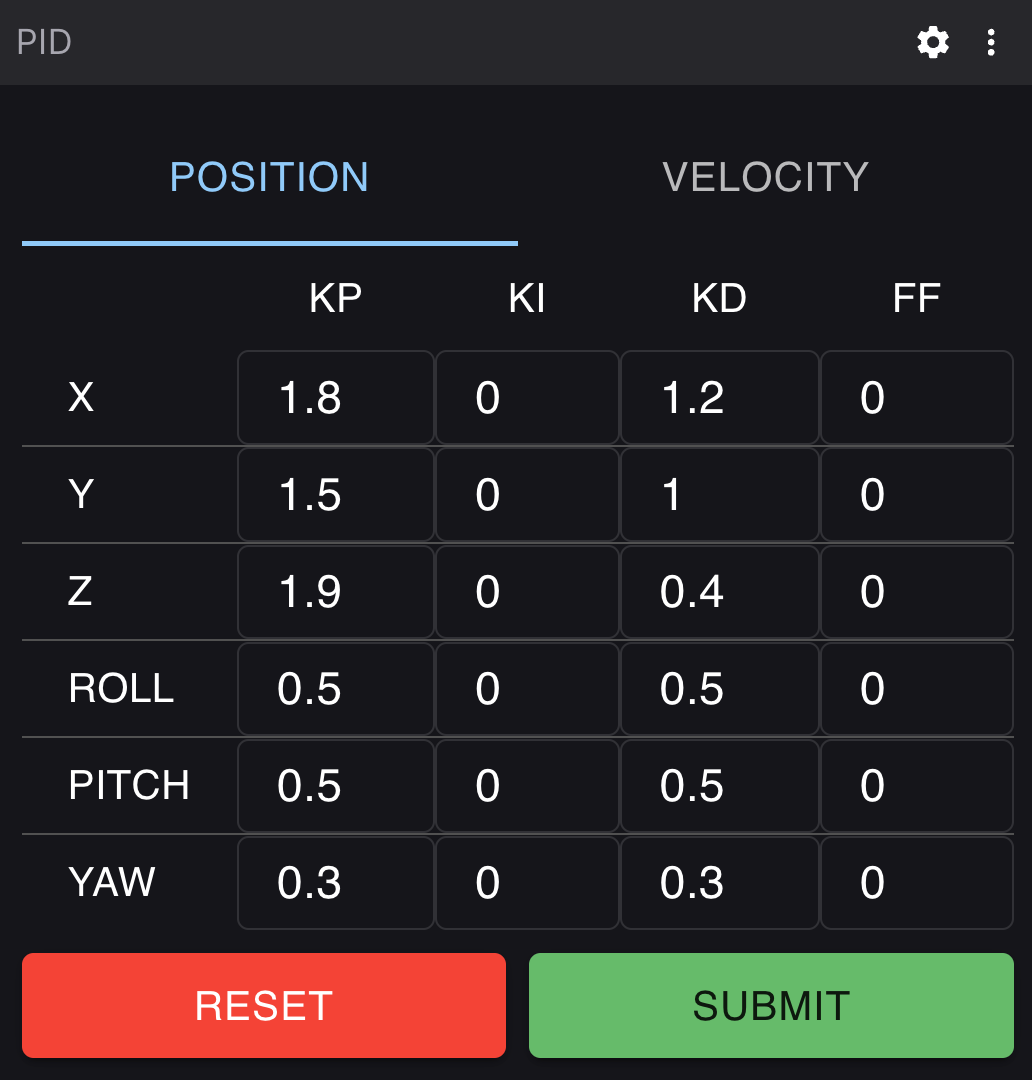}
\caption{The PID Panel allows an operator to read and set PID gains. Operators can thus view real-time data using the PID Position Layout (see Fig.~\ref{fig:plots}) and adjust PID constants using the PID Panel to instantly see how they affect Oogway's movement.}
\end{figure}

\onecolumn
\begin{figure}[h!]
\centering
\includegraphics[width=0.8\textwidth]{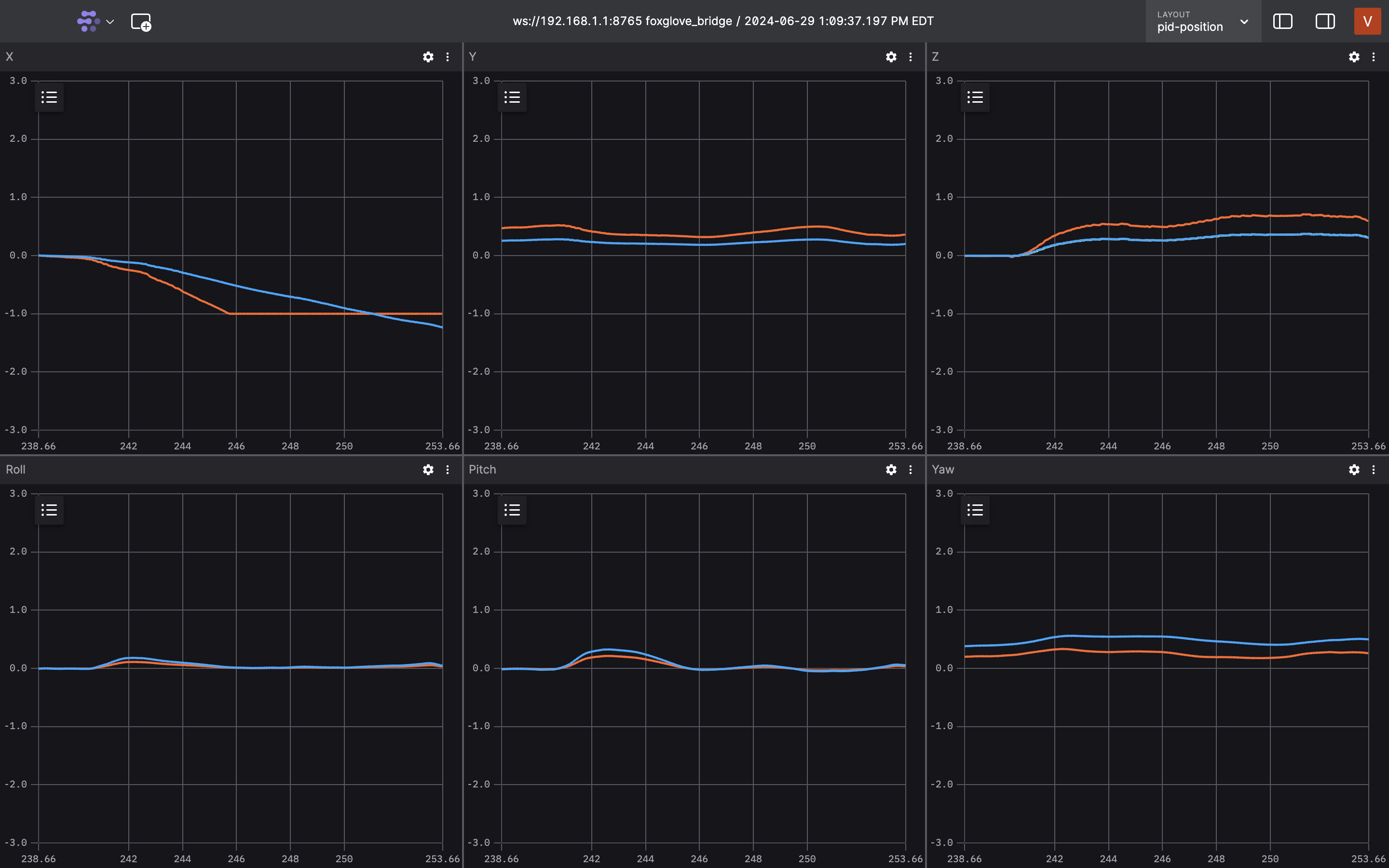}
\caption{PID Position Layout. Error is plotted in blue; control effort is plotted in orange. Each subplot corresponds to one of the six degrees of freedom: x, y, z, roll, pitch, and yaw.}
\label{fig:pid-plots}
\end{figure}

\begin{figure}[h!]
\centering
\includegraphics[width=0.8\textwidth]{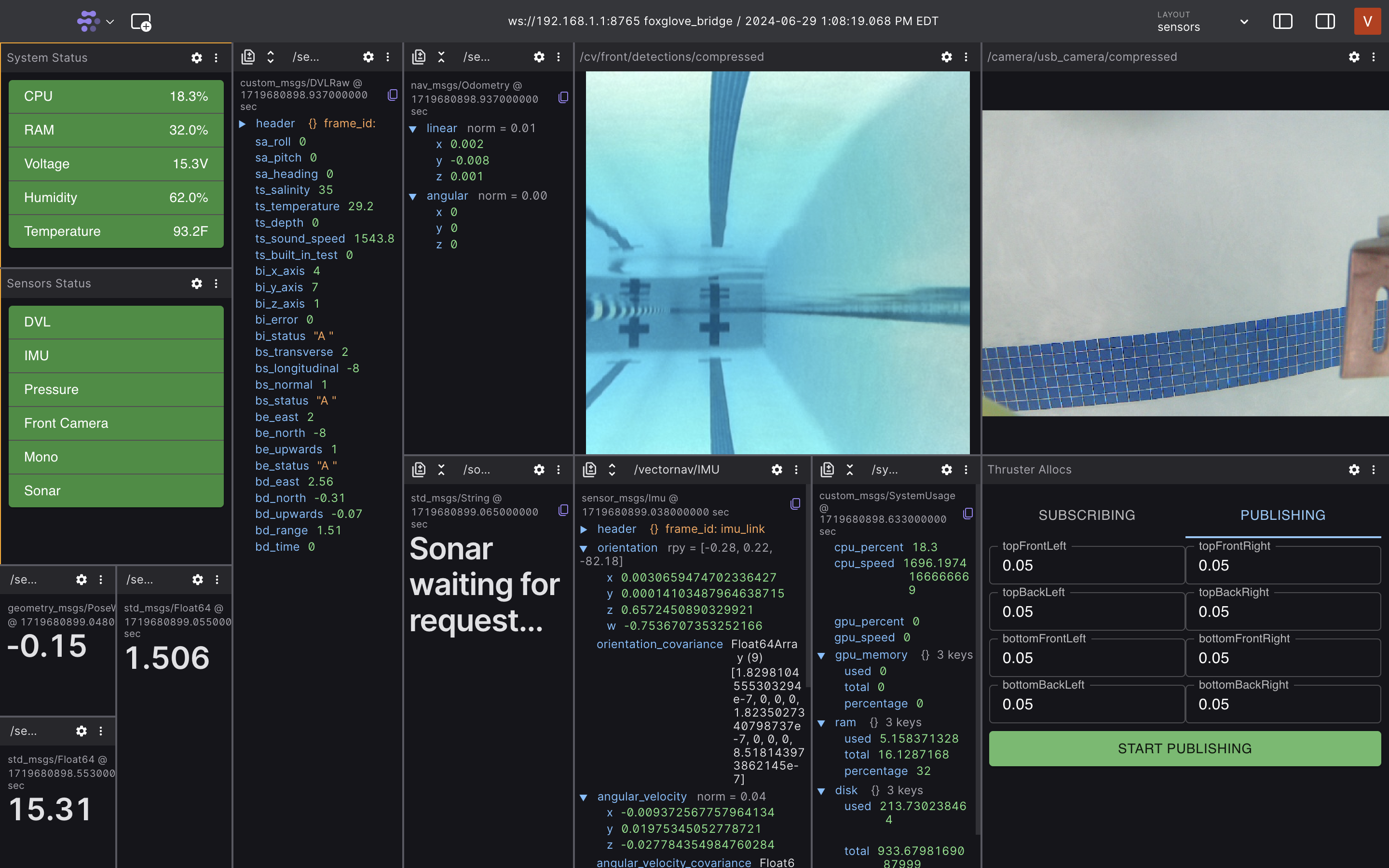}
\caption{Sensors Layout. This layout includes the System Status and Sensors Status panels along with panels to observe individual, raw sensor feed.}
\label{fig:sensors-layout}
\end{figure}

\newpage
\section*{Appendix D: Full Electrical System Diagrams}

\clearpage

\begin{figure}[h!]
\centering
\includegraphics[width=0.75\textwidth]{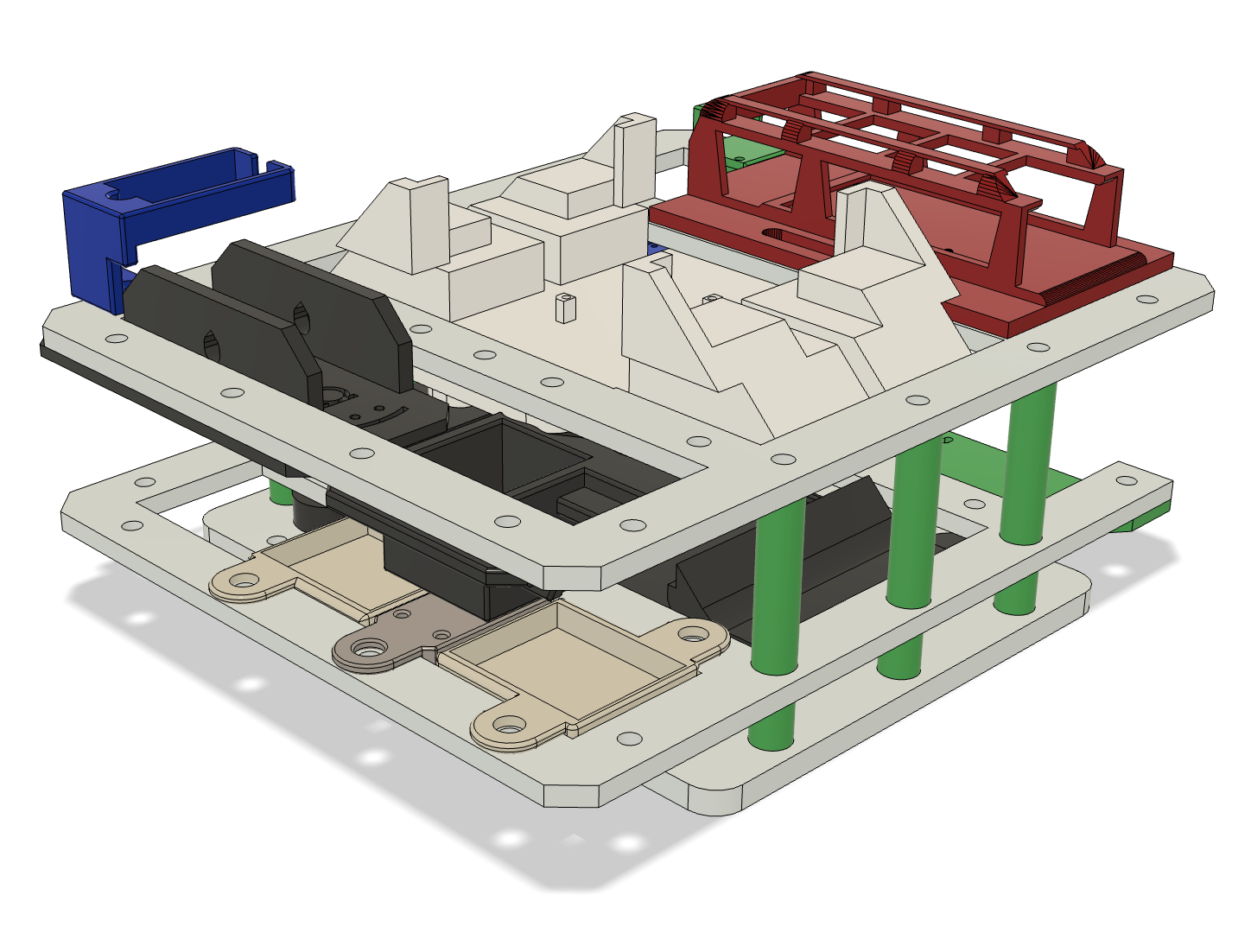}
\caption{New Electrical Stack Mounts -- CAD}
\label{fig:newstackcad}
\end{figure}

\begin{figure}[h!]
\centering
\includegraphics[width=0.75\textwidth]{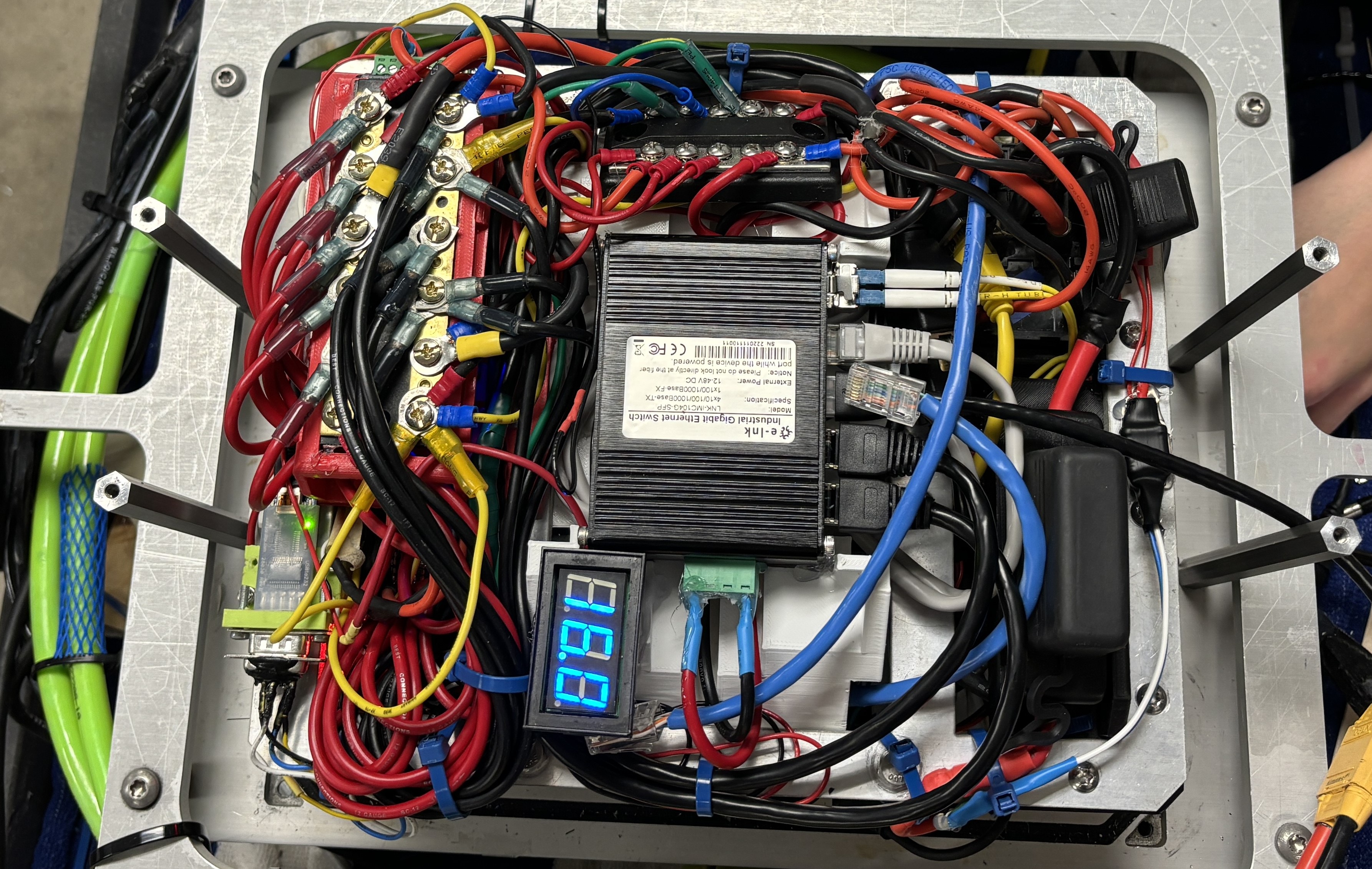}
\caption{New Electrical Stack -- Populated, Top View}
\label{fig:newstack}

\end{figure}
\twocolumn

\newpage
\onecolumn
\section*{Appendix E: Sonar Pipeline}

\begin{figure}[h!]
\centering
\includegraphics[width=0.8\textwidth]{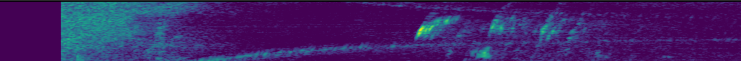}
\caption{Original sonar sweep data of a flat panel perpendicular to the robot underwater. The panel can be seen as the first light green blob in the image. The following darker green blobs are acoustic reflections.}
\label{fig:plots}
\end{figure}

\begin{figure}[h!]
\centering
\includegraphics[width=0.8\textwidth]{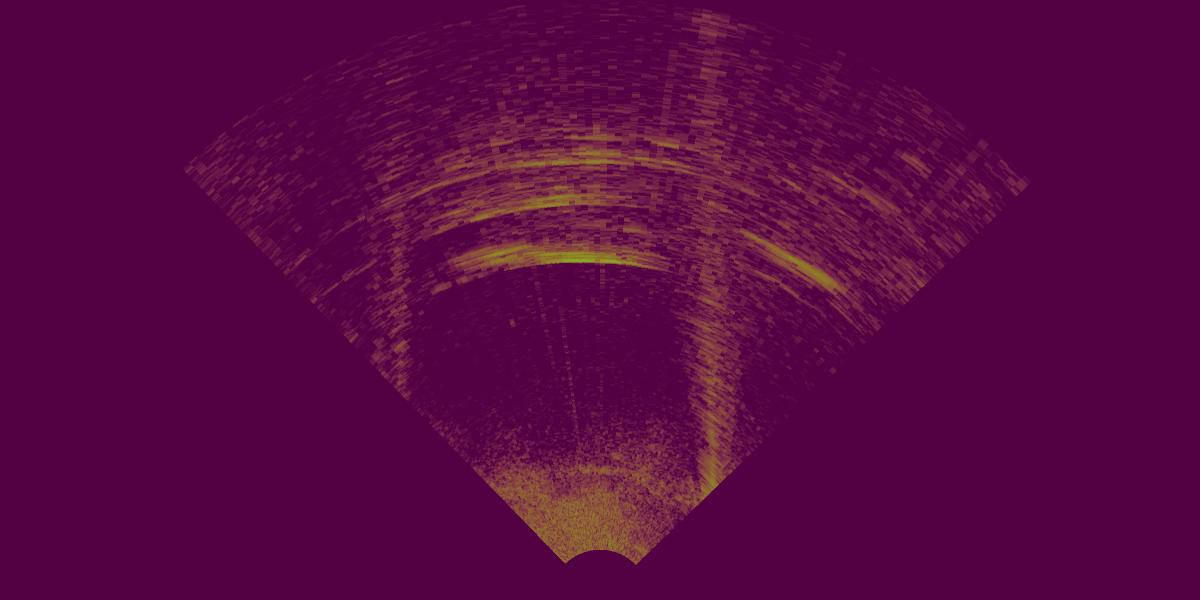}
\caption{Conversion from polar image to Cartesian image. Curved lines in the polar image become straight lines in the Cartesian image which allows for easy classification. The straight line on the right of the image is the wall of the pool.}
\label{fig:buoy_detection}
\end{figure}

\begin{figure}[h!]
\centering
\includegraphics[width=0.8\textwidth]{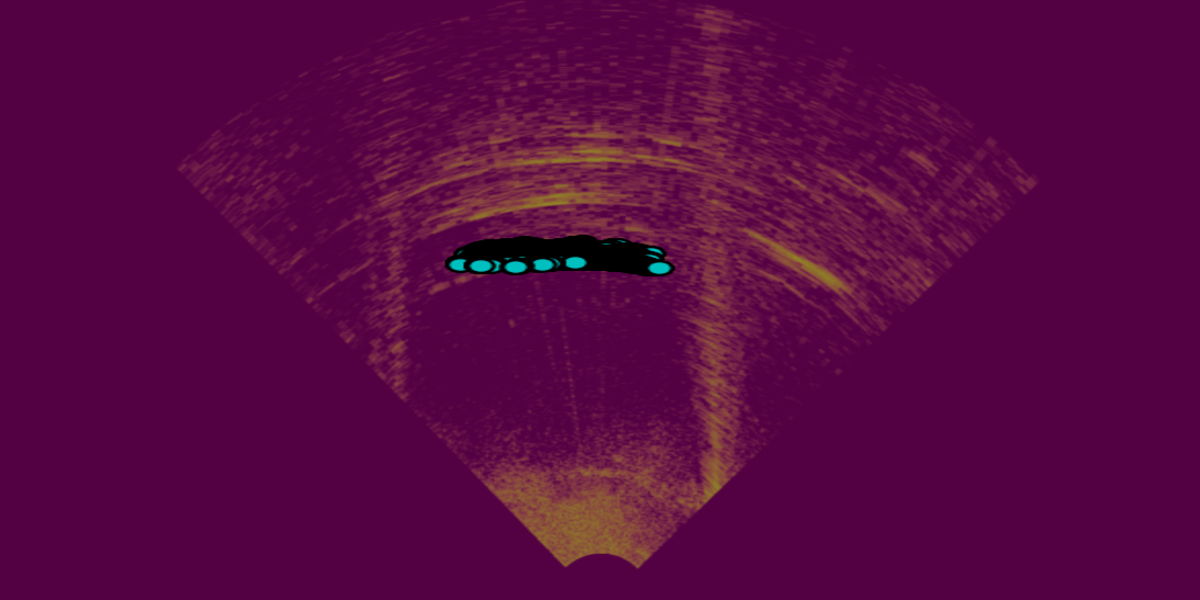}
\caption{Panel identified using cyan circles. The blobs behind the panel are ignored as they are acoustic reflections.}
\label{fig:buoy_detection}
\end{figure}

\newpage
\section*{Appendix F: Control Flow Diagram}

\newpage
\section*{Appendix G: Controls Overview Diagram}

\newpage
\section*{Appendix H: Oogway Render and Photo}

\begin{figure}[h!]
\centering
\includegraphics[width=0.67\textwidth]{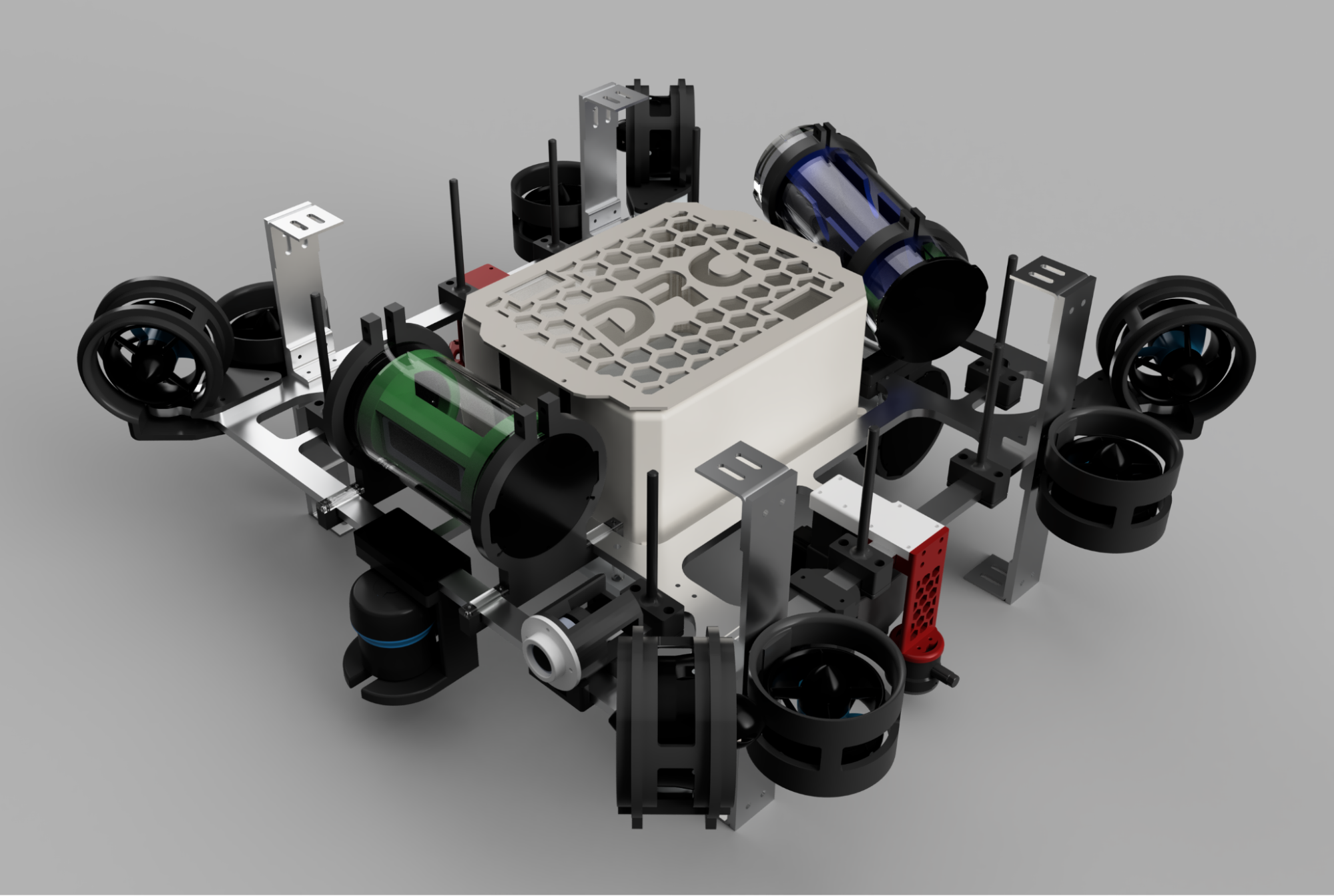}
\caption{Render of Oogway}
\label{fig:buoy_detection}
\end{figure}

\begin{figure}[h!]
\centering
\includegraphics[width=0.67\textwidth]{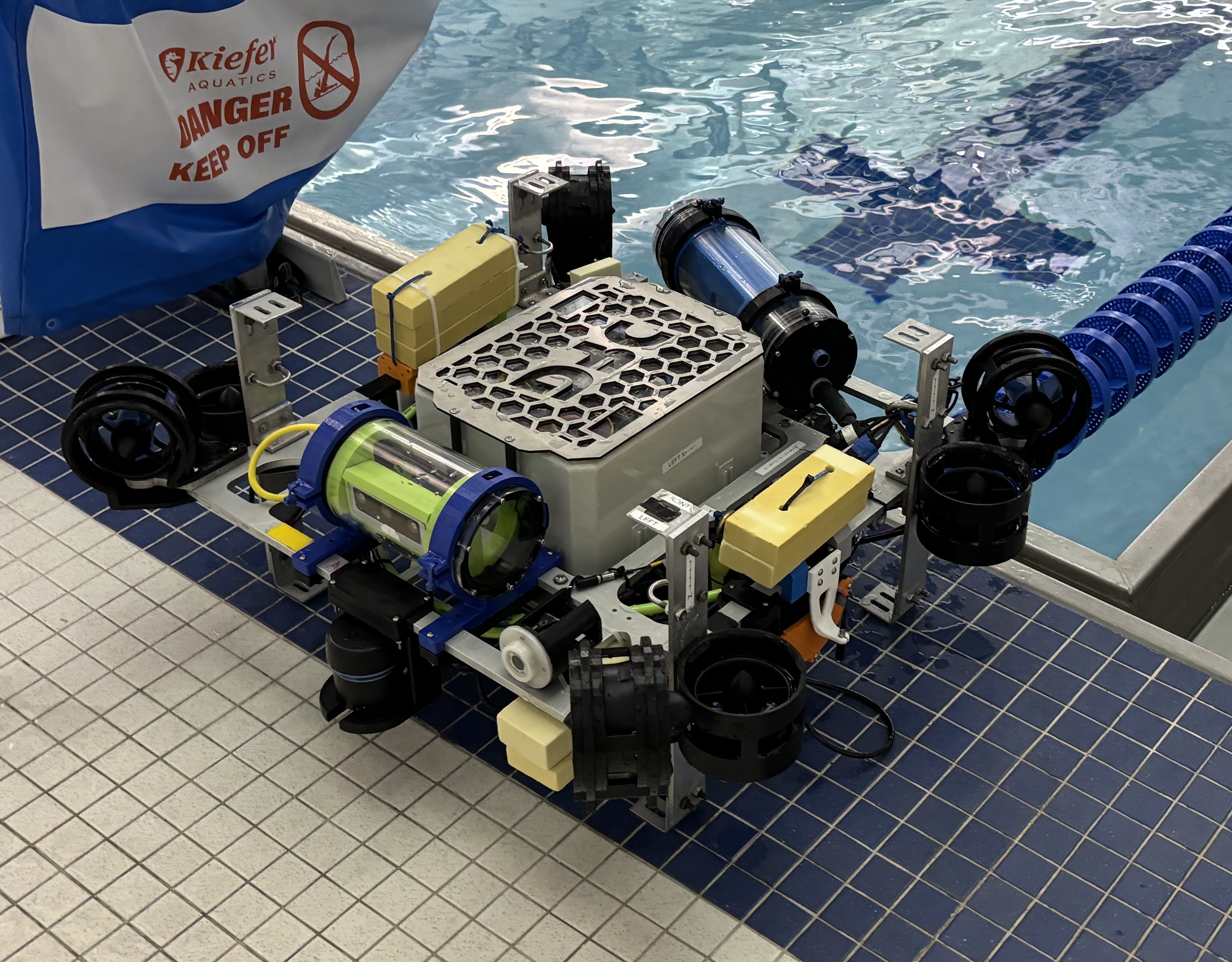}
\caption{Photo of Oogway}
\label{fig:oogway-water}
\end{figure}

\newpage
\section*{Appendix I: Community Outreach}

The Duke Robotics Club has a longstanding commitment to improving our community and fostering young students’ interest in STEM fields. In the past we have volunteered in middle schools, helped Durham high schoolers learn about coding, and mentored middle and high school robotics teams. This year, multiple Duke Robotics members are proud mentors of Valence Robotics, a FIRST® (For Inspiration and Recognition of Science and Technology) robotics team. The FIRST Robotics Competition (FRC) is an annual competition specifically aimed at high school students with the intent to inspire them to pursue a career in STEM. Duke Robotics members regularly attend Valence Robotics’ weekly meetings, helping explain advanced concepts to them, such as machining and software task planning. Many members of our team were part of FIRST themselves, and we value giving back to a community that has impacted our lives so much. 

In our second year mentoring the team, it has been inspiring witnessing the students grow as leaders and engineers within the team. Our approach to mentorship is completely hands-off, letting them make mistakes and learn from them–an approach that we have found most valuable to our engineering education in the past. Throughout the season, they have gained valuable insights into the iterative design process, confronting challenges and refining their robot design through practical experience. We always make ourselves available for an in depth conversation on design or just life in general, fostering a deep connection between our teams. 

We value our outreach as an important part of our team culture. By actively participating in mentorship and outreach programs, we not only share our passion for engineering but also continuously refine our own skills as engineers and communicators. Moreover, the relationships we build through these initiatives foster a sense of responsibility and commitment to the broader community. As we look to the future, we remain dedicated to expanding our impact and inspiring others to embrace the possibilities of robotics.

\begin{figure}[h!]
\centering
\includegraphics[width=0.67\textwidth]{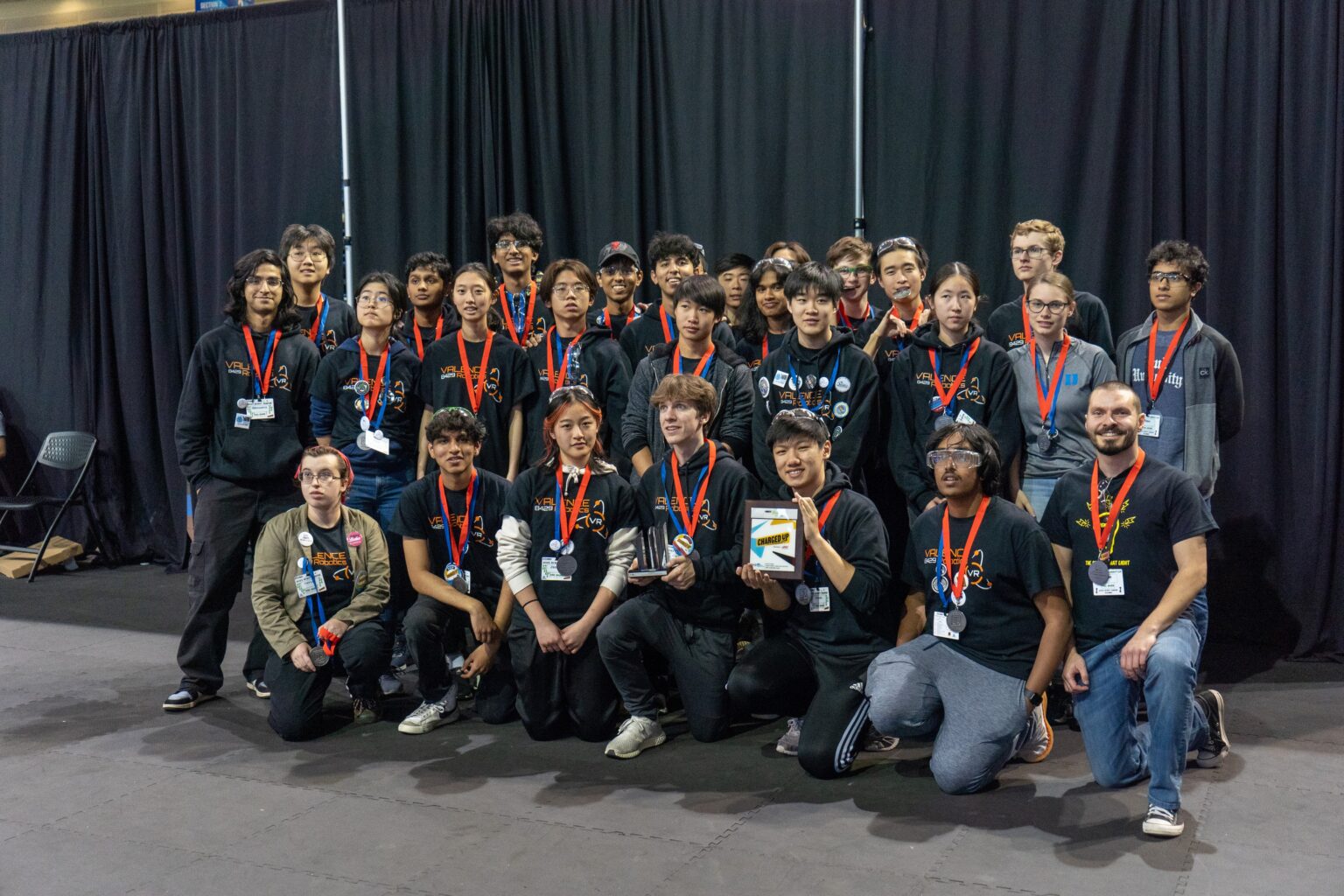}
\caption{Members of Duke Robotics with Valence Robotics at an FRC competition.}
\label{fig:valence_team}
\end{figure}

\end{document}